\UseRawInputEncoding

\documentclass{article}
\usepackage{amsmath,graphicx,amsthm}
\usepackage{lineno,hyperref}
\usepackage{epsfig}
\usepackage{amssymb, amsfonts,mathrsfs,euscript, bm}
\usepackage{algorithm}
\usepackage{algorithmic}
\usepackage{subfigure}
\usepackage{amsthm}
\usepackage{url, color, soul}

\usepackage{multirow}
\usepackage{array}
\usepackage{breqn}
\usepackage{hyperref}
\usepackage{amssymb}
\usepackage{comment}
\usepackage{float}
\usepackage{times}
\usepackage{float}

\newtheorem{theorem}{Theorem}
\newtheorem{definition}{Definition}
\newtheorem{lemma}{Lemma}
\newtheorem{corollary}{Corollary}
\newtheorem{assumption}{Assumption}

\newtheoremstyle{remarkupright}%
  {}{}            
  {\normalfont}   
  {}              
  {\bfseries}     
  {.}             
  {.5em}          
  {}              

\theoremstyle{remarkupright}
\newtheorem{remark}{Remark}

\title{A Unified Analysis of Generalization and Sample Complexity for Semi-Supervised Domain Adaptation}
\author{Elif Vural\thanks{Department of Electrical and Electronics Engineering, METU, Ankara}, H\"useyin Karaca\thanks{Department of Electrical and Electronics Engineering, Bilkent University, Ankara}}
\date{}

\newcommand{\R}{\ensuremath{\mathbb{R}}}

\newcommand{\B}{\ensuremath{\mathcal{B}}}
\newcommand{\wvect}{\ensuremath{\mathbf{w}}}
\newcommand{\Xs}{\ensuremath{\mathcal{X}^s}} 
\newcommand{\Xt}{\ensuremath{\mathcal{X}^t}} 
\newcommand{\Y}{\ensuremath{\mathcal{Y}}} 
\newcommand{\Zs}{\ensuremath{\mathcal{Z}^s}} 
\newcommand{\Zt}{\ensuremath{\mathcal{Z}^t}} 
\newcommand{\mus}{\ensuremath{\mu_s}} 
\newcommand{\mut}{\ensuremath{\mu_t}} 
\newcommand{\nus}{\ensuremath{\nu_s}} 
\newcommand{\nut}{\ensuremath{\nu_t}} 

\newcommand{\vars}{\ensuremath{\sigma_s^2}}
\newcommand{\vart}{\ensuremath{\sigma_t^2}}
\newcommand{\Cs}{\ensuremath{C_s}}
\newcommand{\Ct}{\ensuremath{C_t}}

\newcommand{\xs}{\ensuremath{x^s}} 
\newcommand{\xt}{\ensuremath{x^t}} 
\newcommand{\xis}{\ensuremath{x^s_i}} 
\newcommand{\xjs}{\ensuremath{x^s_j}} 
\newcommand{\xjt}{\ensuremath{x^t_j}} 
\newcommand{\xit}{\ensuremath{x^t_i}} 
\newcommand{\ys}{\ensuremath{\mathbf{y}^s}} 
\newcommand{\yt}{\ensuremath{\mathbf{y}^t}} 
\newcommand{\yis}{\ensuremath{\mathbf{y}^s_i}} 
\newcommand{\y}{\ensuremath{\mathbf{y}}} 
\newcommand{\yjt}{\ensuremath{\mathbf{y}^t_j}} 
\newcommand{\Ns}{\ensuremath{N_s}} 
\newcommand{\Nt}{\ensuremath{N_t}} 
\newcommand{\Ms}{\ensuremath{M_s}} 
\newcommand{\Mt}{\ensuremath{M_t}} 
\newcommand{\as}{\ensuremath{a_s}}
\newcommand{\at}{\ensuremath{a_t}}
\newcommand{\X}{\ensuremath{\mathcal{X}}} 
\newcommand{\fs}{\ensuremath{f^s}} 
\newcommand{\ft}{\ensuremath{f^t}} 
\newcommand{\h}{\ensuremath{h}} 
\newcommand{\Fs}{\ensuremath{\mathcal{F}^s}} 
\newcommand{\Ft}{\ensuremath{\mathcal{F}^t}} 
\newcommand{\Hs}{\ensuremath{\mathcal{H}}} 
\newcommand{\F}{\ensuremath{\mathcal{F}}} 
\newcommand{\gs}{\ensuremath{g^s}} 
\newcommand{\gt}{\ensuremath{g^t}} 
 \newcommand{\Gs}{\ensuremath{\mathcal{G}^s}} 
\newcommand{\Gt}{\ensuremath{\mathcal{G}^t}} 
 \newcommand{\Dspace}{\ensuremath{\mathcal{D}}} 
 \newcommand{\vs}{\ensuremath{v^s}} 
\newcommand{\vt}{\ensuremath{v^t}} 
 \newcommand{\Vs}{\ensuremath{\mathcal{V}^s}} 
\newcommand{\Vt}{\ensuremath{\mathcal{V}^t}}

\newcommand{\loss}{\ensuremath{\ell}} 
\newcommand{\Ls}{\ensuremath{\mathcal{L}^s}} 
\newcommand{\Lt}{\ensuremath{\mathcal{L}^t}} 
\newcommand{\Lw}{\ensuremath{\mathcal{L}_{\alpha}}} 
\newcommand{\hLs}{\ensuremath{\hat{\mathcal{L}}^s}} 
\newcommand{\hLt}{\ensuremath{\hat{\mathcal{L}}^t}} 
\newcommand{\hLw}{\ensuremath{\hat{\mathcal{L}}_{\alpha}}} 
\newcommand{\D}{\ensuremath{D}} 
\newcommand{\Ddan}{\ensuremath{\D_\ddan}} 
\newcommand{\hD}{\ensuremath{\hat{D}}} 
\newcommand{\hDdan}{\ensuremath{\hat{\D}_\ddan}} 

\newcommand{\LLs}{\ensuremath{R}} 
\newcommand{\Lls}{\ensuremath{L_\ell}} 
\newcommand{\Lact}{\ensuremath{L}} 
\newcommand{\bls}{\ensuremath{A_\ell}} 
\newcommand{\Lk}{\ensuremath{L_K}} 
\newcommand{\LLsdan}{\ensuremath{\LLs_{A}}}

\newcommand{\N}{\ensuremath{\mathcal{N}}}

\newcommand{\dmetric}{\ensuremath{\mathfrak{d}}} 
\newcommand{\ds}{\ensuremath{\mathfrak{d}^s}} 
\newcommand{\dt}{\ensuremath{\mathfrak{d}^t}} 
\newcommand{\dXs}{\ensuremath{\ds_{\mathcal{X}} }} 
\newcommand{\dXt}{\ensuremath{\dt_{\mathcal{X}} }} 
\newcommand{\dVs}{\ensuremath{\ds_{\mathcal{V}} }} 
\newcommand{\dVt}{\ensuremath{\dt_{\mathcal{V}} }} 
\newcommand{\Ks}{\ensuremath{\kappa^s}} 
\newcommand{\Kt}{\ensuremath{\kappa^t}} 
\newcommand{\Kl}{\ensuremath{\kappa^l}} 
\newcommand{\Kgen}{\ensuremath{\kappa}} 
\newcommand{\radTheta}{\ensuremath{\delta}} 
\newcommand{\grid}{\ensuremath{\mathfrak{G}}}

\newcommand{\fsl}{\ensuremath{f^{sl}}} 
\newcommand{\ftl}{\ensuremath{f^{tl}}} 
\newcommand{\fsone}{\ensuremath{f^{s1}}} 
\newcommand{\ftone}{\ensuremath{f^{t1}}} 
\newcommand{\fsLm}{\ensuremath{f^{s(L-1)}}} 
\newcommand{\ftLm}{\ensuremath{f^{t(L-1)}}} 

\newcommand{\hsl}{\ensuremath{\boldsymbol{\xi}^{sl}}} 
\newcommand{\htl}{\ensuremath{\boldsymbol{\xi}^{tl}}} 
\newcommand{\hslm}{\ensuremath{\boldsymbol{\xi}^{s(l-1)}}} 
\newcommand{\htlm}{\ensuremath{\boldsymbol{\xi}^{t(l-1)}}} 
\newcommand{\hsz}{\ensuremath{\boldsymbol{\xi}^{s0}}}
\newcommand{\hsone}{\ensuremath{\boldsymbol{\xi}^{s1}}}
\newcommand{\htz}{\ensuremath{\boldsymbol{\xi}^{t0}}}
\newcommand{\hidl}{\ensuremath{\boldsymbol{\xi}^{l}}} 
\newcommand{\hidlm}{\ensuremath{\boldsymbol{\xi}^{l-1}}} 
\newcommand{\hidz}{\ensuremath{\boldsymbol{\xi}^{0}}}
\newcommand{\hidone}{\ensuremath{\boldsymbol{\xi}^{1}}}
\newcommand{\hidL}{\ensuremath{\boldsymbol{\xi}^{L}}}
\newcommand{\hsL}{\ensuremath{\boldsymbol{\xi}^{sL}}} 
\newcommand{\htL}{\ensuremath{\boldsymbol{\xi}^{tL}}}
\newcommand{\hsLm}{\ensuremath{\boldsymbol{\xi}^{s(L-1)}}} 
\newcommand{\htLm}{\ensuremath{\boldsymbol{\xi}^{t(L-1)}}} 
\newcommand{\hidLm}{\ensuremath{\boldsymbol{\xi}^{L-1}}} 

\newcommand{\Wsl}{\ensuremath{\mathbf{W}^{sl}}} 
\newcommand{\Wtl}{\ensuremath{\mathbf{W}^{tl}}} 
\newcommand{\Wl}{\ensuremath{\mathbf{W}^{l}}} 
\newcommand{\WsL}{\ensuremath{\mathbf{W}^{sL}}} 
\newcommand{\WtL}{\ensuremath{\mathbf{W}^{tL}}} 
\newcommand{\WL}{\ensuremath{\mathbf{W}^{L}}} 
\newcommand{\Wone}{\ensuremath{\mathbf{W}^{1}}} 

\newcommand{\Wldan}{\ensuremath{\mathbf{W}^{l}_\ddan}}

\newcommand{\bldan}{\ensuremath{\mathbf{b}^{l}_\ddan}}

\newcommand{\Thetasl}{\ensuremath{\boldsymbol{\Theta}^{sl}}} 
\newcommand{\Thetatl}{\ensuremath{\boldsymbol{\Theta}^{tl}}} 
\newcommand{\Thetas}{\ensuremath{\boldsymbol{\Theta}^{s}}} 
\newcommand{\Thetat}{\ensuremath{\boldsymbol{\Theta}^{t}}} 
\newcommand{\ThetaCom}{\ensuremath{\boldsymbol{\Theta}}} 
\newcommand{\ThetaSetl}{\ensuremath{\mathbf{\boldsymbol{\Theta}}^l}} 
\newcommand{\Thetal}{\ensuremath{\boldsymbol{\Theta}^{l}}} 
\newcommand{\Phis}{\ensuremath{\mathbf{\Phi}^{s}}} 
 
\newcommand{\PhiCom}{\ensuremath{\mathbf{\Phi}}} 
\newcommand{\mapFs}{\ensuremath{\mathcal{M}_{\Fs}}} 
\newcommand{\mapGs}{\ensuremath{\mathcal{M}_{\Gs}}} 

\newcommand{\bsl}{\ensuremath{\mathbf{b}^{sl}}} 
\newcommand{\btl}{\ensuremath{\mathbf{b}^{tl}}} 
\newcommand{\bl}{\ensuremath{\mathbf{b}^{l}}} 
\newcommand{\bsL}{\ensuremath{\mathbf{b}^{sL}}} 
\newcommand{\btL}{\ensuremath{\mathbf{b}^{tL}}} 
\newcommand{\bL}{\ensuremath{\mathbf{b}^{L}}} 
\newcommand{\bone}{\ensuremath{\mathbf{b}^{1}}} 

\newcommand{\numL}{\ensuremath{L}} 
\newcommand{\numLdan}{\ensuremath{K}} 
\newcommand{\actl}{\ensuremath{\eta^l}} 
\newcommand{\act}{\ensuremath{\eta}} 
\newcommand{\actone}{\ensuremath{\eta^1}} 
\newcommand{\lay}{\ensuremath{l}} 

\newcommand{\dl}{\ensuremath{{d_l}}}
\newcommand{\dlm}{\ensuremath{{d_{l-1}}}}
\newcommand{\dz}{\ensuremath{{d_0}}}
\newcommand{\done}{\ensuremath{{d_1}}}
\newcommand{\dk}{\ensuremath{{d_k}}}
\newcommand{\dkm}{\ensuremath{{d_{k-1}}}}
\newcommand{\di}{\ensuremath{{d_i}}}
\newcommand{\dimm}{\ensuremath{{d_{i-1}}}}
\newcommand{\dkplone}{\ensuremath{{d_{k+1}}}}
\newcommand{\dL}{\ensuremath{{d_L}}}
\newcommand{\dLm}{\ensuremath{{d_{L-1}}}}
\newcommand{\dlmax}{\ensuremath{{d_{\max}}}}
\newcommand{\Rdl}{\ensuremath{\mathbb{R}^{d_l}}}
\newcommand{\dcom}{\ensuremath{{d}}}

\newcommand{\dlddan}{\ensuremath{{d^{\ddan}_l}}}
\newcommand{\dlmddan}{\ensuremath{{d^{\ddan}_{l-1}}}}
\newcommand{\dKddan}{\ensuremath{{d^{\ddan}_K}}}
\newcommand{\Bnet}{\ensuremath{{A_\Theta}}} 
\newcommand{\Leta}{\ensuremath{{L_\eta}}} 
\newcommand{\Binp}{\ensuremath{{A_x}}}  
\newcommand{\Beta}{\ensuremath{{C_\eta}}} 
\newcommand{\Bddan}{\ensuremath{{C_\Dspace}}} 
\newcommand{\Bddansq}{\ensuremath{{C^2_\Dspace}}} 

\newcommand{\Bopeta}{\ensuremath{{A_\eta}}} 
\newcommand{\Bdiml}{\ensuremath{{R_l}}} 
\newcommand{\Bdimlm}{\ensuremath{{R_{l-1}}}} 
\newcommand{\BdimLm}{\ensuremath{{R_{L-1}}}} 
\newcommand{\Bdimz}{\ensuremath{{R_0}}}
\newcommand{\Bdimone}{\ensuremath{{R_1}}}
 
\newcommand{\Bdimim}{\ensuremath{{R_{i-1}}}} 
\newcommand{\BQdiml}{\ensuremath{{Q_l}}} 
\newcommand{\BQdimone}{\ensuremath{{Q_1}}} 
\newcommand{\BQdimL}{\ensuremath{{Q_L}}} 
\newcommand{\BQdimLm}{\ensuremath{{Q_{L-1}}}} 
\newcommand{\BQ}{\ensuremath{{Q}}}

\newcommand{\phil}{\ensuremath{\phi^l}} 
\newcommand{\kr}{\ensuremath{k}} 
\newcommand{\krl}{\ensuremath{k^l}} 
\newcommand{\Xl}{\ensuremath{\mathcal{X}^l}}

\newcommand{\ddan}{\ensuremath{\Delta}} 
\newcommand{\dom}{\ensuremath{\mathcal{\D}}} 
\newcommand{\ydoms}{\ensuremath{l^s}} 
\newcommand{\ydomt}{\ensuremath{l^t}} 
\newcommand{\actldan}{\ensuremath{\eta^l_\ddan}} 
\newcommand{\hidldan}{\ensuremath{\boldsymbol{\xi}^{l}_\ddan}}
\newcommand{\hidlmdan}{\ensuremath{\boldsymbol{\xi}^{l-1}_\ddan}}
\newcommand{\hidzdan}{\ensuremath{\boldsymbol{\xi}^{0}_\ddan}}
\newcommand{\hidKdan}{\ensuremath{\boldsymbol{\xi}^{K}_\ddan}}
\newcommand{\actKdan}{\ensuremath{\eta^\numLdan_\ddan}} 

\begin{document}
%

\maketitle
\begin{abstract}

Domain adaptation seeks to leverage the abundant label information in a source domain to improve classification performance in a target domain with limited labels. While the field has seen extensive methodological development, its theoretical foundations remain relatively underexplored. Most existing theoretical analyses focus on simplified settings where the source and target domains share the same input space and relate target-domain performance to measures of domain discrepancy. Although insightful, these analyses may not fully capture the behavior of modern approaches that align domains into a shared space via feature transformations. In this paper, we present a comprehensive theoretical study of domain adaptation algorithms based on \textit{domain alignment}. We consider the joint learning of domain-aligning feature transformations and a shared classifier in a semi-supervised setting. We first derive generalization bounds in a broad setting, in terms of covering numbers of the relevant function classes. We then extend our analysis to characterize the sample complexity of domain-adaptive neural networks employing maximum mean discrepancy (MMD) or adversarial objectives. Our results rely on a rigorous analysis of the covering numbers of these architectures. We show that, for both MMD-based and adversarial models, the sample complexity admits an upper bound that scales quadratically with network depth and width. Furthermore, our analysis suggests that in semi-supervised settings, robustness to limited labeled target data can be achieved by scaling the target loss proportionally to the square root of the number of labeled target samples. Experimental evaluation in both shallow and deep settings lends support to our theoretical findings.

\end{abstract}
%

\section{Introduction}
\label{sec:intro}

Domain adaptation is a subfield of machine learning that aims to improve model performance in a target domain by leveraging the greater availability of labeled samples in a source domain. The main challenge in domain adaptation is to address the discrepancy between the source and target distributions, which can take various forms such as covariate shift \cite{KouwL21}, label shift \cite{Azizzadenesheli19}, \cite{Combes0WG20}, as well as more challenging heterogeneous settings  with source and target samples originating from different data spaces \cite{SinghalWRK23}. 
Early work in domain adaptation explored instance reweighting methods for covariate shift \cite{HuangSGBS06}, \cite{SunCPY11}, feature augmentation approaches \cite{DaumeIII07}, \cite{DaumeKS10}, \cite{DuanXT12}, and techniques for learning feature projections or transformations \cite{BaktashmotlaghHLS13},  \cite{PanTKY11}, \cite{YaoPNLM15}. More recently, in line with broader advances in data science, domain adaptation research over the last decade has largely shifted towards deep learning-based techniques \cite{SinghalWRK23}, \cite{WangD18}.  Metrics such as maximum mean discrepancy (MMD)   \cite{LongCWJ15}, \cite{TzengHZSD14}, \cite{GhifaryKZ14} lead to efficient solutions for aligning source and target domains across various applications \cite{ZengSRGFZ25}, \cite{WangYXWZW23}, \cite{DXiaLZSL25}, \cite{YangZLLLC25}. Adversarial architectures \cite{GaninUAGLLML16}, \cite{TzengHSD17}, \cite{TangJ20},  \cite{ZonooziS23}  and reconstruction-based approaches using encoder-decoder structures \cite{GhifaryKZBL16}, \cite{BousmalisTSKE16}, \cite{ZonooziSD25} are also commonly employed. 

Despite the variety of models and the diversity of solutions, the basic paradigm in domain adaptation - whether using shallow methods or neural networks- often boils down to first aligning the source and target domains by mapping them to a common space through feature transformations, followed by learning a hypothesis function, typically a classifier, in that shared domain. The alignment of the source and target distributions is achieved by minimizing a suitably defined \textit{distribution distance} (also referred to as \textit{domain discrepancy} or \textit{distribution divergence}), with common choices including MMD \cite{LongCWJ15}, covariance-based metrics \cite{SunS16}, and the Wasserstein distance \cite{CourtyFTR17}, \cite{DamodaranKFTC18}, \cite{HamriBF25}. Although domain adaptation algorithms have been successfully applied across a wide range of fields  including computer vision, time-series analysis, and natural language processing  \cite{SinghalWRK23}, \cite{ZonooziS23}, surprisingly, the literature still lacks a thorough theoretical characterization of their performance. In particular, there is a notable gap in understanding the behavior of \textit{domain alignment algorithms}, which we define as methods that explicitly map source and target domains to a common representation through feature transformations. In this paper, we focus on this important class of algorithms, and aim to provide a rigorous theoretical analysis of their performance.

Most existing theoretical analyses focus on understanding how the discrepancy between source and target domains affects the target-domain performance of classifiers trained to perform well on the source domain \cite{RedkoMHS20}, \cite{BenDavidBCP06},  \cite{MansourMR09},  \cite{ZhangLLJ19},  \cite{DhouibRL20},  \cite{WangM24}. While these studies provide useful insight into how models trained with abundant source labels generalize to a target domain with limited or no labeled data, they inherently assume that source and target data reside in the same space. Consequently, their results do not straightforwardly extend to the prevalent framework where source and target domains are aligned through feature transformations or mappings -whether shallow or deep- prior to classification. Only a few studies have investigated the performance of domain alignment algorithms  \cite{ZhouTPT19}, \cite{FangLLZ23}, \cite{WangS15}; however, these works rather focus on specific transformation types, such as linear mappings \cite{ZhouTPT19} or location and scale changes \cite{WangS15}. Some literature has investigated the performance and sample complexity of transfer learning via deep learning approaches \cite{TomerWH16}, \cite{McNamaraB17}, \cite{JiaoLLY24}. However, domain adaptation and transfer learning remain distinct problems:  transfer learning deals with  differing source and target tasks, unlike domain adaptation. Notably, the characterization of the sample complexity of domain-adaptive neural networks remains an important yet largely unexplored subject in current learning theory. It is well established that the amount of data required to successfully train a neural network increases with the size of the network to prevent overfitting, and many studies have addressed this issue in classical single-domain settings \cite{AnthonyB02}, \cite{NeyshaburTS15}, \cite{WeiM19}, \cite{VardiSS22}, \cite{DanielyG24}.  To the best of our knowledge, however, the scaling of labeled and unlabeled source and target sample requirements with respect to the width and depth of domain-adaptive networks has not been extensively studied yet.

In this work, we aim to fill this gap by providing a comprehensive theoretical analysis of domain adaptation in the widely used setting where the source and target domains are mapped to a common space through feature transformations, and a hypothesis is learnt in that shared space after alignment.  We consider a semi-supervised setting where labels are largely available for the source samples but limited (or unavailable) for the target samples. The structure of the paper along with our main contributions are summarized below:

\begin{itemize}

\item In Section \ref{sec:gen_bounds}, we study a general setting that involves learning a source feature transformation $\fs \in \Fs$, a target feature transformation $\ft \in \Ft$ and a hypothesis $\h \in \Hs$ in the common domain. The learning objective minimizes a loss function composed of a weighted (convex) combination of the source and target classification losses, along with a distribution distance term that measures the discrepancy between the aligned domains. At this stage, our analysis remains general and does not assume any specific structure for the learning algorithm. In Section \ref{ssec:gen_bnd_arb_dist} (Theorem \ref{thm:gen_defect_target}), we present a probabilistic bound on the expected target loss in terms of the empirical weighted loss and the expected distribution discrepancy.  

\item  In Section \ref{ssec_gen_bnd_mmd} we develop these results for the setting where the distribution distance is selected as the popular maximum mean discrepancy (MMD) metric. In Theorem \ref{thm:main_result_mmd}, we show that  the expected target loss can be effectively bounded in terms of the empirical classification and distribution losses alone. This bound holds provided that the number of labeled source samples $\Ms$  scales logarithmically with the covering number of the composite hypothesis class $\Hs \circ \Fs$, while the total number of source and target samples, $\Ns$ and $\Nt$, must scale logarithmically with the covering numbers of the feature transformation classes $\Fs$ and $\Ft$. 

\item In Sections \ref{ssec_samp_comp_mmd_net}-\ref{ssec_adv_da_net} we extend our analysis to domain-adaptive deep learning algorithms and, in particular, investigate their sample complexity. We consider two pioneering approaches that have inspired a large body of follow-up work: MMD-based domain adaptation networks \cite{LongCWJ15}, \cite{TzengHZSD14}, \cite{GhifaryKZ14} and adversarial domain adaptation networks \cite{GaninUAGLLML16}, \cite{TzengHSD17}, \cite{TangJ20}. Our results in Theorems  \ref{thm_main_result_da_mmd} and \ref{thm_main_result_dann} show that, in both MMD-based and adversarial domain adaptation settings, the sample complexities for the number of labeled source samples $\Ms$ and the total number of source and target samples, $\Ns$ and  $\Nt$, scale quadratically with the width $\dcom$ and the depth $\numL$ of the network. Our results also offer insight into the optimal choice for the weight $\alpha$ of the target classification loss, indicating it should decrease at rate $\alpha=O(\sqrt{\Mt})$ to effectively handle the scarcity of labeled target samples.  Our proof technique extends Theorem \ref{thm:main_result_mmd} by thoroughly analyzing the covering numbers of the relevant function classes.  To the best of our knowledge, these are the first results to provide a comprehensive characterization of the sample complexity of domain-adaptive neural networks.

\end{itemize}

We defer a detailed discussion of closely related literature to Section \ref{sec_rel_work}, where we also compare and contrast our results with previous findings.  Section \ref{sec:exp_results} presents some simulation results for the experimental validation of our findings, and Section \ref{sec:conclusion} concludes the paper. A preliminary version of our study was presented in \cite{Vural18}, which laid the groundwork for the results in Section \ref{ssec:gen_bnd_arb_dist}.

\section{General performance bounds for domain alignment}
\label{sec:gen_bounds}

\subsection{Problem formulation}
\label{ssec:problem_form}

Let $\Xs$ and $\Xt$ denote two compact metric spaces representing respectively a source domain and a target domain, and let $\Y \subset \R^m$ be a label set. Let $\mus$ be a source Borel probability measure and $\mut$ be a target Borel probability measure respectively on the sets $\Zs = \Xs \times \Y$ and $\Zt = \Xt \times \Y$. We consider the family of learning algorithms that aim to learn two mappings (transformations) $\fs: \Xs \rightarrow \X$ and $\ft: \Xt \rightarrow \X$ from the source and target domains to a common set $\X$ together with a hypothesis function $h: \X \rightarrow \Y $ estimating class labels on $\X$. The expected losses of the transformations $\fs$, $\ft$, and the hypothesis $h$ at the source and target are respectively given by
\begin{equation*}
\begin{split}
\Ls( \fs, \h) &= \int_{\Zs} \loss( \h \circ \fs(\xs), \ys ) \, d \mus  \\
\Lt( \ft, \h) &= \int_{\Zt} \loss( \h \circ \ft(\xt), \yt ) \, d \mut 
\end{split}
\end{equation*}
where $\loss: \Y \times \Y \rightarrow [0, \infty)$ is a loss function.  Assuming that $\fs$ and $\ft$ are measurable mappings, the probability measures $\mus$ and $\mut$ on the source and target domains induce corresponding probability measures $\nus$ and $\nut$ on the domain $\X$. Let $\D$ be a function such that $\D (\fs, \ft)$ represents the distance between the measures $\nus$ and $\nut$ on $\X$ induced via the mappings $\fs$ and $\ft$ with respect to some distribution discrepancy criterion.

Let $\{ \xis \}_{i=1}^{\Ns}$ be a set of source samples and $\{ \xjt \}_{j=1}^{\Nt}$ be a set of target samples drawn independently from the probability measures $\mus$ and $\mut$, where $\{ \xis \}_{i=1}^{\Ms}$ are the $\Ms$ labeled samples in the source with labels $\{ \yis \}_{i=1}^{\Ms}$, and $\{ \xjt \}_{j=1}^{\Mt}$ are the $\Mt$ labeled samples in the target with labels $\{ \yjt \}_{j=1}^{\Mt}$. We consider learning algorithms that minimize a convex combination of the source and target empirical losses, while minimizing the distance between the transformed source and target samples in the domain $\X$ as

\begin{equation}
\label{eq:obj_learning}
\begin{split}
\min_  {\fs \in \Fs,  \  \ft \in \Ft, \ \h \in  \Hs}
(1-\alpha) \hLs (\fs, \h) + \alpha \hLt (\ft, \h) 
+ \beta \hD (\fs, \ft).
\end{split}
\end{equation}

Here $\Fs$ and $\Ft$ are function classes consisting of a family of transformations, respectively from the source and target domains $\Xs$ and $\Xt$ to $\X$; $\Hs$ is a hypothesis class consisting of hypotheses; $\alpha$ is a weight parameter with $0 \leq \alpha \leq 1$; $\hLs (\fs, \h)$ and $\hLt (\ft, \h) $ are the empirical source and target losses given by
\begin{equation}
\label{eq_emp_cl_loss}
\begin{split}
\hLs( \fs, \h) &= \frac{1}{\Ms} \sum_{i=1}^{\Ms}  \loss( \h \circ \fs(\xis), \yis )  \\
\hLt( \ft, \h) &=  \frac{1}{\Mt} \sum_{j=1}^{\Mt} \loss( \h \circ \ft(\xjt),  \yjt  ) 
\end{split}
\end{equation}
and the distance $\hD$ is an estimate of the distribution distance $\D (\fs, \ft)$ computed with all (labeled and unlabeled) samples $\{ \xis \}_{i=1}^{\Ns}$ and $\{ \xjt \}_{j=1}^{\Nt}$. As discussed in Section \ref{sec:intro}, the distribution distance $\D(\fs, \ft)$ has been chosen in different ways in previous works such as the MMD or Wasserstein distance along with the corresponding estimates $\hD(\fs, \ft)$ that lead to practical learning algorithms. In Section \ref{ssec:gen_bnd_arb_dist}, we provide generalization bounds for  learning algorithms with an arbitrary distribution distance function. Then in Section \ref{ssec_gen_bnd_mmd}, we focus on the kernel mean matching (KMM) methods in particular, and propose bounds for algorithms using a KMM-based distribution distance.

\subsection{Generalization bounds for arbitrary distribution distances}
\label{ssec:gen_bnd_arb_dist}

\begin{figure}[t]
\begin{center}
     \subfigure[]
       {\label{fig:illus_domain_relation_a}\includegraphics[height=3.0cm]{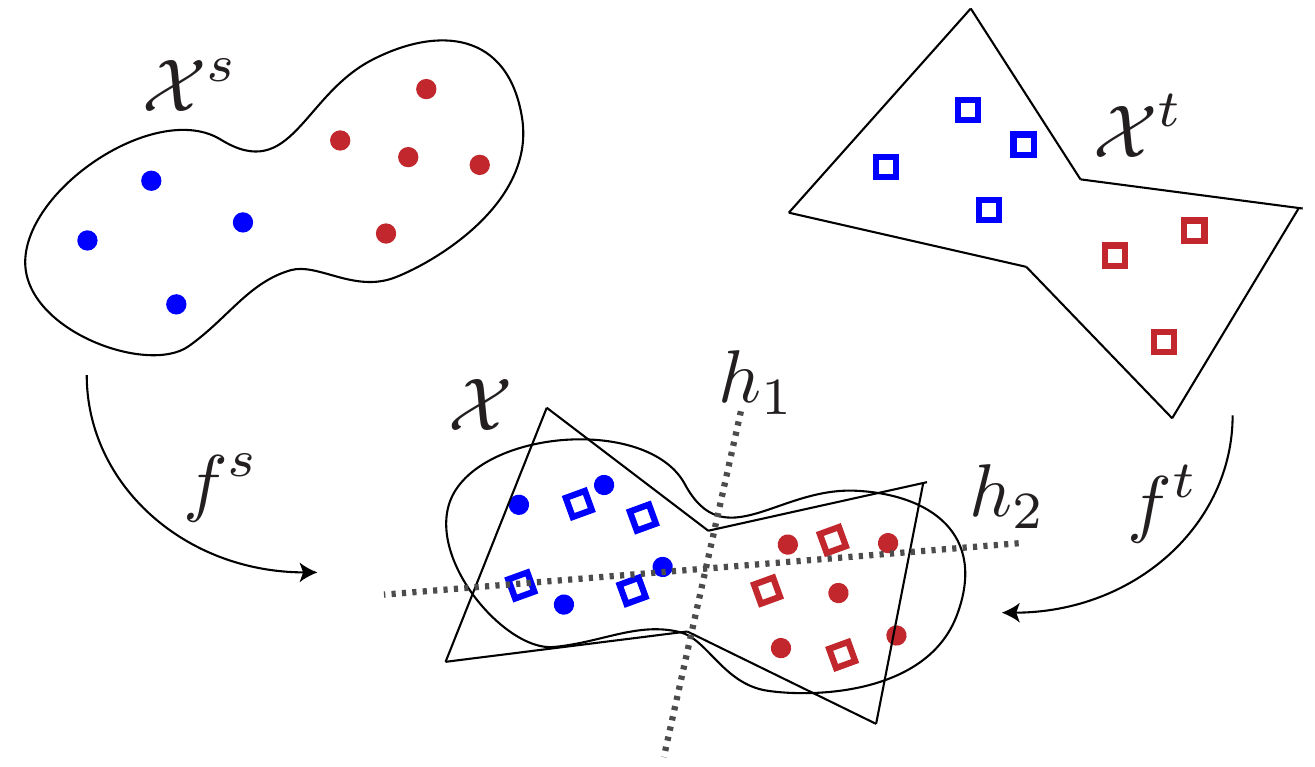}}
       \hspace{1cm}
     \subfigure[]
       {\label{fig:illus_domain_relation_b}\includegraphics[height=3.2cm]{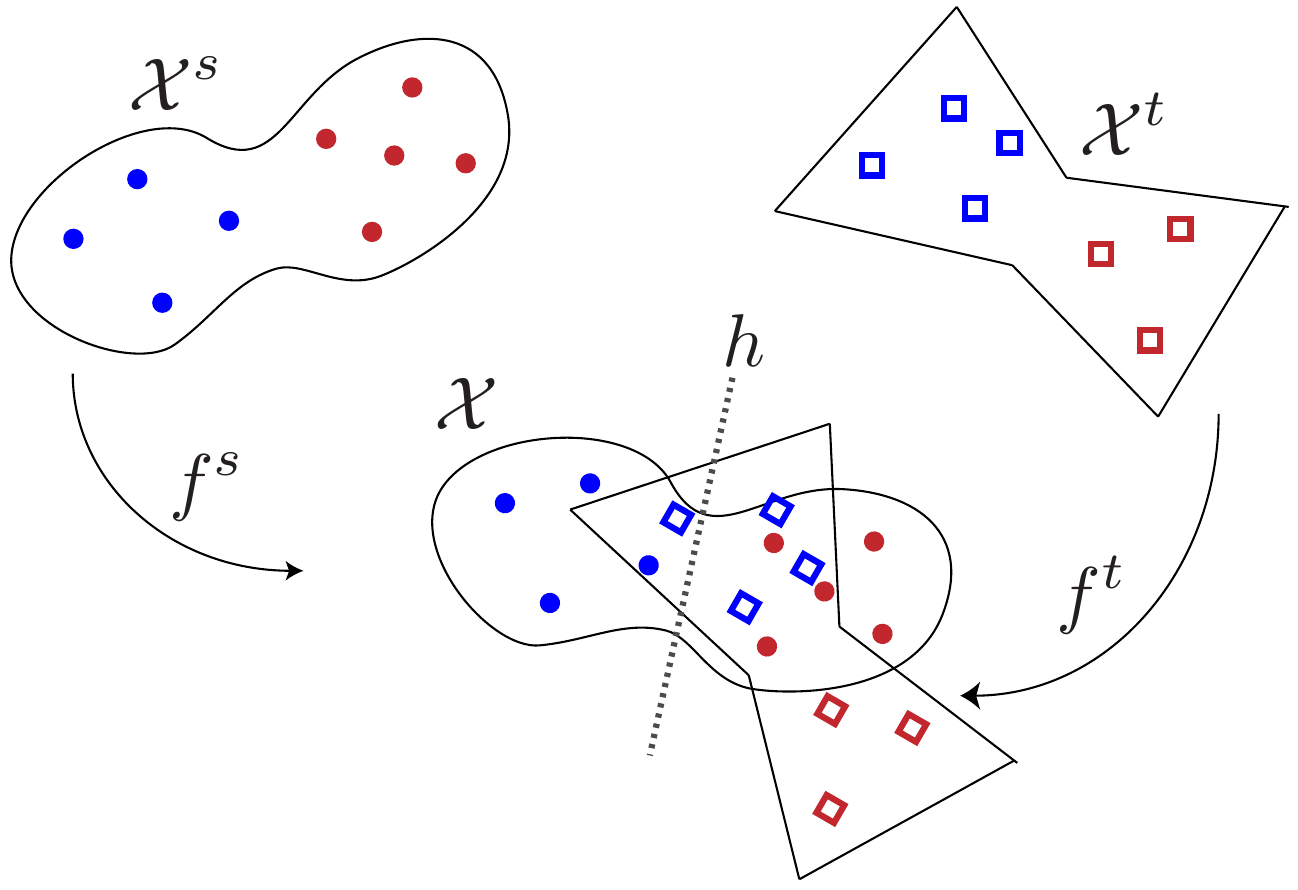}}
 \end{center}
 \vspace{-0.3cm}
 \caption{Illustration of Assumption \ref{assum_existence_LLs}. Red and blue colors represent two different classes in the source and target domains $\Xs$ and $\Xt$. In (a), the two domains are well-aligned by the learnt transformations; therefore, the source and target losses are similar. In (b), the learnt transformations do not align the domains well; therefore, the difference between the source and target losses can be high.}
 \label{fig:illus_domain_relation}
\end{figure}

In order to analyze the performance of algorithms that aim to solve  \eqref{eq:obj_learning}, we first assume that the expected loss has a bounded rate of variation with respect to the chosen distribution distance: 

\begin{assumption}
\label{assum_existence_LLs}
There exists a constant $\LLs>0$ such that, for any transformations $\fs \in \Fs$, $\ft \in \Ft$ and any hypothesis $\h \in \Hs$, we have
\begin{equation}
\label{eq:Lipsc_reg_loss}
| \Ls(\fs, \h) - \Lt(\ft, \h) | \leq \LLs \ \D(\fs, \ft). 
\end{equation}
\end{assumption}

Assumption \ref{assum_existence_LLs} imposes the presence of a relation between the source and target distributions: The source and target distributions must be ``related'' in such a way that, when their distance is reduced in the common domain after going through the transformations in $\Fs$, $\Ft$,  their resulting losses should not differ too much compared to the distribution distance $D(\fs, \ft)$. This assumption is illustrated in Figure \ref{fig:illus_domain_relation}. The figure depicts a simple setting where the source and target domains are aligned by geometric transformations $\fs$, $\ft$, which are respectively in the geometric transformation families $\Fs$ and $\Ft$. The hypothesis family $\Hs$ consists of linear classifiers $\h$.  In Figure \ref{fig:illus_domain_relation_a}, the learnt transformations $\fs$ and $\ft$  suitably align the two domains, so that the distribution distance $\D(\fs, \ft)$ is small. Consequently, a hypothesis $h_1$ that yields a small loss $ \Ls(\fs, \h_1) $ in the source domain also yields a small loss $\Lt(\ft, \h_1)$ in the target domain; and a hypothesis $h_2$ that yields a large loss $ \Ls(\fs, \h_2) $ in the source domain also yields a large loss $\Lt(\ft, \h_2)$ in the target domain. Meanwhile, in Figure \ref{fig:illus_domain_relation_b} the learnt transformations $\fs$ and $\ft$ do not align the two domains well. In this case, the distribution distance $\D(\fs, \ft)$ is large, which allows the loss difference $| \Ls(\fs, \h) - \Lt(\ft, \h) |$ also to be large by Assumption \ref{assum_existence_LLs}. Indeed, one may find a hypothesis $h$ that yields a small loss  $ \Ls(\fs, \h) $ in the source domain, but a large loss  $\Lt(\ft, \h)$ in the target domain. Since the loss difference $| \Ls(\fs, \h) - \Lt(\ft, \h) |$ can be bounded in terms of the distribution distance $\D(\fs, \ft)$, the transformation families  $\Fs, \Ft$, and the hypothesis family $\Hs$ considered in this example satisfy Assumption \ref{assum_existence_LLs}. In brief, the assumption dictates that there should be a sufficiently strong relation between the source and target domains, the function classes $\Fs$ and $\Ft$ must be chosen suitably to respect this relation, and the hypothesis family $\Hs$ must also be compatible with the problem.

In the following, we first bound the expected target loss in terms of the expected weighted loss and the distribution distance.

\begin{lemma}
\label{lem:dev_exptar_expweigh}
Consider that Assumption \ref{assum_existence_LLs} holds. Let $\Lw(\fs, \ft, \h)$ denote the expected weighted loss in the source and target domains given by
\[
\Lw(\fs, \ft, \h) \triangleq  (1-\alpha) \Ls(\fs, \h) + \alpha \Lt(\ft, \h).
\]
Then the expected target loss is bounded as
\[\Lt(\ft, \h) \leq   \Lw(\fs, \ft, \h)+ (1-\alpha) \LLs  \D(\fs, \ft).\]
\end{lemma}

\begin{proof}
We have
$ 
\Lt(\ft, \h) = \alpha \Lt(\ft, \h) +  (1-\alpha) \Lt(\ft, \h).
$
From Assumption \ref{assum_existence_LLs}, we get
\[
 \Lt(\ft, \h) \leq \Ls (\fs, \h) + \LLs \ \D(\fs, \ft).
\]
Using this above, we obtain
\begin{equation*}
\begin{split}
\Lt(\ft, \h) &\leq \alpha \Lt(\ft, \h) 
+  (1-\alpha) \left( \Ls (\fs, \h) + \LLs \ \D(\fs, \ft) \right)\\
&= \Lw(\fs, \ft, \h)+ (1-\alpha) \LLs  \D(\fs, \ft).
\end{split}
\end{equation*}
\end{proof}

We use the above relation to bound the expected target loss in terms of the empirical losses given by the learning algorithm. We characterize the complexity of the transformation and hypothesis classes in terms of their covering numbers, defined as follows \cite{CuckerS02}:%

\begin{definition}
Let $\F$ be a compact metric space with metric $\dmetric $, and let $B_\epsilon(f)$ denote an open ball of radius $\epsilon$ around $f \in \F$. Then the covering number $\N(\F, \epsilon, \dmetric)$ of $\F$ is defined as
\[
\N(\F, \epsilon, \dmetric) \triangleq \min \{ k: \exists f_1, \dots f_k \in \F, \ \F \subset \cup_{i=1}^k B_\epsilon(f_i) \}.
\]
\end{definition}

In order to study the discrepancy between the expected and the empirical losses, we next make the following assumptions. 
\begin{assumption}
\label{assum_HF_comp_Ll_Al}
The composite function classes $\Hs \circ \Fs \triangleq \{ \gs =  \h \circ \fs : \h \in \Hs, \fs \in \Fs \}$ and $\Hs \circ \Ft \triangleq \{ \gt = \h \circ \ft : \h \in \Hs, \ft \in \Ft \}$ are compact metric spaces with respect to the metrics
\begin{equation}
\label{eq_defn_ds_dt}
\begin{split}
\ds(\gs_1, \gs_2) & \triangleq \sup_{\xs \in \Xs} \| \gs_1(\xs) - \gs_2(\xs)  \|  \\
\dt(\gt_1, \gt_2) & \triangleq \sup_{\xt \in \Xt} \| \gt_1(\xt) - \gt_2(\xt)  \|
\end{split}
\end{equation}
where $\| \cdot \|$ denotes the $l_2$-norm in $\R^m$.
Also, the loss function $\loss$ is bounded by $\bls$ and Lipschitz continuous with respect to the first argument with constant $\Lls$, such that
\begin{equation*}
\begin{split}
\loss(\y_1, \y_2) &\leq \bls, \  \forall \y_1, \y_2 \in \Y  \\
| \loss(\y_1, \y) -  \loss(\y_2, \y) | &\leq \Lls \| \y_1 - \y_2 \|,\  \forall \y_1, \y_2, \y \in \Y  .
\end{split}
\end{equation*}
\end{assumption}

We can now present the following result that bounds the deviation between the expected and  empirical weighted losses.

\begin{lemma}
\label{lem:weight_loss_gen}
Let the conditions in Assumption \ref{assum_HF_comp_Ll_Al} hold. Let
\[
\hLw(\fs, \ft, \h) \triangleq  (1-\alpha) \hLs(\fs, \h) + \alpha \hLt(\ft, \h)
\]
denote the empirical weighted loss. Then, we have
\begin{equation*}
\begin{split}
&P\left (\sup_{\fs \in \Fs, \ft \in \Ft, \h \in \Hs} |  \Lw( \fs, \ft, \h ) - \hLw( \fs,  \ft, \h )   |  \leq \epsilon \right) \\
& \geq 1 - 2 \N( \Hs \circ \Ft, \frac{\epsilon}{8 \alpha \Lls}, \dt) e^{-\frac{\Mt \epsilon^2}{8 \alpha^2 \bls^2}} 
-2 \N( \Hs \circ \Fs, \frac{\epsilon}{8 (1-\alpha) \Lls}, \ds) e^{-\frac{\Ms \epsilon^2}{8 (1-\alpha)^2 \bls^2}}.
\end{split}
\end{equation*}
\end{lemma}

The proof of Lemma \ref{lem:weight_loss_gen} is given in Appendix \ref{pf_lem_weight_loss_gen}.

We can now simply combine Lemmas \ref{lem:dev_exptar_expweigh} and \ref{lem:weight_loss_gen}  to bound the expected target loss in terms of the empirical weighted loss and the distribution distance in the following main result.
\begin{theorem}
\label{thm:gen_defect_target}
Let Assumptions \ref{assum_existence_LLs}, \ref{assum_HF_comp_Ll_Al} hold. Then for any transformations $\fs \in \Fs$, $\ft \in \Ft$ and hypothesis $\h \in \Hs$, with probability at least
\begin{equation}
\label{eq_prob_expr_thm1}
\begin{split}
1 &- 2 \N( \Hs \circ \Ft, \frac{\epsilon}{8 \alpha \Lls}, \dt) e^{-\frac{\Mt \epsilon^2}{8 \alpha^2 \bls^2}} 
-2 \N( \Hs \circ \Fs, \frac{\epsilon}{8 (1-\alpha) \Lls}, \ds) e^{-\frac{\Ms \epsilon^2}{8 (1-\alpha)^2 \bls^2}}
\end{split}
\end{equation}
the expected target loss is bounded as
\[
\Lt(\ft, \h) \leq   \hLw(\fs, \ft, \h)+ (1-\alpha) \LLs  \D(\fs, \ft) + \epsilon.
\]
\end{theorem}
The main result in Theorem \ref{thm:gen_defect_target} states the following: For any algorithm that computes transformations $\fs$, $\ft$, and a hypothesis $\h$ by attempting to solve a problem such as in \eqref{eq:obj_learning}, the actual expected loss obtained at the target  by applying the learnt transformation $\ft$ and hypothesis $\h$ to target test samples cannot differ from the empirical weighted loss  $\hLw(\fs, \ft, \h)$ obtained over training samples by more than $\epsilon$ plus an error term involving the distance $ \D(\fs, \ft) $. This statement holds with probability approaching 1 at an exponential rate with the increase in number of labeled samples $\Ms$. Note that in the very typical case where $\Mt$ is limited, the target term in the probability expression \eqref{eq_prob_expr_thm1} can be controlled by suitably scaling down the weight parameter $\alpha$ proportionally to $O(\sqrt{\Mt})$. \\

\begin{remark}
An important question is how much the learning algorithm is expected to reduce the distribution distance $\D(\fs, \ft)$. This depends on the chosen distance; nevertheless, in many practical learning problems, the number of unlabeled samples $\Ns, \Nt$ is much larger than the number of labeled samples $\Ms, \Mt$. If we assume that $N = \min( \Ns, \Nt) $ is sufficiently large, then we may expect the deviation between the expected and empirical distribution distances to decay such that
\begin{equation*}
\begin{split}
P(  | \D(\fs, \ft) - \hD(\fs, \ft) |  \geq \epsilon ) 
&\leq (\N_{\Fs, \epsilon} + \N_{\Ft, \epsilon}) \ O \left( e^{-  N \epsilon^2 } \right)  \\
&\leq O \left( e^{-  \Mt \epsilon^2 } \right) 
+ O \left( e^{-  \Ms \epsilon^2 } \right) 
\end{split}
\end{equation*}
for some appropriate complexity measures $\N_{\Fs, \epsilon}$ , $\N_{\Ft, \epsilon}$ for the transformation function classes. In this case, the result in Theorem \ref{thm:gen_defect_target} would imply that with  probability $1- O( e^{-  \Mt \epsilon^2 } ) - O( e^{-  \Ms \epsilon^2 } ) $, the expected target loss would be bounded in terms of the empirical losses and the empirical distribution distance as
\begin{equation}
\begin{split}
\Lt(\ft, \h) &\leq   \hLw(\fs, \ft, \h) 
+ (1-\alpha) \LLs  \hD(\fs, \ft) + \epsilon + (1-\alpha) \LLs \epsilon.
\end{split}
\end{equation}
Our purpose in the next section is to establish such a result for the particular setting where the distribution distance is chosen as the MMD.
\end{remark}

\subsection{Generalization bounds for maximum mean discrepancy measures}
\label{ssec_gen_bnd_mmd}

We now extend the results of Section \ref{ssec:gen_bnd_arb_dist} for a setting where the distribution discrepancy in the common domain of transformation is measured with respect to the maximum mean discrepancy (MMD) criterion. The MMD criterion is widely used in domain adaptation. In particular, a popular family of methods called kernel mean mathcing (KMM) algorithms aim to map the source and target data to a shared domain via a kernel function such that the distance between the source and target samples measured with respect to the MMD criterion is minimized.

KMM methods set the source and target mappings $\fs: \Xs \rightarrow \X$ and $\ft: \Xt \rightarrow \X$ as a kernel-induced feature map $\phi$. The source and target domains $\Xs = \Xt$ are often assumed to be the same and the transformations are set as $\fs=\ft=\phi$. The shared domain $\X$ is typically a Hilbert space with a kernel $\kr: \Xs \times \Xt \rightarrow \R$ satisfying $\kr(\xs, \xt) = \langle \phi(\xs), \phi(\xt) \rangle_{\X}$ with respect to the inner product $\langle \cdot, \cdot \rangle_{\X}$ in $\X$.

Given the source and target probability measures $\mus$, $\mut$ on the sets $\Zs = \Xs \times \Y$ and $\Zt = \Xt \times \Y$; and the probability measures $\nus$, $\nut$ these respectively induce over the domain $\X$; KMM algorithms characterize the distance between $\nus$ and $\nut$ via the MMD given by 
\begin{equation}
\label{eq:defn_D_MMD}
\D(\fs, \ft) = \|  E_{\xs}[\fs(\xs)] - E_{\xt}[ \ft(\xt) ] \|_{\X}
\end{equation}
where $\| \cdot \|_{\X}$ stands for the inner-product-induced norm in the Hilbert space $\X$. For notational simplicity, we will drop the subscript $(\cdot)_{\X}$ when there is no ambiguity over the space in consideration.  The notation $ E_{\xs}[\cdot] $ and $ E_{\xt}[\cdot] $ indicates that the expectations are taken with respect to the probability measures $\mus$ and $\mut$ in the source and the target domains, respectively. We will simply write $E[\cdot]$ whenever the meaning is clear. Given the source and target sample sets $\{ \xis \}_{i=1}^{\Ns} $ and $\{ \xjt \}_{j=1}^{\Nt}$, the empirical estimate of the MMD is given by
\begin{equation}
\label{eq:defn_hD_MMD}
\hD(\fs, \ft) = \left \| 
\frac{1}{\Ns} \sum_{i=1}^{\Ns} \fs(\xis) -  \frac{1}{\Nt} \sum_{j=1}^{\Nt} \ft(\xjt) 
\right \|.
\end{equation}

\begin{remark}
 Although most KMM methods assume the source and target domains to be the same ($\Xs = \Xt$), and also the source and target transformations to be the same ($\fs=\ft=\phi$), we do not make use of these assumptions in the analysis presented in this section. Here, we only assume that the distribution discrepancy between $\nus$ and $\nut$ is taken as in \eqref{eq:defn_D_MMD} for any two transformations $\fs \in \Fs$ and $\ft \in \Ft$, and the empirical estimate of the MMD is computed as in \eqref{eq:defn_hD_MMD}.
\end{remark}

In order to study the performance of KMM algorithms,  we would like to first derive a bound on the deviation between the actual distribution discrepancy $\D(\fs, \ft)$ and its empirical estimate $\hD(\fs, \ft)$. We make the following assumption on the data distributions:

\begin{assumption}
\label{assum_fx_bdd_moments}
The expected deviations of the random variables $\{ \fs(\xis) \}_{i=1}^{\Ns}$ and  $\{ \ft(\xjt) \}_{j=1}^{\Nt}$ from their means $E[\fs(\xs)]$ and $E[\ft(\xt)]$ are bounded such that there exist constants $\vars$ and $\vart$ satisfying
\begin{equation}
\label{eq:var_bnd_fs_ft}
\begin{split}
E \left[  \| \fs(\xis) - E[\fs(\xs)]  \|^2 \right]  &\leq \vars \\
E \left[  \| \ft(\xjt) - E[\ft(\xt)]  \|^2 \right]  &\leq \vart.
\end{split}
\end{equation}
Also, for the higher order powers of the deviation, there exist constants $\Cs$ and $\Ct$ satisfying
\begin{equation}
\label{eq:mom_bnd_fs_ft}
\begin{split}
E \left[  \| \fs(\xis) - E[\fs(\xs)]  \|^k \right]  &\leq \frac{k!}{2} \vars \, \Cs^{k-2} \\
E \left[  \| \ft(\xjt) - E[\ft(\xt)]  \|^k \right]  &\leq \frac{k!}{2} \vart \, \Ct^{k-2}.
\end{split}
\end{equation}
\end{assumption}
The condition \eqref{eq:var_bnd_fs_ft} can be seen as a finite variance assumption for a distribution over a Hilbert space, and the condition \eqref{eq:mom_bnd_fs_ft} bounds the growth of the $k$-th central moment by a rate of $O(k! \, C^{k})$. These assumptions hold for many common data distributions in practice.

We first present the following lemma, which bounds the deviation between the expectation and the empirical mean of the source and the target data mapped to the common domain $\X$ via the transformations $\fs$ and $\ft$.

\begin{lemma}
\label{lem:bernstein_src_trg}
Let the source and target distributions and the transformations $\fs: \Xs \rightarrow \X$ and $\ft: \Xt \rightarrow \X$ be such that Assumption \ref{assum_fx_bdd_moments} holds. Also, for given $\epsilon>0$, let the number of source and target samples be such that
\begin{equation*}
\Ns > \frac{\vars}{\epsilon^2}
, \quad
\Nt > \frac{\vart}{\epsilon^2}.
\end{equation*}
Then for the source domain we have
\begin{equation}
\label{eq:lem_src_empmean}
\begin{split}
&P\left( \left \|  \frac{1}{\Ns}  \sum_{i=1}^{\Ns}  \fs(\xis)  -  E[\fs(\xs)] \right \|  \geq  \epsilon \right) \\
&\leq
  \exp \left(  
  -\frac{1}{8}\left( \frac{\sqrt{\Ns} \epsilon}{\sigma_s} -1\right)^2 \frac{1}{1+\left( \frac{\sqrt{\Ns} \epsilon}{\sigma_s} -1 \right)\frac{\Cs}{2 \sqrt{\Ns} \sigma_s }}
   \right) 
   \end{split}
\end{equation} 
and for the target domain we have
\begin{equation}
\label{eq:lem_trg_empmean}
\begin{split}
&P\left( \left \|  \frac{1}{\Nt}  \sum_{j=1}^{\Nt}  \ft(\xjt)  -  E[\ft(\xt)] \right \|  \geq  \epsilon \right) \\
&\leq
  \exp \left(  
  -\frac{1}{8}\left( \frac{\sqrt{\Nt} \epsilon}{\sigma_t} -1\right)^2 \frac{1}{1+\left( \frac{\sqrt{\Nt} \epsilon}{\sigma_t} -1 \right)\frac{\Ct}{2 \sqrt{\Nt} \sigma_t }}
   \right) .
\end{split}
\end{equation}
\end{lemma}

The proof of Lemma \ref{lem:bernstein_src_trg} is given in Appendix \ref{pf_lem_bernstein_src_trg}. Lemma \ref{lem:bernstein_src_trg} provides a bound on the deviation between the sample mean and the expectation of the source and target samples transformed to the shared Hilbert space $\X$. In particular, it states that as the number $\Ns, \Nt$ of source and target samples increases, this deviation can be upper bounded with probability improving at an exponential rate with $\Ns$ and $\Nt$. We next build on this result to present in Lemma \ref{lem:unif_bnd_D_hD} a uniform upper bound on the deviation $| \D(\fs, \ft) - \hD (\fs, \ft)  |$ between the expected and empirical MMD distances, which is valid for any $\fs \in \Fs$ and $\ft \in \Ft$. We first need an assumption on the compactness of the function classes $\Fs$ and $\Ft$:

\begin{assumption}
\label{assum_Fs_Ft_compact}
The function classes $\Fs$ and $\Ft$ are compact metric spaces with respect to the metrics
\begin{equation}
\label{eq_defn_dXs_dXt}
\begin{split}
\dXs (\fs_1 , \fs_2) & \triangleq \sup_{\xs \in \Xs}  \|  \fs_1(\xs) - \fs_2(\xs) \| \\
\dXt (\ft_1 , \ft_2) & \triangleq \sup_{\xt \in \Xt}  \|  \ft_1(\xt) - \ft_2(\xt) \| .
\end{split}
\end{equation}

\end{assumption}

\begin{lemma}
\label{lem:unif_bnd_D_hD}

Let Assumptions \ref{assum_fx_bdd_moments}, \ref{assum_Fs_Ft_compact} hold. Given $\epsilon>0$, let the number of source and target samples be such that
\begin{equation*}
\Ns > \frac{16 \vars}{\epsilon^2},
\quad
\Nt > \frac{16 \vart}{\epsilon^2}.
\end{equation*}
Let us define the functions
\begin{equation*}
\begin{split}
\as(\Ns, \epsilon) &\triangleq 
  \frac{1}{8}\left( \frac{\sqrt{\Ns} \epsilon}{4 \sigma_s} -1\right)^2 \frac{1}{1+\left( \frac{\sqrt{\Ns} \epsilon}{4 \sigma_s} -1 \right)\frac{\Cs}{2 \sqrt{\Ns} \sigma_s }}
 \\
\at(\Nt, \epsilon) &\triangleq  
 \frac{1}{8}\left( \frac{\sqrt{\Nt} \epsilon}{4 \sigma_t} -1\right)^2 \frac{1}{1+\left( \frac{\sqrt{\Nt} \epsilon} 
 {4 \sigma_t} -1 \right)\frac{\Ct}{2 \sqrt{\Nt} \sigma_t }}.
\end{split}
\end{equation*}
Then 
\begin{equation*}
\begin{split}
& P \left(
\sup_{\fs \in \Fs, \ft \in \Ft}  | \D(\fs, \ft) - \hD (\fs, \ft)  | < \epsilon
\right) \\
& \geq 1 - \N(\Fs, \frac{\epsilon}{8}, \dXs) \exp(-\as(\Ns, \epsilon)) 
- \N(\Ft, \frac{\epsilon}{8}, \dXt) \exp(-\at(\Nt, \epsilon)).
\end{split}
\end{equation*}
\end{lemma}

Lemma \ref{lem:unif_bnd_D_hD} is proved in Appendix \ref{pf_lem_unif_bnd_D_hD}. The lemma provides a probabilistic upper bound on the deviation between the actual MMD and its estimate from a finite sample set, which holds for all functions in the transformation function classes $\Fs$ and $\Ft$.  We are now ready to combine this bound with our results in Section \ref{ssec:gen_bnd_arb_dist}. We recall that in Theorem \ref{thm:gen_defect_target}, the expected target loss $\Lt(\ft, \h)$ was bounded in terms of the empirical weighted loss $\Lw(\fs, \ft, \h)$ and the true distribution discrepancy $\D(\fs, \ft)$ after the transformations. However, in practice, for two transformations $\fs$, $\ft$ computed by a domain adaptation method, the true distribution discrepancy $\D(\fs, \ft)$ is often unknown. We are now in a position to extend Theorem \ref{thm:gen_defect_target} in the following result, where we bound the expected target loss in terms of the empirical MMD measure $\hD( \fs, \ft)$.

\begin{theorem}
\label{thm:main_result_mmd}

Consider a domain adaptation algorithm where the distribution discrepancy is taken as the MMD measure, and the loss function and data distributions satisfy Assumptions \ref{assum_existence_LLs}-\ref{assum_Fs_Ft_compact}. For $\epsilon >0$, let the number of source and target samples satisfy
\begin{equation*}
\Ns > \frac{16 \vars}{\epsilon^2},
\quad \quad
\Nt > \frac{16 \vart}{\epsilon^2}.
\end{equation*}
Then for any transformations $\fs \in \Fs$, $\ft \in \Ft$, and hypothesis $\h \in \Hs$, with probability at least
\begin{equation*}
\begin{split}
1 &- 2 \N( \Hs \circ \Ft, \frac{\epsilon}{8 \alpha \Lls}, \dt) e^{-\frac{\Mt \epsilon^2}{8 \alpha^2 \bls^2}} 
-2 \N( \Hs \circ \Fs, \frac{\epsilon}{8 (1-\alpha) \Lls}, \ds) e^{-\frac{\Ms \epsilon^2}{8 (1-\alpha)^2 \bls^2}} \\
& - \N(\Fs, \frac{\epsilon}{8}, \dXs) \exp(-\as(\Ns, \epsilon)) 
- \N(\Ft, \frac{\epsilon}{8}, \dXt) \exp(-\at(\Nt, \epsilon))
\end{split}
\end{equation*}
the expected target loss is upper bounded as
\begin{equation*}
\begin{split}
\Lt(\ft, \h) \leq   \hLw(\fs, \ft, \h)+ (1-\alpha) \LLs  \hD(\fs, \ft) + (1-\alpha) \LLs \epsilon + \epsilon.
\end{split}
\end{equation*}
\end{theorem}

\begin{proof}
The stated result follows simply from Theorem \ref{thm:gen_defect_target} and Lemma \ref{lem:unif_bnd_D_hD} by applying the union bound.
\end{proof}

The result in Theorem \ref{thm:main_result_mmd} states that the target loss can be bounded in terms of the empirical weighted loss and the empirical distribution discrepancy, with probability approaching 1 at an exponential rate as the number of labeled and unlabeled samples increases. The dependence of this rate on the number of unlabeled samples follows from the relations $\as(\Ns,\epsilon)=O(\Ns \epsilon^2)$ and $\at(\Nt, \epsilon) = O(\Nt \epsilon^2)$. In particular, our result points to the following practical fact: If a domain adaptation algorithm efficiently minimizes the empirical weighted loss and the empirical distribution discrepancy, the true loss obtained in the target domain will also be small, provided that the number of samples is sufficiently high with respect to the complexity of the transformation and hypothesis classes, characterized by their covering numbers.\\

\section{Sample complexity of domain-adaptive neural networks}
\label{sec_samp_dan}

In this section, we build on the results in Section \ref{sec:gen_bounds} and extend our analysis to examine the performance of domain-adaptive neural networks. In particular, we study the sample complexity of two common neural network types, namely, MMD-based and adversarial architectures, respectively in Section \ref{ssec_samp_comp_mmd_net} and Section \ref{ssec_adv_da_net}.

\subsection{MMD-based domain adaptation networks}
\label{ssec_samp_comp_mmd_net}

\begin{figure}[t]
  \centering
  \centerline{\includegraphics[width=14.0cm]{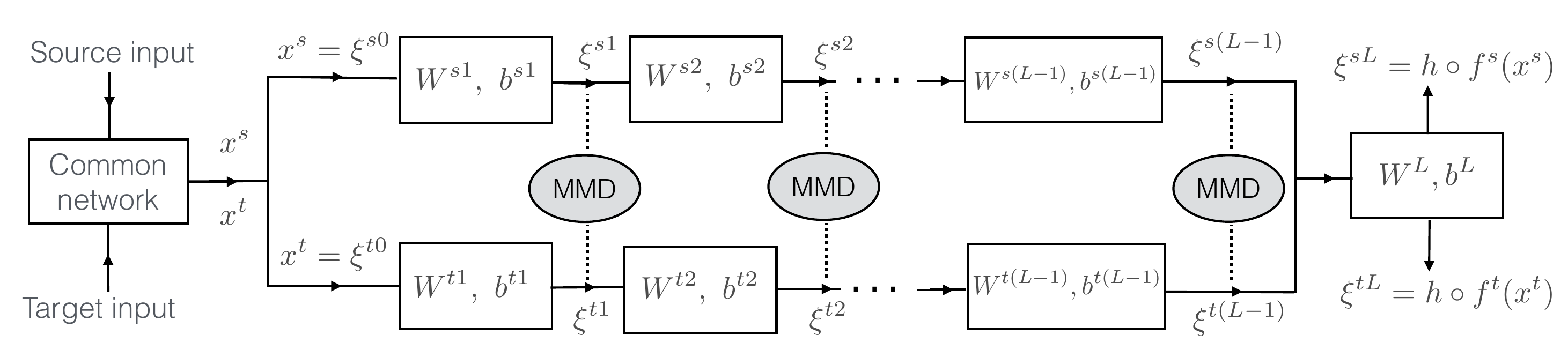}}
  \caption{Illustration of MMD-based domain adaptation networks}
  \label{fig_illus_mmd_network}
\end{figure}

We begin with studying the implications of Theorem \ref{thm:main_result_mmd} on deep domain adaptation networks that learn domain-invariant features based on the MMD distance measure. We consider the network model depicted in Figure \ref{fig_illus_mmd_network}, which serves as a commonly adopted foundation for many MMD-based neural network architectures.  The source and target samples first pass through a common network, possibly comprising multiple convolutional and fully connected layers. The common network output is then provided to a source network and a target network consisting of $\numL-1$ fully connected layers in the corresponding domain, with the $\numL$-th (output) layer consisting of a classifier that is shared between the two domains. The action of the common network remains out of the scope of our study, as its parameters are often adopted from a pre-trained network or fine-tuned using only a set of source samples in the literature \cite{LongCWJ15}, \cite{TzengHZSD14}, \cite{GhifaryKZ14}. We hence consider the feature representations at the output of the common network as our source and target domain samples $\xs$ and $\xt$. Defining $\hsz \triangleq \xs \in \R^\dz$ and $\htz \triangleq \xt \in \R^\dz$, the relation between the features of layers $\lay$ and $\lay-1$ is given by
\begin{equation}
\label{eq_defn_hsl_htl}
\begin{split}
\hsl&=\actl( \Wsl \hslm + \bsl)\\
\htl&=\actl( \Wtl \htlm + \btl)
\end{split}
\end{equation}
for $\lay=1, \dots, \numL$, where $\hsl, \htl \in \Rdl$ are $\dl$-dimensional source and target features in layer $l$; the parameters $\Wsl, \Wtl \in \R^{\dl \times \dlm}$ are source and target weight matrices; the parameters $\bsl, \btl \in \Rdl $ are source and target bias vectors; $\actl: \Rdl \rightarrow \Rdl$ is a nonlinear activation function; $\numL$ is the depth of the network; and $\dl$ is the width of the network at layer $\lay$. We assume that the parameters of the output layer $\numL$ are common between the source and the target domains, such that $\WsL=\WtL=\WL \in \R^{m \times \dLm}$ and $\bsL=\btL=\bL \in \R^{m}$, where $m=\dL$ is the number of classes.

Let $\Thetasl=[\Wsl \ \bsl] \in \R^{\dl \times (\dlm+1)}$ and $\Thetatl =[\Wtl \ \btl] \in \R^{\dl \times (\dlm+1)}$ denote the matrices containing the network parameters of layer $\lay$. Let us also define the overall parameter structures 
\begin{equation*}
\begin{split}
\Thetas&=(\boldsymbol{\Theta}^{s1}, \dots, \boldsymbol{\Theta}^{s \numL}) \\
\Thetat&=(\boldsymbol{\Theta}^{t1}, \dots, \boldsymbol{\Theta}^{t \numL})
 \end{split}
\end{equation*}
containing the parameters of the entire source and target networks, respectively. We model the source and target domains to be compact sets and the network parameters to be bounded.
\begin{assumption}
\label{assum_Ax_Atheta}
The source and target domains are given by
\begin{equation}
\label{eq_bnd_xs_xt}
\Xs = \{  \xs \in \R^\dz : \| \xs \| \leq \Binp \},
\qquad \qquad
\Xt = \{  \xt \in \R^\dz : \| \xt \| \leq \Binp \}
\end{equation}
for some bound $\Binp>0$. Also, the network parameters $\Thetasl$, $\Thetatl $ in each layer belong to a closed and bounded set in $\R^{\dl \times (\dlm+1)} $ such that 
\begin{equation}
\label{eq_bnd_Thetaij}
\begin{split}
|\Thetasl_{ij}| , |\Thetatl_{ij}| \leq \Bnet
\end{split}
\end{equation}
for some magnitude bound parameter $\Bnet>0$, for $\lay=1, \dots, \numL$ and $i=1, \dots, \dl$; $j=1, \dots, \dlm+1$. 
\end{assumption}

Clearly, the features $\hsl$, $\htl$ in all layers depend on both the input vectors $\xs$, $\xt$ and the network parameters $\Thetas $, $\Thetat$. In the following, with a slight abuse of notation we write $\hsl_{\Thetas}$ when we would like emphasize the dependence of $\hsl$ on the network parameters $\Thetas$, and we write $\hsl (\xs)$ when we would like to refer to the dependence of $\hsl$ on the input $\xs$. The notation is set similarly for the target domain variables.

MMD-based deep domain adaptation networks employ a feature mapping $\phil: \Rdl \rightarrow \Xl $ between the hidden layer feature vectors $\hsl, \htl$ and a Reproducing Kernel Hilbert Space (RKHS) $\Xl$ \cite{LongCWJ15, GrettonBRSS12}. The RKHS $\Xl$ of each layer $\lay$ has a symmetric, positive definite characteristic kernel $\krl: \Rdl \times \Rdl \rightarrow \R$ such that
\begin{equation*}
\begin{split}
\krl(\hidl_1, \hidl_2 ) = \langle \phil(\hidl_1), \phil(\hidl_2) \rangle_{\Xl}
\end{split}
\end{equation*}
for any $\hidl_1, \hidl_2 \in \Rdl$, where $\langle \cdot, \cdot \rangle_{\Xl}$ denotes the inner product in the RKHS $\Xl$ \cite{GrettonBRSS12}. The feature mapping $\phil $ and the characteristic kernel $\krl$ are related as $\phil(\hidl)=\krl(\hidl,\cdot): \Rdl \rightarrow \R$ \cite{GrettonBRSS12}. The feature mapping $\phil $ has the property that  $\langle \phil(\hidl), \psi  \rangle_{\Xl} = \psi(\hidl)$  for any  $\psi \in \Xl$ and $\hidl \in \Rdl$.

In order to study this common framework within the setting of Section \ref{ssec_gen_bnd_mmd}, let us first define the functions $\fsl: \Xs \rightarrow \Xl$ and  $\ftl: \Xt \rightarrow \Xl$ as

\begin{equation}
\label{eq_defn_fsl_ftl}
\begin{split}
\fsl(\xs) \triangleq \phil(\hsl(\xs)) \in \Xl ,
\qquad \qquad
\ftl(\xt) \triangleq \phil(\htl(\xt))\in \Xl 
\end{split}
\end{equation}
for $\lay=1, \dots \, , \numL-1$. Note that the direct sum 
\begin{equation*}
\X=\bigoplus_{\lay=1}^{\numL-1} \Xl = \{ (f^1, f^2, \dots \, , f^{\numL-1}) : f^\lay \in \Xl, \, \lay=1, \dots, \numL-1 \}
\end{equation*}
of the RKHSs $\X^1, \dots \, , \X^{\numL-1}$ is also a Hilbert space with inner product $\langle \cdot , \cdot \rangle_{\X}$ given by \cite{DunfordS88}
\begin{equation}
\label{eq_defn_inn_prod_X}
\langle (f^1,  \dots \, ,  f^{\numL-1}) ,  (g^1,  \dots \, , g^{\numL-1}) \rangle_{\X}
= \sum_{\lay=1}^{\numL-1} 
\langle f^\lay, g^\lay \rangle_{\Xl}.
\end{equation}

Let us use the notation $\fsl_{\Thetas}(\xs)$ and $\ftl_{\Thetat}(\xt)$ for the functions $\fsl(\xs)$ and $\ftl(\xt)$ defined in \eqref{eq_defn_fsl_ftl} whenever we would like to emphasize their dependence on the network parameters. We can now define the function spaces
\begin{equation}
\label{eq_Fs_Ft_defn_dl}
\begin{split}
\Fs &= \{ \fs: \Xs \rightarrow \X \ | \ 
\fs(\xs)=\big( \fsone_{\Thetas} (\xs), \dots \, , \fsLm_{\Thetas}(\xs)    \big) \in \X, 
\ |\Thetasl_{ij}|  \leq \Bnet, \forall i, j\}  \\
\Ft &= \{ \ft: \Xt \rightarrow \X \ | \ 
\ft(\xt)= \big( \ftone_{\Thetat} (\xt), \dots \, , \ftLm_{\Thetat}(\xt)    \big) \in \X, 
\ |\Thetatl_{ij}|  \leq \Bnet, \forall i, j \}
\end{split}
\end{equation}
which define the mapping from the source and target domains to the feature representations composed of all layers from $\lay=1$ up to $ \lay= \numL-1$. As these features are passed through layer $\lay=\numL$ for the final classification stage, we can regard the network outputs  $\hsL $, $\htL $ as the composition of the mappings $\fs$, $\ft$ with the hypothesis function $\h$, i.e.,
\begin{equation}
\label{eq_gs_gt_defn}
\begin{split}
\gs(\xs) &= (\h \circ \fs)(\xs) \triangleq \hsL(\xs) \\
\gt(\xt) &= (\h \circ \ft)(\xt) \triangleq \htL(\xt).
\end{split}
\end{equation}
Let us also define the corresponding function spaces
\begin{equation}
\label{eq_Gs_Gt_defn_dl}
\begin{split}
\Gs &= \Hs \circ \Fs = \{ \gs: \Xs \rightarrow \Y \ | \ 
\gs(\xs)= \hsL_{\Thetas}(\xs) \in \Y \subset \R^m, 
\ |\Thetasl_{ij}|  \leq \Bnet, \forall i, j\} \\
\Gt &= \Hs \circ \Ft = \{ \gt: \Xt \rightarrow \Y \ | \ 
\gt(\xt)= \htL_{\Thetat}(\xt) \in \Y \subset \R^m, 
\ |\Thetatl_{ij}|  \leq \Bnet, \forall i, j\} .
\end{split}
\end{equation}

In the following, we first assume the continuity of the kernels and the activations. 

\begin{assumption}
\label{assum_krl_actl_cont}
The kernels $\krl(\cdot , \cdot ) $ for layers $\lay=1, \dots, \numL-1$ and the activation functions $\actl (\cdot)$ for layers $\lay=1, \dots, \numL$ are continuous.
\end{assumption}

As demonstrated in Lemma \ref{lem_fs_ft_measble}, this assumption ensures that $ E[\fs(\xs)]$ and $ E[\ft(\xt)]$ are in $ \X $, whose proof is presented in Appendix \ref{pf_lem_fs_ft_measble}.

\begin{lemma}
\label{lem_fs_ft_measble}
Let the condition in Assumption \ref{assum_krl_actl_cont} hold. Then the mappings $\fsl: \Xs \rightarrow \Xl$ and $\ftl:  \Xt \rightarrow \Xl$ for $\lay=1, \dots, \numL-1$, and the mappings $\fs:  \Xs \rightarrow \X $ and $\ft:  \Xt \rightarrow \X $ are measurable. Moreover, assuming that $E[\sqrt{\krl(\hsl, \hsl)}]<\infty$ and $E[\sqrt{\krl(\htl, \htl)}]<\infty$, the functions $E[\fsl(\xs)]: \Rdl \rightarrow \R$ and $E[\ftl(\xt)]: \Rdl \rightarrow \R$ defined as
\begin{equation*}
\begin{split}
E[\fsl(\xs)](\cdot) & \triangleq E[\fsl(\xs)(\cdot)] \\
E[\ftl(\xt)](\cdot) & \triangleq E[\ftl(\xt)(\cdot)] 
\end{split}
\end{equation*}
through the Borel probability measures $\mus$ and $\mut$ in the source and target domains  are in the RKHSs $\Xl$. Consequently, the functions 
\begin{equation*}
\begin{split}
 E[\fs(\xs)]& \triangleq ( E[\fsone (\xs)], \dots \, , E[\fsLm(\xs)]    )  \\ 
 E[\ft(\xt)]& \triangleq ( E[\ftone (\xt)], \dots \, , E[\ftLm(\xt)]    ) 
\end{split}
\end{equation*}
are in  the Hilbert space $\X$.
\end{lemma}

We next revisit the distribution discrepancy definition in Section  \ref{ssec_gen_bnd_mmd} for MMD-based neural networks. Let us define the distribution discrepancy in layer $\lay$ as
\begin{equation*}
\begin{split}
\D^\lay (\fsl, \ftl) \triangleq  \|  E_{\xs}[\fsl(\xs)] - E_{\xt}[ \ftl(\xt) ] \|_{\Xl}.
\end{split}
\end{equation*}
MMD-based domain adaptation algorithms typically seek to minimize the empirical estimate $\hD^\lay$ of $\D^\lay $ at each layer \cite{LongCWJ15}, \cite{TzengHZSD14}, \cite{GhifaryKZ14}. The empirical distribution discrepancy $\hD^\lay$ is obtained from the source and target sample sets $\{ \xis \}_{i=1}^{\Ns}$ and $\{ \xjt \}_{j=1}^{\Nt}$ as
\begin{equation*}
\begin{split}
& (\hD^\lay)^2 (\fsl, \ftl) = \left \|  \frac{1}{\Ns} \sum_{i=1}^{\Ns} \fsl(\xis) -  \frac{1}{\Nt} \sum_{j=1}^{\Nt} \ftl(\xjt)  \right \|_{\Xl}^2 \\
& =  \frac{1}{\Ns^2}  \sum_{i=1}^{\Ns}  \sum_{j=1}^{\Ns} \krl(\hsl_i, \hsl_j)
 -  \frac{2}{\Ns \Nt}  \sum_{i=1}^{\Ns}  \sum_{j=1}^{\Nt} \krl(\hsl_i, \htl_j)
 +  \frac{1}{\Nt^2}  \sum_{i=1}^{\Nt}  \sum_{j=1}^{\Nt} \krl(\htl_i, \htl_j)
\end{split}
\end{equation*}
where $\hsl_i$ and $\htl_j$ denote the source and target features in layer $\lay$ corresponding respectively to the samples $\xis$ and $\xjt$.  The second equality follows from the relations $\fsl(\xis)=\phil(\hsl_i)$ and $\ftl(\xjt)=\phil(\htl_j)$.

The overall distribution discrepancy between the source and the target domains defined in \eqref{eq:defn_D_MMD} is given by
\[
\D(\fs, \ft) = \|  E_{\xs}[\fs(\xs)] - E_{\xt}[ \ft(\xt) ] \|_\X
\]
following the definitions in Lemma \ref{lem_fs_ft_measble} in the current setting. Its empirical estimate $\hD(\fs, \ft)$ defined in \eqref{eq:defn_hD_MMD} is then obtained as
\begin{equation}
\begin{split}
& \hD^2(\fs, \ft) = \left \| 
\frac{1}{\Ns} \sum_{i=1}^{\Ns} \fs(\xis) -  \frac{1}{\Nt} \sum_{j=1}^{\Nt} \ft(\xjt) 
\right \|_\X^2 \\
&= \frac{1}{\Ns^2}  \sum_{i=1}^{\Ns}  \sum_{j=1}^{\Ns} \langle \fs(\xis), \fs(\xjs) \rangle_\X
 -  \frac{2}{\Ns \Nt}  \sum_{i=1}^{\Ns}  \sum_{j=1}^{\Nt}  \langle \fs(\xis), \ft(\xjt) \rangle_\X   \\
& +  \frac{1}{\Nt^2}  \sum_{i=1}^{\Nt}  \sum_{j=1}^{\Nt}  \langle \ft(\xit), \ft(\xjt) \rangle_\X  \\
 &= \sum_{\lay=1}^{\numL-1}  (\hD^\lay)^2 (\fsl, \ftl) 
\end{split}
\label{eq_hD2_dda}
\end{equation}
where the last equality follows from the definition \eqref{eq_defn_inn_prod_X} of the inner product in $\X$. 

Most MMD-based deep domain adaptation networks rely on aligning the source and the target domains by minimizing the total MMD distance \eqref{eq_hD2_dda} summed over all layers \cite{WangD18}, \cite{LongCWJ15}, \cite{TzengHZSD14}, \cite{GhifaryKZ14}. We thus consider a learning algorithm that minimizes the overall loss
\begin{equation}
\label{eq_obj_learning_mmd}
\begin{split}
\min_  {\fs \in \Fs,  \  \ft \in \Ft, \ \h \in  \Hs}
(1-\alpha) \hLs (\fs, \h) + \alpha \hLt (\ft, \h) 
+ \beta \sum_{\lay=1}^{\numL-1}  (\hD^\lay)^2 (\fsl, \ftl).
\end{split}
\end{equation}
Hence, the above analysis provides the bridge between the results in Section \ref{ssec_gen_bnd_mmd} and the current setting with MMD-based domain adaptation networks, so that the statement of Theorem \ref{thm:main_result_mmd} applies to the current problem. Before we proceed with the implications of Theorem \ref{thm:main_result_mmd}, we need two additional assumptions.

\begin{assumption}
\label{assum_Lk_Leta}
The symmetric kernel $\krl(\cdot, \cdot): \Rdl \times \Rdl \rightarrow \R$ is Lipschitz continuous with constant $\Lk$ in each argument, such that
\begin{equation}
\label{eq_assum_kernel_lip_cont}
\begin{split}
|  \krl( \boldsymbol{\xi}_1, \boldsymbol{\xi})  -  \krl( \boldsymbol{\xi}_2, \boldsymbol{\xi})   | \leq \Lk \| \boldsymbol{\xi}_1 - \boldsymbol{\xi}_2 \|
\end{split}
\end{equation}
for all $\boldsymbol{\xi}_1, \boldsymbol{\xi}_2, \boldsymbol{\xi} \in \Rdl$.  Also, the nonlinear activation functions $\actl$ in \eqref{eq_defn_hsl_htl} are Lipschitz-continuous with constant $\Leta$, such that
\begin{equation}
\label{eq_Lipcont_act}
\begin{split}
\| \actl(\mathbf{u}) - \actl(\mathbf{v}) \| \leq \Leta \, \| \mathbf{u}-\mathbf{v} \|
\end{split}
\end{equation}
for all $\mathbf{u}, \mathbf{v} \in \Rdl$, for $\lay=1, \dots, \numL$. 
\end{assumption}

\begin{assumption}
\label{assum_bnd_act_val_op}
The nonlinear activation functions $\actl$ in \eqref{eq_defn_hsl_htl} are bounded either in value (e.g., sigmoid, softmax) or as an operator (e.g., ReLU). In the former case, we assume that there exists a constant $\Beta>0$ with
\begin{equation}
\label{eq_bnd_act_value}
\begin{split}
| \actl_i(\mathbf{u})  | \leq \Beta 
\end{split}
\end{equation}
for all $\mathbf{u} \in \Rdl$, for $\lay=1, \dots, \numL-1$ and $i = 1, \dots, \dl$, where $\actl_i(\mathbf{u})$ denotes the $i$-th component of $\actl(\mathbf{u})$. In the latter case, we assume that there exists $\Bopeta >0$ such that
\begin{equation}
\label{eq_bnd_act_op}
\begin{split}
\| \actl(\mathbf{u})  \| \leq \Bopeta  \| \mathbf{u} \| 
\end{split}
\end{equation}
for all $\mathbf{u} \in \Rdl$, for $\lay=1, \dots, \numL-1$.

\end{assumption}

The Lipschitz continuity condition \eqref{eq_assum_kernel_lip_cont} holds for many widely used kernels such as Gaussian kernels. As for condition \eqref{eq_Lipcont_act}, the Lipschitz constants of the commonly used rectified linear unit, softmax and softplus activation functions are derived in Appendix \ref{sec_app_lip_nonlinact}. In the following result we show that the transformation function classes $\Fs, \Ft$ as well as the composite function classes $\Gs$, $\Gt$  are compact metric spaces.

\begin{lemma}
\label{lem_Fs_Ft_HFs_HFt_comp}
Let Assumptions \ref{assum_Ax_Atheta}-\ref{assum_Lk_Leta} hold. Then, the transformation function classes $\Fs, \Ft$ in \eqref{eq_Fs_Ft_defn_dl} and the composite function classes $\Gs, \Gt$ in \eqref{eq_Gs_Gt_defn_dl} are compact metric spaces, respectively under the metrics $\dXs, \dXt$ in \eqref{eq_defn_dXs_dXt}, and the metrics $\ds, \dt$ in \eqref{eq_defn_ds_dt}.
\end{lemma}

The proof of Lemma \ref{lem_Fs_Ft_HFs_HFt_comp} is presented in Appendix \ref{pf_lem_Fs_Ft_HFs_HFt_comp}. Having established the compactness of the function classes, we can now study the corresponding covering numbers.

\begin{lemma} 
\label{lem_cov_num_Fs_Ft}
Let Assumptions \ref{assum_Ax_Atheta}, \ref{assum_Lk_Leta}, \ref{assum_bnd_act_val_op} hold. Then, the covering numbers of the function classes $\Fs$ and $\Ft$ are upper bounded as 
\begin{equation*}
\begin{split}
\N (  \Fs, \epsilon, \dXs) \leq \prod_{\lay=1}^{\numL-1}  \left( \frac{4 \Bnet  \Lk \BQ}{\epsilon^2}+1  \right)^{ \dl(\dlm+1) } \\
\N(  \Ft, \epsilon, \dXt) \leq \prod_{\lay=1}^{\numL-1}  \left( \frac{4 \Bnet  \Lk \BQ}{\epsilon^2}+1  \right)^{ \dl(\dlm+1) }
\end{split}
\end{equation*}
where the dimension-dependent constant $\BQ$ is defined as
\begin{equation*}
\begin{split}
\BQ \triangleq 
\sum_{\lay=1}^{\numL-1} \BQdiml
\end{split}
\end{equation*}
with
\begin{equation}
\label{eq_defn_Ql}
\begin{split}
\BQdiml
&\triangleq 
(\Leta \Bdimlm \sqrt{\dl \dlm}+ \Leta \sqrt{\dl}) \\
&+
\sum_{i=1}^{\lay-1}
(\Leta \Bdimim \sqrt{\di \dimm}
+ \Leta \sqrt{\di})
\prod_{k=i+1}^{\lay}
\Leta \Bnet \sqrt{\dk \dkm}
\end{split}
\end{equation}
for $\lay=2, \dots, \numL$ and 
$\BQdimone \triangleq 
\Leta \sqrt{\done \dz} \, \Bdimz  +\Leta \sqrt{\done} $.
 Here
\begin{equation*}
\begin{split}
\Bdiml &\triangleq
(\Bopeta \Bnet) ^l  (\Binp \sqrt{\dz}+1) \sqrt{\done}
\prod_{k=1}^{\lay-1} \sqrt{\dkplone \dk} \\
&+
\sum_{i=2}^{\lay-1} (\Bopeta \Bnet)^{\lay+1-i} \sqrt{\di}
\prod_{k=i}^{\lay-1} \sqrt{\dkplone \dk}
+
\Bopeta \Bnet \sqrt{\dl}
\end{split}
\end{equation*}
under condition  \eqref{eq_bnd_act_op}
and $\Bdiml \triangleq\Beta \sqrt{\dl} $ under condition \eqref{eq_bnd_act_value} for $\lay=2, \dots, \numL-1$, where $\Bdimz \triangleq \Binp$ and $\Bdimone \triangleq \Bopeta  \Bnet \sqrt{\done \dz}  \Binp +  \Bopeta \Bnet \sqrt{\done}$. 
\end{lemma} 

Lemma \ref{lem_cov_num_Fs_Ft} is proved in Appendix \ref{pf_lem_cov_num_Fs_Ft}. A similar result is obtained for the function spaces $\Hs \circ \Fs$ and $\Hs \circ \Ft$  in the following lemma, which is proved in Appendix \ref{pf_lem_cov_num_HFs_HFt}.

\begin{lemma} 
\label{lem_cov_num_HFs_HFt}
Let Assumptions \ref{assum_Ax_Atheta}, \ref{assum_Lk_Leta}, \ref{assum_bnd_act_val_op} hold. Then, the covering numbers of the function classes $\Hs \circ \Fs$ and $\Hs \circ \Ft$ are upper bounded as 
\begin{equation*}
\begin{split}
\N( \Hs \circ \Fs, \epsilon, \ds) &\leq    \prod_{\lay=1}^{\numL}  \left( \frac{2 \Bnet \BQdimL}{\epsilon}+1  \right)^{ \dl(\dlm+1) } \\
\N( \Hs \circ \Ft, \epsilon, \dt) & \leq  \prod_{\lay=1}^{\numL}  \left( \frac{2 \Bnet \BQdimL}{\epsilon}+1  \right)^{ \dl(\dlm+1) }.
\end{split}
\end{equation*}
\end{lemma}

\begin{corollary}
\label{cor_covnum_rate}

Consider that the feature dimensions $\dl$ are such that $\dl = O(\dcom)$ for $\lay=1, \dots, \numL$, for some common network width parameter $\dcom$. Then, the rate of growth of the covering numbers for the function spaces $\N (  \Fs, \epsilon, \dXs) $, $\N (  \Ft, \epsilon, \dXt) $, $\N ( \Hs \circ \Fs, \epsilon, \ds) $, $\N ( \Hs \circ \Ft, \epsilon, \dt) $  with the width $\dcom$ and the depth $\numL$ of the network is upper bounded by 
\[
O\left( \left( \frac{  \numL }{\epsilon}  \right)^{\dcom^2 \numL}   (c \dcom)^{\dcom^2 \numL^2} \right)
\]
where $c$ denotes a constant.
\end{corollary}

Corollary \ref{cor_covnum_rate} is proved in Appendix \ref{pf_cor_covnum_rate}. Combining Corollary \ref{cor_covnum_rate} and Theorem \ref{thm:main_result_mmd}, we are now ready to state our main result about the sample complexity of MMD-based domain adaptation networks in Theorem \ref{thm_main_result_da_mmd} below, whose proof is presented in Appendix \ref{pf_thm_main_result_da_mmd}.

\begin{theorem}
\label{thm_main_result_da_mmd}

Consider a learning algorithm relying on the minimization of a loss function of the form \eqref{eq_obj_learning_mmd} via an MMD-based domain adaptation network. Assume that the classification loss function $\loss$ is bounded by a constant $\bls$ and Lipschitz continuous with respect to the first argument with constant $\Lls$.  Suppose that the source and target data distributions satisfy Assumptions \ref{assum_existence_LLs} and \ref{assum_fx_bdd_moments}. Assume also that the network parameters, activation functions  and the kernels satisfy Assumptions \ref{assum_Ax_Atheta}-\ref{assum_bnd_act_val_op}. 

Consider that the weight parameter $\alpha$ in the loss function is chosen such that
\begin{equation*}
\begin{split}
\alpha
= O \left(
\left(
\frac
{\Mt \epsilon^2}
{\dcom^2 \numL \log \left( \frac{  \numL }{\epsilon} \right)  \, 
+ 
\dcom^2 \numL^2 \log(\dcom) }
\right)^{1/2}
\right)
\end{split}
\end{equation*}
according to the number $\Mt$ of available labeled target samples. Then in order to bound the expected target loss with a  generalization gap of $O(\epsilon)$ as
\begin{equation}
\label{eq_accuracy_thm4}
\begin{split}
\Lt(\ft, \h) \leq   \hLw(\fs, \ft, \h)+ (1-\alpha) \LLs  \hD(\fs, \ft) + (1-\alpha) \LLs \epsilon + \epsilon,
\end{split}
\end{equation}
the sample complexities in terms of the number $\Ms$ of labeled source samples, the number $\Ns$ of all (labeled and unlabeled) source samples, and the number $\Nt$  of all target samples are upper bounded by
\begin{equation*}
\begin{split}
O\left(
\frac{\dcom^2 \numL \log \left(\frac{\numL}{\epsilon}\right) 
+ \dcom^2 \numL^2 \log(\dcom) }
{\epsilon^2}
\right) . 
\end{split}
\end{equation*}
\end{theorem}

Note that the assumption of the existence of the constants $\bls$ and $\Lls$ in Theorem \ref{thm_main_result_da_mmd} is satisfied in many common settings.  In Appendix \ref{sec_app_crossent}, we derive these constants for the commonly used cross-entropy loss function.  We can draw several conclusions from the statement of Theorem \ref{thm_main_result_da_mmd}. The sample complexity expressions obtained in the theorem indicate that, as the network depth $\numL$ and the network width $\dcom$ increase,  $\Ms$, $\Ns$, and $\Nt$ must increase at rate $O(\dcom^2 \numL^2)$, if the logarithmic terms are ignored for simplicity. This result shows that the number of labeled source samples and the number all source and target samples required for preventing overfitting must grow quadratically with both $\numL$ and $\dcom$ as the network size increases. On the other hand, the number $\Mt$ of available labeled target samples is typically limited in domain adaptation scenarios. Regarding this, Theorem \ref{thm_main_result_da_mmd} also has some implications on the optimal choice of the weight parameter $\alpha$ that finds a suitable balance between the target and source classification losses. As the number $\Mt$ of labeled target samples decreases, the weight $\alpha$ of the target classification loss must also shrink at rate $\alpha=O(\sqrt{\Mt})$ in order to avoid overfitting the model to the few available target labels. Similarly, as the network size grows, the weight parameter $\alpha$ must also shrink at rate $\alpha=O((\dcom \numL)^{-1})$ with $\dcom$ and $\numL$. The parameter $\epsilon$ in the theorem is a probability constant that sets the tradeoff between the desired accuracy level and the number of required training samples. In order for the expected target loss not to exceed the empirical losses by more than $O(\epsilon)$ in \eqref{eq_accuracy_thm4}, the number of samples $\Ms, \Ns, \Nt$ must scale at an inverse quadratic rate $O(\epsilon^{-2})$ with $\epsilon$.

\subsection{Adversarial domain adaptation networks}
\label{ssec_adv_da_net}

\begin{figure}[t]
  \centering
  \centerline{\includegraphics[width=13.0cm]{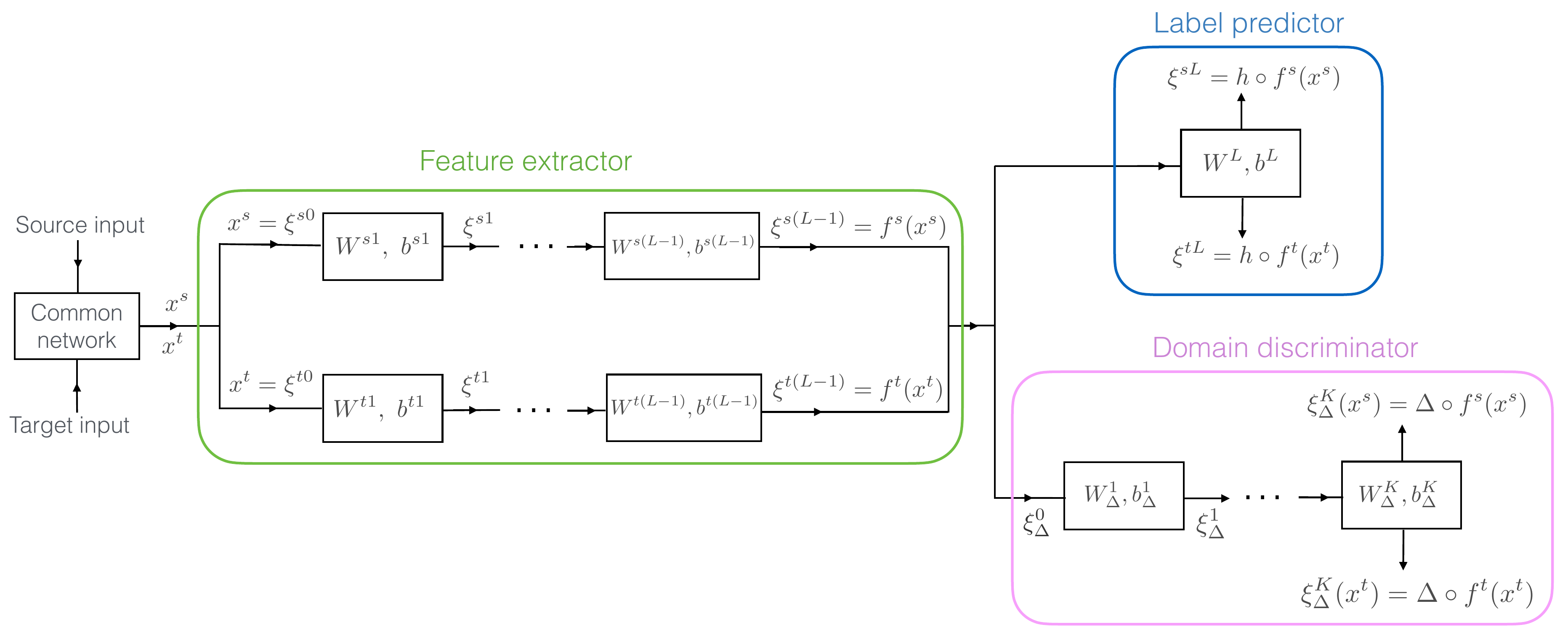}}
  \caption{Illustration of adversarial domain adaptation networks}
  \label{fig_illus_ddan_network}
\end{figure}

In this section, we extend our results to analyze the sample complexity of adversarial domain adaptation networks.  Adversarial models have been widely used in domain adaptation since the leading studies \cite{GaninUAGLLML16}, \cite{TzengHSD17}, \cite{LongC0J18}, and have been applied to a variety of problems in recent works \cite{SinghalWRK23}. Domain-adversarial neural networks aim to compute domain-invariant representations $\fs: \Xs \rightarrow \X$, $\ft: \Xt \rightarrow \X$ through a feature extractor network, followed by a label predictor $\h: \X \rightarrow \Y$ that provides the class label at its output as illustrated in Figure \ref{fig_illus_ddan_network}. The domain-invariance of the learnt features is ensured by a domain discriminator network, which is trained to determine whether the features belong to the source domain or the target domain. The feature extractor and the domain discriminator networks are trained in an adversarial fashion, such that the feature extractor aims to learn domain-invariant representations whose domains are indistinguishable by the domain discriminator. The domain discriminator $\ddan: \X \rightarrow \R$ seeks to minimize the domain discrimination loss
\[
\Ls_\dom( \fs, \ddan) + \Lt_\dom( \ft, \ddan)
\]
where
\begin{equation*}
\begin{split}
\Ls_\dom( \fs, \ddan)=E[\loss_\dom(\ddan \circ \fs(\xs), \ydoms)],
\qquad
\Lt_\dom( \ft, \ddan) = E[ \loss_\dom( \ddan \circ \ft(\xt), \ydomt )]
\end{split}
\end{equation*}
respectively denote the expected domain discrimination losses in the source and the target domains;  
$\loss_\dom: \R \times \R \rightarrow [0, \infty)$ is a domain discrimination loss function; and $\ydoms,  \ydomt \in \R$ denote the domain labels of the source and the target domains. It is common practice to set the domain discrimination loss $\loss_\dom$ as a logarithmic penalty on the deviation between the estimated domain labels and the true domain labels $\ydoms =0, \ydomt =1$ as \cite{GaninUAGLLML16}, \cite{TzengHSD17}, \cite{LongC0J18}
\begin{equation}
\label{eq_log_penalty_domainloss}
\begin{split}
\loss_\dom( \ddan \circ \fs(\xs), \ydoms ) = - \log(1-\ddan \circ \fs(\xs)) \\
\loss_\dom( \ddan \circ \ft(\xt), \ydomt ) = - \log(\ddan \circ \ft(\xt)). 
\end{split}
\end{equation}
Meanwhile, the feature extractor network is trained to maximize the domain classification loss so that the learnt features are domain-invariant, leading to the overall optimization problem 
\begin{equation}
\label{eq_ddan_obj_func}
\begin{split}
\min_  {\fs,  \ft, \h, \ddan}
(1-\alpha) \hLs (\fs, \h) + \alpha \hLt (\ft, \h) 
- \beta \big( \hLs_\dom (\fs, \ddan) +\hLt_\dom( \ft, \ddan) \big)
\end{split}
\end{equation}
where $\hLs, \hLt$ denote the empirical source and target classification losses defined in \eqref{eq_emp_cl_loss}. Here  $\hLs_\dom,  \hLt_\dom$ are the empirical domain discrimination losses given by
\begin{equation*}
\begin{split}
\hLs_\dom( \fs, \ddan) &= \frac{1}{\Ns} \sum_{i=1}^{\Ns}  \loss_\dom( \ddan \circ \fs(\xis), \ydoms_i )  \\
\hLt_\dom( \ft, \ddan) &=  \frac{1}{\Nt} \sum_{j=1}^{\Nt} \loss_\dom( \ddan \circ \ft(\xjt), \ydomt_j  ) 
\end{split}
\end{equation*}
where $\ydoms_i $ and $\ydomt_j $ respectively denote the domain labels of the source samples $\xis $ and the target samples $\xjt$. 

In order to study domain-adversarial network models within our framework, we consider that the transformations $\fs, \ft$ are given by the feature representations at layer $L-1$ of the feature extractor network. The corresponding function spaces are then 
\begin{equation*}
\begin{split}
\Fs &= \{ \fs: \Xs \rightarrow \R^{\dLm} \, | \,  \fs(\xs)= \hsLm_{\Thetas}(\xs), \ |\Thetasl_{ij}|  \leq \Bnet, \forall i, j\}  \\
\Ft &= \{ \ft: \Xt \rightarrow \R^{\dLm} \, | \,  \ft(\xt)= \htLm_{\Thetat}(\xt), \ |\Thetatl_{ij}|  \leq \Bnet, \forall i, j\}.  
\end{split}
\end{equation*}
Similarly, the hypotheses $\h \circ \fs$ and $\h \circ \ft$ are given by the output of the last layer $L$ 
\begin{equation*}
\begin{split}
\h \circ \fs(\xs)&= \hsL(\xs), 
\qquad
\h \circ \ft(\xt)= \htL(\xt)
\end{split}
\end{equation*}
with the function spaces $\Hs \circ \Fs$ and $\Hs \circ \Ft$ defined\footnote{Note that, the definitions of the function spaces $\Fs, \Ft$ in this section are different from those in Section \ref{ssec_samp_comp_mmd_net}, as they take different roles between MMD-based and adversarial  networks. Nevertheless, the composite function spaces $\Gs = \Hs \circ \Fs$ and $\Gt = \Hs \circ \Ft$ in this section are the same as those of Section \ref{ssec_samp_comp_mmd_net}, since the functions $\gs, \gt$ are defined through the classification layer output in both the MMD-based and the adversarial settings.} in \eqref{eq_Gs_Gt_defn_dl}. Here, the features between layers $\lay-1$ and $\lay$ are related as in \eqref{eq_defn_hsl_htl} through the network parameters $\Wsl, \Wtl, \bsl, \btl$ and the nonlinear activation functions $\actl$. While feature extractor networks typically consist of several convolutional layers followed by fully connected layers in many common architectures \cite{SinghalWRK23}; in domain adaptation applications it is a common strategy to adopt convolutional layer weights from pretrained networks or to train or fine-tune them using only source data \cite{TzengHSD17}. Therefore, we leave the training of convolutional layers out of the scope of our analysis. We consider the input source and target samples $\xs, \xt \in \R^{\dz}$ to be the response generated at the output of the convolutional network common between the two domains as illustrated in Figure \ref{fig_illus_ddan_network} and focus on the action of the fully connected layers of the feature extractor networks.

The domain discriminator network typically consists of several fully connected layers \cite{GaninUAGLLML16}, \cite{TzengHSD17}. Denoting the weight parameters of these layers as $\Wldan \in \R^{\dlddan \times \dlmddan}$, $\bldan \in \R^{\dlddan}$, the relation between the responses $\hidlmdan \in \R^{\dlmddan},  \hidldan \in \R^{\dlddan}$ at layers $\lay-1$ and $ \lay$ is given by
\begin{equation*}
\begin{split}
\hidldan= \actldan( \Wldan \hidlmdan + \bldan)
\end{split}
\end{equation*}
for $\lay=1, \dots, \numLdan $, where $\numLdan $ denotes the number of layers and $\actldan: \R^{\dlddan} \rightarrow \R^{\dlddan}  $ denotes the activation function of the domain discriminator network at layer $\lay$.  Here, the input $\hidzdan $  to the domain discriminator network corresponds to the outputs $\hsLm, \htLm$ of the feature extractor networks. The domain discriminator output is then given by
\begin{equation*}
\begin{split}
\ddan \circ \fs(\xs) = \hidKdan (\xs),
\qquad
\ddan \circ \ft(\xt) = \hidKdan (\xt)
\end{split}
\end{equation*}
for the source and the target domains, where the dimension of the output layer of the domain discriminator is $\dKddan=1$. Still using Assumption \ref{assum_Ax_Atheta} and extending it to the domain discriminator network as well, we define the function class of domain discriminators with bounded network weights as

\begin{equation}
\label{eq_Dspace_defn}
\begin{split}
\Dspace &= \{ \ddan: \R^{\dLm} \rightarrow \R \ | \ 
\ddan(\hidzdan)=\hidKdan , 
\
| ({\Wldan})_{ij}|  \leq \Bnet, \
 |({\bldan})_{i}|   \leq \Bnet,
  \forall i, j\}  .
\end{split}
\end{equation}

Provided that the adversarial domain adaptation network is well-trained, the mappings $\fs(\xs)$, $\ft(\xt)$ specialize in the extraction of domain-invariant features such that the domain discriminator cannot distinguish between the source and the target samples. The discriminator outputs  $\ddan \circ \fs(\xs)$ and $\ddan \circ \ft(\xt) $ then take similar values. Based on this observation, we build our analysis on the following definition of the distribution distance
\begin{equation*}
\begin{split}
\Ddan(\fs, \ft) \triangleq \left | E[\ddan \circ \fs(\xs) ]  - E [\ddan \circ \ft(\xt) ]  \right |.
\end{split}
\end{equation*}
The distribution distance $\Ddan(\fs, \ft)$  measures how well the source and target distributions are aligned once they are mapped to the shared feature space by the mappings $\fs$ and $\ft$. Note that the above definition of the distribution distance $\Ddan(\fs, \ft)$ depends also on the domain discriminator $\ddan$. We make the following assumption about the domain discriminator.

\begin{assumption}
\label{assum_ddan_bdd}
The domain discriminator output is bounded, i.e., there exists a constant $\Bddan>0$ such that
\begin{equation*}
\begin{split}
|\ddan(\hidzdan)|=|\hidKdan| \leq \Bddan
\end{split}
\end{equation*}
for all $\hidzdan \in \R^{\dLm}$.
\end{assumption}
 
Note that Assumption \ref{assum_ddan_bdd} is satisfied for many domain-adversarial networks, as the activation function $\actKdan $ of the final domain discriminator layer is often selected as a bounded function such as the sigmoid  \cite{GaninUAGLLML16} or the softmax function \cite{TzengHDS15}. Let us denote the composition of the domain discriminator and the feature extractor as
 \begin{equation*}
\begin{split}
\vs(\xs) \triangleq \ddan \circ \fs (\xs),
\qquad
\vt(\xt) \triangleq \ddan \circ \ft (\xt)
\end{split}
\end{equation*}
and the corresponding function spaces as
\begin{equation*}
\begin{split}
\Vs &= \Dspace \circ \Fs = 
\{
\vs: \vs= \ddan \circ \fs,  \ddan \in \Dspace, \fs \in \Fs
\} \\
\Vt &= \Dspace \circ \Ft = 
\{
\vt: \vt= \ddan \circ \ft,  \ddan \in \Dspace, \ft \in \Ft
\}.
\end{split}
\end{equation*}

In order to study the sample complexity of adversarial domain adaptation networks,  we first characterize in the following lemma the deviation between the expected distribution distance $\Ddan(\fs, \ft)$ and its finite-sample estimate 
\begin{equation*}
\begin{split}
\hDdan(\fs, \ft) = \left |
 \frac{1}{\Ns} 
  \sum_{i=1}^{\Ns} \ddan \circ \fs(\xis)   - 
   \frac{1}{\Nt}     \sum_{j=1}^{\Nt} \ddan \circ \ft(\xjt)  
 \right |.
\end{split}
\end{equation*}
\begin{lemma}
\label{lem_ddan_hddan_dev}
Let Assumption \ref{assum_ddan_bdd} hold. Assume also that the composite function classes $\Vs$ and $\Vt$ are compact with respect to the metrics 
\begin{equation*}
\begin{split}
\dVs(\vs_1, \vs_2) & \triangleq \sup_{\xs \in \Xs}  | \vs_1(\xs) - \vs_2(\xs) | \\
\dVt(\vt_1, \vt_2) & \triangleq \sup_{\xt \in \Xt}  | \vt_1(\xt) - \vt_2(\xt) | \\
\end{split}
\end{equation*}
where $\vs_1, \vs_2 \in \Vs$ and $\vt_1, \vt_2 \in \Vt$.
Then,  
\begin{equation*}
\begin{split}
P & \left(
\sup_{\fs \in \Fs, \ft \in \Ft, \ddan \in \Dspace} 
 |  \Ddan(\fs, \ft)  - \hDdan(\fs, \ft)  | \leq  \epsilon
 \right ) \\
& \geq
 1 - 2 \N(\Vs, \frac{\epsilon}{6}, \dVs)  \exp \left( -\frac{\Ns \epsilon^2}{72 \Bddansq} \right) 
- 2 \N(\Vt, \frac{\epsilon}{6}, \dVt) \exp \left( -\frac{\Nt \epsilon^2}{72 \Bddansq} \right) .
\end{split}
\end{equation*}
\end{lemma}

The proof of Lemma \ref{lem_ddan_hddan_dev} is presented in Appendix \ref{pf_lem_ddan_hddan_dev}. Note that Lemma  \ref{lem_ddan_hddan_dev}  is the counterpart of Lemma \ref{lem:unif_bnd_D_hD} in the domain-adversarial setting. Before stating the main result of this section, we formalize the following conditions. 

\begin{assumption}
\label{assum_actddan_cont_Lip}
The activation functions $\actl (\cdot)$ for layers $\lay=1, \dots, \numL$ and the activation functions $\actldan (\cdot)$ for layers  $\lay=1, \dots, \numLdan$ are continuous and also Lipschitz-continuous with constant $\Leta$, such that
\begin{equation}
\label{eq_Lipcont_act_ddan}
\begin{split}
\| \actl(\mathbf{u}) - \actl(\mathbf{v}) \| \leq \Leta \, \| \mathbf{u}-\mathbf{v} \|
\end{split}
\end{equation}
for all $\mathbf{u}, \mathbf{v} \in \Rdl$, for $\lay=1, \dots, \numL$ and
\begin{equation}
\label{eq_Lipcont_actddan_ddan}
\begin{split}
\| \actldan(\mathbf{u}) - \actldan(\mathbf{v}) \| \leq \Leta \, \| \mathbf{u}-\mathbf{v} \|
\end{split}
\end{equation}
for all $\mathbf{u}, \mathbf{v} \in \R^{\dlddan}$, for $\lay=1, \dots, \numLdan$.

\end{assumption}

\begin{figure}[t]
\begin{center}
     \subfigure[Poor alignment]
       {\label{fig:illus_domain_ddan_a}\includegraphics[height=4.5cm]{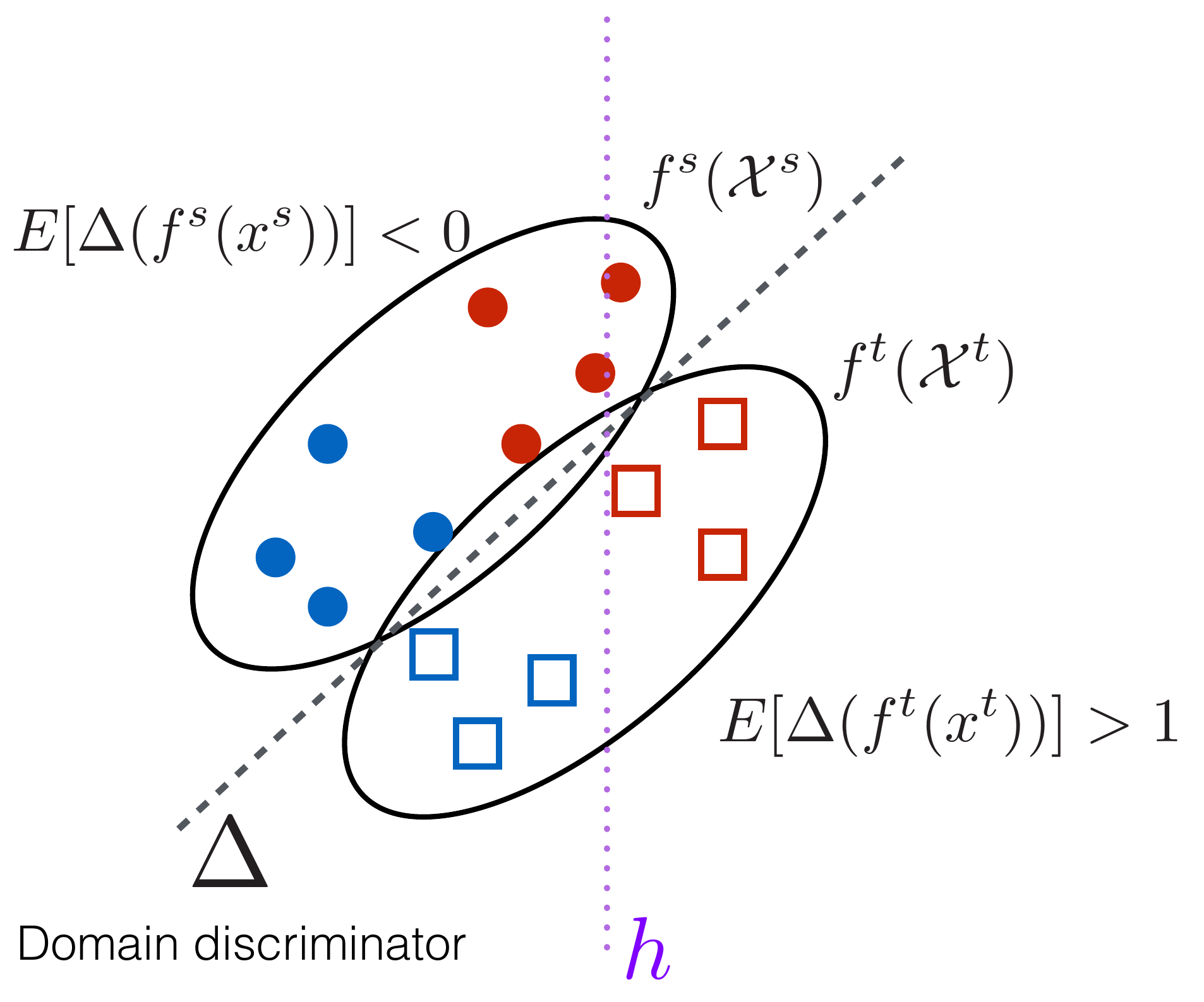}}
       \hspace{1cm}
     \subfigure[Proper alignment]
       {\label{fig:illus_domain_ddan_b}\includegraphics[height=4cm]{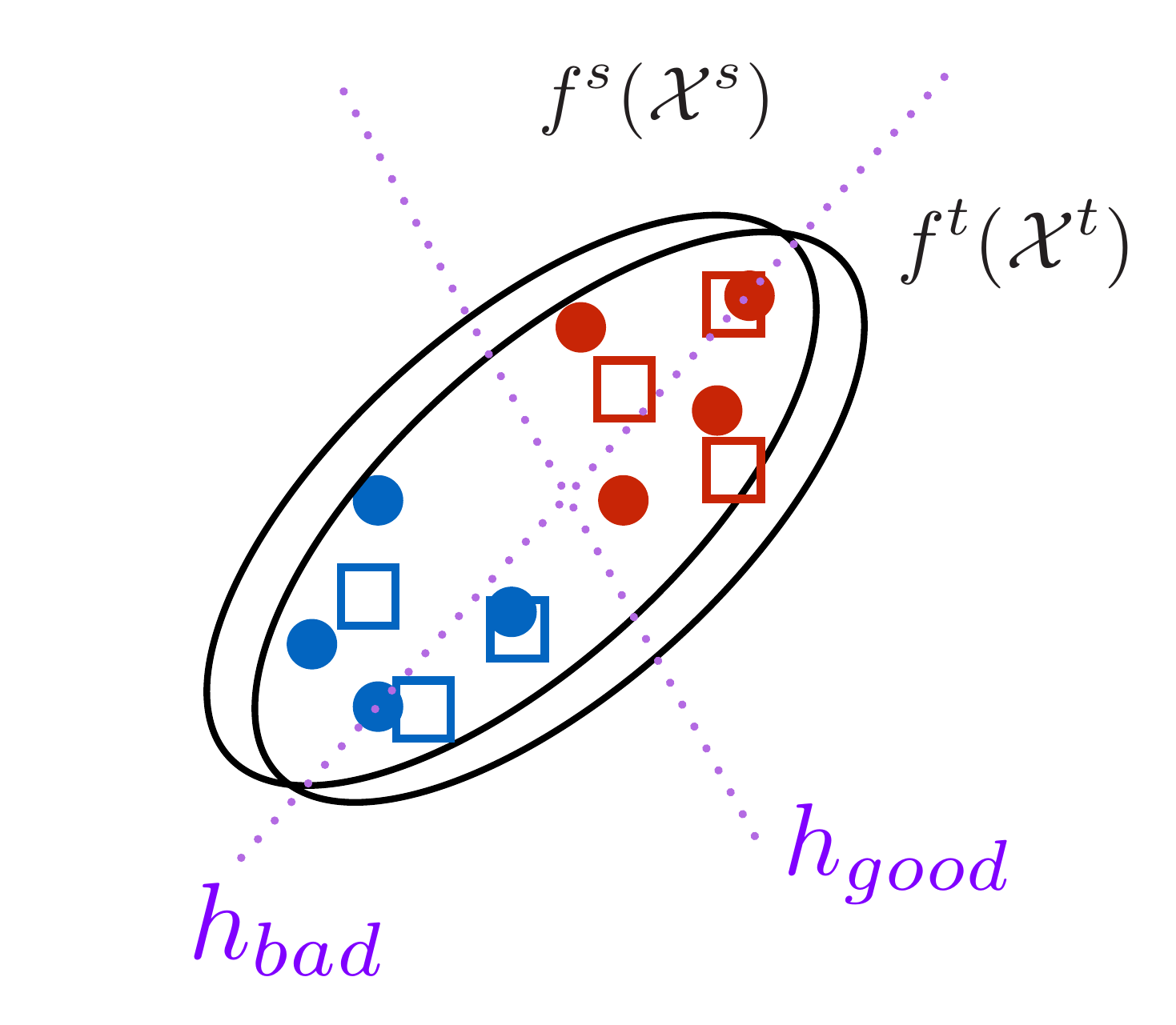}}
 \end{center}
 \vspace{-0.5cm}
 \caption{Illustration of Assumption \ref{assum_existence_LLsdan}. Red and blue colors represent two different classes in the source and target domains. In (a), the two domains are poorly aligned by the mappings $\fs$ and $\ft$, therefore, the algorithm learns a domain discriminator $\ddan $ that can separate the two domains well. The domain distance $\Ddan(\fs, \ft)$ is then high, and consequently, there may exist hypotheses $\h$ yielding a small loss in one domain and a large loss in the other domain. In (b), the domains are well-aligned and the domain distance $\Ddan(\fs, \ft)$ is small. The source and target losses are then similar for any hypothesis $\h$.}
 \label{fig_illus_domain_ddan}
\end{figure}

\begin{assumption}
\label{assum_bnd_act_val_op_ddan}
The nonlinear activation functions $\actldan$  are bounded either in value or as an operator, for $\lay=1, \dots, \numLdan-1$. In the former case, there exists a constant $\Beta>0$ with
\begin{equation}
\label{eq_bnd_act_value_ddan}
\begin{split}
| ({\actldan})_i(\mathbf{u)}  | \leq \Beta 
\end{split}
\end{equation}
for all $\mathbf{u} \in \R^{\dlddan}$, where $({\actldan})_i(\mathbf{u})$ denotes the $i$-th component of $\actldan(\mathbf{u})$. In the latter case,  there exists $\Bopeta >0$ such that
\begin{equation}
\label{eq_bnd_act_op_ddan}
\begin{split}
\| \actldan(\mathbf{u})  \| \leq \Bopeta  \| \mathbf{u} \| 
\end{split}
\end{equation}
for all $\mathbf{u} \in \R^{\dlddan}$.
\end{assumption}

Note that Assumption \ref{assum_actddan_cont_Lip} is an adaptation of the conditions in Assumptions \ref{assum_krl_actl_cont} and \ref{assum_Lk_Leta} to the domain-adversarial setting in consideration. Similarly, Assumption \ref{assum_bnd_act_val_op_ddan} simply adapts the condition in Assumption \ref{assum_bnd_act_val_op} to the domain discriminator network. We lastly make the following assumption about the link between the distribution distance and the deviation between the source and target losses.

\begin{assumption}
\label{assum_existence_LLsdan}
There exists a constant $\LLsdan>0$ such that, for the domain discriminator $\ddan \in \Dspace $ learnt by the algorithm, we have
\begin{equation}
\label{eq:Lipsc_reg_loss_ddan}
| \Ls(\fs, \h) - \Lt(\ft, \h) | \leq \LLsdan \, \Ddan(\fs, \ft)
\end{equation}
for any transformations $\fs \in \Fs$, $\ft \in \Ft$, and any hypothesis $\h \in \Hs$. 
\end{assumption} 
Assumption \ref{assum_existence_LLsdan} is the counterpart of Assumption \ref{assum_existence_LLs} in the context of adversarial domain adaptation networks, which is illustrated in Figure \ref{fig_illus_domain_ddan}. The assumption asserts that the source and the target distributions be related in such a way that, when efficiently aligned via the feature mappings $\fs$ and $\ft$ so as to minimize the domain discrepancy $\Ddan(\fs, \ft)$, the classification losses arising in the source and the target domains are also comparable. Note that the assumption is not limited to the ideal scenario where the domains are well-aligned: In  case of poor alignment,  $\Ddan(\fs, \ft)$ may be high, possibly leading to significantly different losses in the two domains. We, however, assume that the domain discriminator network is sufficiently well-trained; i.e., the learnt discriminator $\ddan $ is able to distinguish between the source and target domains if the mappings $\fs$ and $\ft$ result in poor feature alignment.

We can now state our main result about the sample complexity of adversarial domain adaptation networks.

\begin{theorem}
\label{thm_main_result_dann}

Consider a learning algorithm relying on the minimization of a loss function of the form \eqref{eq_ddan_obj_func} via an adversarial domain adaptation network. Assume that the classification loss function $\loss$  is bounded by a constant $\bls$ and Lipschitz continuous with respect to the first argument with constant $\Lls$.  Suppose that the source and target data distributions satisfy Assumption \ref{assum_existence_LLsdan} and the network parameters and activation functions  satisfy Assumptions \ref{assum_Ax_Atheta} and \ref{assum_bnd_act_val_op}- \ref{assum_bnd_act_val_op_ddan}. 

Let the feature dimensions be such that $\dl = O(\dcom)$ for $\lay=1, \dots, \numL$ and $\dlddan = O(\dcom)$ for $\lay=1, \dots, \numLdan$ for some common width parameter $\dcom$. Consider that the weight parameter $\alpha$ in the loss function is chosen such that
\begin{equation}
\label{eq_alpha_exrp_thm_ddan}
\begin{split}
\alpha
= O \left(
\left(
\frac
{\Mt \epsilon^2}
{\dcom^2 \numL \log \left( \frac{  \numL }{\epsilon} \right)  \, 
+ 
\dcom^2 \numL^2 \log(\dcom) }
\right)^{1/2}
\right)
\end{split}
\end{equation}
according to the number $\Mt$ of available labeled target samples. Then, in order to bound the expected target loss with a generalization gap of $O(\epsilon)$ as
\begin{equation}
\label{eq_accuracy_thm_ddan}
\begin{split}
\Lt(\ft, \h) \leq   \hLw(\fs, \ft, \h)+ (1-\alpha) \LLsdan  \hDdan(\fs, \ft) + (1-\alpha) \LLsdan \epsilon + \epsilon,
\end{split}
\end{equation}
the sample complexities in terms of the number $\Ms$ of labeled source samples, the number $\Ns$ of all (labeled and unlabeled) source samples, and the number $\Nt$  of all target samples are upper bounded by
\begin{equation*}
\begin{split}
\Ms &= O\left(
\frac{\dcom^2 \numL \log \left(\frac{\numL}{\epsilon}\right) 
+ \dcom^2 \numL^2 \log(\dcom) }
{\epsilon^2}
\right) \\
\Ns, \Nt &= 
O\left(
\frac{\dcom^2 (\numL+\numLdan) \log \left(\frac{\numL+ \numLdan}{\epsilon}\right) 
+ \dcom^2 (\numL+\numLdan)^2 \log(\dcom) }
{\epsilon^2}
\right).
\end{split}
\end{equation*}
\end{theorem}

The proof of Theorem \ref{thm_main_result_dann} is presented in Appendix \ref{pf_thm_main_result_dann}. The findings of Theorem  \ref{thm_main_result_dann} on the sample complexity of domain-adversarial networks are in line with those of Theorem \ref{thm_main_result_da_mmd}, which studied MMD-based networks. The optimal choice for the weight parameter $\alpha$ scales as $O(\sqrt{\Mt})$ as the number of labeled target samples varies, similarly to Theorem \ref{thm_main_result_da_mmd}. In order to prevent overfitting, $\Ms$ must increase at rate $\Ms=O(\dcom^2 \numL^2)$ with $\dcom$ and $\numL$, which indicates that the number of labeled source samples must increase quadratically with the width $\dcom$  and the depth $\numL$ of the feature extractor network, ignoring the logarithmic factors. Likewise, the number  of source and target samples $\Ns$ and $\Nt$  must also increase at a quadratic rate $O(\dcom^2 (\numL+\numLdan)^2)$ with the width $\dcom$ and the depth $\numL + \numLdan$ of the combination of feature extractor and domain discriminator networks, in order to avoid overfitting to the empirical domain discrimination loss of training samples. Similarly to the result in Theorem \ref{thm_main_result_da_mmd}, for the difference between the expected target loss and the sum of the empirical losses to be bounded by an amount of $O(\epsilon)$, the number of samples $\Ms, \Ns, \Nt$ must scale at rate $O(\epsilon^{-2})$.  

\begin{remark} 
In our analysis, we have considered the label predictor network to consist of a single layer as illustrated in Figure \ref{fig_illus_ddan_network}, as common practice in adversarial domain adaptation networks.  Nevertheless, it is straightforward to adapt our results to the case where the label predictor network consists of more than one layer. This is due to the fact that our analysis is based on the covering numbers of the function spaces $\Gs, \Gt$ and $\Vs, \Vt$, where $\N(\Gs, \epsilon, \ds)$,  $\N(\Gt, \epsilon, \dt)$  depend on only the total number of layers in the cascade of the feature extractor and the label predictor networks, and $\N(\Vs, \epsilon, \dVs)$,   $\N(\Vt, \epsilon, \dVt)$ depend only on the total number of layers in the cascade of the feature extractor and the domain discriminator networks.  Denoting the depth of the label predictor network as $P$ in this alternative setting, the resulting sample complexities would be obtained as  $\Ms=O(\dcom^2 (\numL+P)^2)$, and $\Ns, \Nt = O(\dcom^2 (\numL+\numLdan)^2)$. The optimal choice of the weight parameter $\alpha$ in \eqref{eq_alpha_exrp_thm_ddan} can similarly be obtained by replacing the number of layers $\numL$ with $\numL+P$ in this case.\\
\end{remark}

\section{Discussion of the results in relation with previous literature}
\label{sec_rel_work}

We now discuss our findings in relation with previous literature. To the best of our knowledge, our study is the first to propose an in-depth characterization of the sample complexity of domain-adaptive neural networks. A substantial body of work has focused on the effect of domain discrepancy on generalization performance, while another line of research has examined the sample complexity of neural networks, however, in a single-domain setting. We briefly overview these results below, along with a few relevant studies on the performance of domain alignment methods.  For clarity and consistency, we restate the findings of prior work using our own notation. The presence of the parameter $\delta$ in the bounds signifies that the result holds with probability at least $1-\delta$.

\subsection{Effect of domain discrepancy on generalization performance}

One of the earliest analyses examining the effect of the deviation between the source and target distributions is the study by Ben-David et al.~\cite{BenDavidBCP06}. The gap between the expected target loss and the empirical source loss is shown to be bounded by
\[
O\left(  
\sqrt{ \frac{\text{dim}_{VC}(\mathcal{H})}{\Ms} + \log(\delta^{-1})} \right) + d_{\mathcal{H}}(D_S, D_T) + \lambda
\]
ignoring the logarithmic factors, where $\text{dim}_{VC}(\mathcal{H})$ denotes the VC-dimension of the hypothesis space $\mathcal{H}$, $\Ms$ is the number of of labeled source samples, and $\lambda$ is a  measure of  the proximity of the true label function to the hypothesis class $\mathcal{H}$. Here  $d_{\mathcal{H}}(D_S, D_T)$ is the $\mathcal{A}$-distance \cite{BenDavidBCP06} between the source and target distributions $D_S$ and $D_T$, given by
\[
 d_{\mathcal{H}}(D_S, D_T) = 2 \sup_{A \in \mathcal{A}} | P_{D_S}(A) - P_{D_T}(A) |
\]
where $\mathcal{A}$ is the set of domain subsets with characteristic functions in $\mathcal{H}$, and $P_{(\cdot)}$ denotes probability with respect to a distribution.

In a succeeding study \cite{BenDavidBCKPV10}, this result has been extended to algorithms minimizing a convex combination of source and target losses, where the hypothesis that minimizes the empirical weighted loss is shown to generalize to the target domain within an error of 
\begin{equation*}
\begin{split}
O\bigg(
\sqrt{\frac{\alpha^2}{\gamma} + \frac{(1-\alpha)^2}{1-\gamma}} 
& \sqrt{  \frac{\text{dim}_{VC}(\mathcal{H}) + \log(\delta^{-1}) }{M} } \\
&+ 
(1-\alpha)
\bigg(\sqrt{\frac{\text{dim}_{VC}(\mathcal{H}) \log(\delta^{-1})}{N} }
+
\hat d_{\mathcal{H}\Delta \mathcal{H}}(D_S, D_T) 
+ \lambda
\bigg)
\bigg).
\end{split}
\end{equation*}
Here the distribution distance $\hat d_{\mathcal{H}\Delta \mathcal{H}}(D_S, D_T) $ denotes the  empirical divergence between the source and the target distributions over the symmetric difference hypothesis space $\mathcal{H}\Delta \mathcal{H}$, which corresponds to the set of disagreements \cite{BenDavidBCKPV10}. $N=\Ns=\Nt$ denotes the number of all samples in the two domains, and $M$ is the total number of labeled samples, with $\Ms=(1-\gamma)M$ source samples and $\Mt=\gamma M$ target samples. This result has some implications paralel to our study, in that the optimal weight $\alpha$ of the target loss should decrease with the scarcity of target labels, i.e., as $\gamma$ decreases. A high domain discrepancy  $\hat d_{\mathcal{H}\Delta \mathcal{H}}(D_S, D_T) $ also drives the weighted loss towards the target loss, by decreasing the weight $1-\alpha$ of the source loss. 

Similar findings have been presented in the study of Mansour et al.~in terms of the Rademacher complexities of the hypothesis space \cite{MansourMR09}. However, in \cite{MansourMR09} the deviation between the source and the target domains has been characterized in terms of the discrepancy $\text{disc}_{\loss} (D_S, D_T)$, which quantifies how the loss-induced disagreement between any pair of hypotheses may differ across $D_S$ and $D_T$.

Following these pioneering works, many other domain divergence measures have been proposed in succeeding studies \cite{RedkoMHS20}. Deng et al.~have explored a robust variant of the discrepancy in  \cite{MansourMR09} based on the adversarial Rademacher complexity definition  \cite{DengGHML23}, which has been shown to vary with the number of samples $M$ and the network width $d$ at rate $O(\sqrt{d/M})$ for two-layer ReLU neural networks. Zhang et al.~have proposed an alternative characterization of distribution distance based on the margin disparity discrepancy, leading to generalization bounds in terms of the Rademacher complexities and the covering numbers of hypothesis spaces \cite{ZhangLLJ19}. Zellinger et al.~have presented performance bounds depending on the VC-dimension of the function classes by formulating the domain discrepancy in terms of the difference between the moments of the source and target distributions \cite{ZellingerMS21}. Other recent efforts along this line include studies involving margin-aware risks with links to optimal transport distances \cite{DhouibRL20}, information-theoretic bounds based on mutual information \cite{WangM23, WuMAZ24}, hypothesis-specific divergence measures \cite{WangM24}, and risk definitions based on stochastic predictors \cite{SiciliaAAH22}.

\begin{remark} 
We note that all these aforementioned works assume that a common classifier is learnt in the original source and target domains; i.e., their setting is essentially different from ours as they do not at all consider learning a transformation or a mapping that aligns the two domains. The main distinction among these works lies in the specific distribution discrepancy each one proposes to characterize the misalignment between the domains, with the purpose of deriving tighter error bounds.  Meanwhile, the reported labeled and unlabeled sample complexities, or otherwise the errors, follow the classical dependence on the VC-dimensions or the Rademacher complexities of the hypothesis classes in consideration, consistent with well-established results in learning theory. From the perspective of domain alignment algorithms, one may want to regard the domain discrepancies in these bounds as the distance obtained after mapping the two domains to a shared domain, an interpretation that arguably extends to transformation learning. While this view holds to some extent, many of the discrepancy measures used in these works (including their empirical approximations) are defined in a theoretical manner, and are difficult to estimate in practice. Although efficient computational techniques may exist for some of these discrepancy measures, they often lack accompanying learning guarantees. In contrast, our main results in Theorems \ref{thm:main_result_mmd}-\ref{thm_main_result_dann} offer a practical means of assessing the generalization capability of domain alignment algorithms, as they are based on the empirical distribution distance computed directly on the aligned training data.
\end{remark}

\subsection{Performance bounds for domain alignment algorithms}

To the best of our knowledge,  a very limited number of theoretical analyses have investigated the performance of learning domain-aligning transformations or representations. A multi-task domain adaptation method is proposed in \cite{ZhouTPT19}, which learns the similarity between source and target samples through a linear transformation $\mathbf{G}$. Assuming the incoherence of the projections corresponding to different tasks, the estimation error of the transformation $\mathbf{G}$ is shown to be bounded by $O(d_T \sqrt{\log(d_S)/n})$, where $d_S$ and $d_T$ denote the dimensions of the source and target Euclidean domains, and $n$ is the number of tasks. While this bound is subsequently leveraged in \cite{ZhouTPT19} to design suitable classifiers based on the incoherence principle, the scope of their analysis is limited to linear transformations. 

A performance analysis of conditional distribution matching is presented in \cite{WangS15}, showing that  the generalization gap in the target domain is bounded by
\[
O\left (  1+ \frac{1}{\sqrt{\Mt}} + \sqrt{\frac{\log(\delta^{-1})}{\Ms + \Mt}} \right)
\]
when the source domain is mapped to the target domain through a location and scale transform. 

Fang et al.~have considered semi-supervised domain alignment algorithms  as in our work  \cite{FangLLZ23}. However, their analysis is significantly different from ours since it does not explore the sample complexity of learning domain transformations, but instead treats the sample complexity as a known problem parameter. Their study aims to demonstrate that the need for labeled target data can be alleviated under certain assumptions by relying on the source and unlabeled target data. 

Transferring representations from a source task to a target task is a problem different from but connected to domain adaptation. Wang et al.~have provided an extensive analysis of  transfer learning and multitask learning through domain-invariant feature representations by minimizing a combined empirical loss under regularization \cite{WangMSX23}. The performance gap between the source and target losses is shown to vary at rate
\[
O\left(\text{dist}_{\mathcal{Y}}(\fs, \ft) + \sqrt{ \frac{\log(\delta^{-1})}{\Ms + \Mt}} \right).
\]
Here $\text{dist}_{\mathcal{Y}}(\fs, \ft)$ denotes the $\mathcal{Y}$-discrepancy \cite{MohriM12} between the two domains once transformed to a shared domain, which is, however, not easy to estimate in practice. 

Galanti et al.~have modeled the transfer learning problem in a setting where a target task and multiple source tasks are drawn from the same distribution of distributions, and considered that a neural network architecture is partially transferred to the target task \cite{TomerWH16}. Their analysis implies that for accurate transfer, the number of source tasks and the number of samples  per source task must scale with the number of edges, respectively, in the transferred component and the target-specific component of the network. In a recent work, Jiao et al.~have considered a model that distinguishes between shared and domain-specific features in multi-domain deep transfer learning and shown that transferability between tasks improves the convergence rates in the target task \cite{JiaoLLY24}. McNamara and Balcan have investigated representation learning on a source task and fine-tuning on a target task  \cite{McNamaraB17}. The accuracy on the source task is shown to carry over to the target task within a performance gap of $O(\sqrt{\text{dim}_{VC}(\mathcal{H \circ F}) / \Ms} + \sqrt{\text{dim}_{VC}(\mathcal{H})/\Mt} )$, where $\mathcal{F}$ is the space of feature representations and $\mathcal{H}$ is the space of classifiers. The significance of this result lies in the fact that the number $\Mt$ of labeled target samples should scale with the dimension of only the classifier $\mathcal{H}$, rather than the more complex composite hypothesis space $\mathcal{H \circ F}$. A paralel finding is presented in \cite{TripuraneniJJ20} for the problem of transfer learning in a multi-task setting, demonstrating that the number of labeled samples for a new task needs to scale only with the complexity of its own task-specific map, assuming the abundance of the training data for the previous tasks. 

\begin{remark}
Although our domain adaptation setting differs essentially from that considered in these transfer learning studies, they are comparable in their shared focus on handling the scarcity of labeled target samples. Whereas these works tie sample complexity to the richness of the target function class, which can be still large for deep neural networks, our analysis indicates that in a domain adaptation scenario the limitedness of target labels can be tolerated through strategically choosing the weight parameter as $\alpha=O(\sqrt{\Mt})$,  independently of the complexity of  the target function class.
\end{remark}

\subsection{Sample complexity of neural networks in a single domain}

Sample complexity of neural networks is a well-explored topic in statistical learning theory, a comprehensive overview of which can be found in \cite{AnthonyB02}, \cite{BartlettMR21}. Although this classical line of research pertains to learning algorithms in a single domain and does not extend to domain adaptation scenarios, we find it instructive to briefly review these results and compare them to our bounds on domain adaptive neural networks. 

The sample complexity of a feed-forward network consisting of $W$ weights, $\numL$ layers and $s$ output units, with fixed piecewise-polynomial activation functions is reported as \cite[Theorem 21.5]{AnthonyB02}
\begin{equation}
\label{eq_samp_comp_Bartlett_NN}
\begin{split}
O\left(  \frac{ s (W \numL \log(W) + W \numL^2) \log(\epsilon^{-1}) + \log(\delta^{-1})   }{\epsilon^2}   \right)
\end{split}
\end{equation}
in order to attain an error of $\epsilon$.  Denoting the network width as $\dcom$, the number of weights $W$ in an $\numL$-layer network is  obtained as $W= \dcom^2 \numL$. Then, the  sample complexity $M=O(\dcom^2 \numL^3)$ in \eqref{eq_samp_comp_Bartlett_NN} points to a quadratic dependence on $\dcom$ and a cubic dependence on $\numL$.  This polynomial dependence is in line with our results in Theorems \ref{thm_main_result_da_mmd} and \ref{thm_main_result_dann}, where the sample complexity of labeled source data has been obtained as $\Ms= O(\dcom^2 \numL^2)$. The dependence on $\numL$ is quadratic, hence slightly tighter in our bounds. 

A more recent trend in the exploration of sample complexity of neural networks is the  characterization of the complexity in a dimension-independent way under particular assumptions. Neyshabur et al.~have shown that the sample complexity depends exponentially on the network depth; nevertheless, its dependence on the network width can be removed under group norm regularization of network weights \cite{NeyshaburTS15}. In succeeding studies, the exponential dependence on the network size has been reduced to polynomial \cite{WeiM19}, quadratic \cite{NeyshaburBS18}, linear \cite{GolowichRS18} and logarithmic \cite{BartlettFT17} factors. Harvey et al.~have shown that the VC-dimension of neural networks with ReLU activation functions is $O(W \numL \log(W))$, resulting in comparable bounds to our work \cite{HarveyLM17}.  In some more recent works, it has been shown that the dependence on network width can be removed for one-layer networks \cite{VardiSS22} and reduced to logarithmic factors for two-layer networks \cite{DanielyG24} under bounded Frobenius norm and spectral norm constraints. We note that these results essentially rely on the condition that the norms of the weight matrices be upper bounded in a dimension-independent manner, and would translate to rather pessimistic sample complexities under the removal of this assumption.

\begin{remark} 
While the above studies have contributed to a comprehensive understanding of neural network classifiers, they all focus on the single-domain scenario, assuming identical distributions for training and test data.  To the best of our knowledge, our work is the first to provide a detailed analysis of the sample complexity of domain-adaptive neural networks. We note that our analysis does not impose any special constraints on the weight matrices, such as norm regularization. Under the incorporation of norm constraints, we would expect to arrive at tighter bounds consistently with the approaches in single-domain settings, which is left as a potential future direction of our study.
\end{remark}

\section{Experimental results}
\label{sec:exp_results}

In this section, we present experimental results for the verification of the proposed generalization bounds. In Section \ref{ssec_exp_gen_da}, we study the generic bounds presented in Section \ref{sec:gen_bounds} by considering a shallow (linear) classifier model. Then in Section \ref{ssec_exp_dan}, we examine the sample complexity results proposed in Section \ref{sec_samp_dan}  for domain-adaptive neural networks.

\subsection{General domain alignment methods} 
\label{ssec_exp_gen_da}

We first validate our findings in Section \ref{sec:gen_bounds} on a synthetic data set with two classes. The source and target data sets are generated by applying two different geometric transformations to 400 samples drawn from the standard normal distribution in $\R^2$. We simulate a learning algorithm that learns geometric transformations to map the source and target samples to a common domain and then trains a classifier in the shared domain. Here we emulate a setting where the transformations $\fs$ and $\ft$ are treated as if learnt from data, however, with some error. In practice,  $\fs$ and $\ft$ are formed by perturbing the ground truth geometric transformations with some transformation estimation error $\tau$.  We test a range of estimation error levels  $\tau$ in the experiments. The classifier trained after mapping the samples to the common domain is chosen as a regularized ridge regression algorithm solving
\begin{equation*}
\min_{\wvect \in \R^2} \ \  \frac{1-\alpha}{\Ms} \sum_{i=1}^{\Ms} (\wvect ^T \fs(\xis) - \yis)^2 
+ \frac{\alpha}{\Mt} \sum_{j=1}^{\Mt} (\wvect ^T \ft(\xjt) - \yjt)^2  + \lambda \| \wvect \|^2 .
\end{equation*}
The target misclassification rate is evaluated over 1000 test samples drawn from the target distribution and classified through the learnt hypothesis $\wvect$ and target transformation $\ft$.

\begin{figure}[t]
\begin{center}
     \subfigure[]
       {\label{fig:toy_error_Mt}\includegraphics[height=4.5cm]{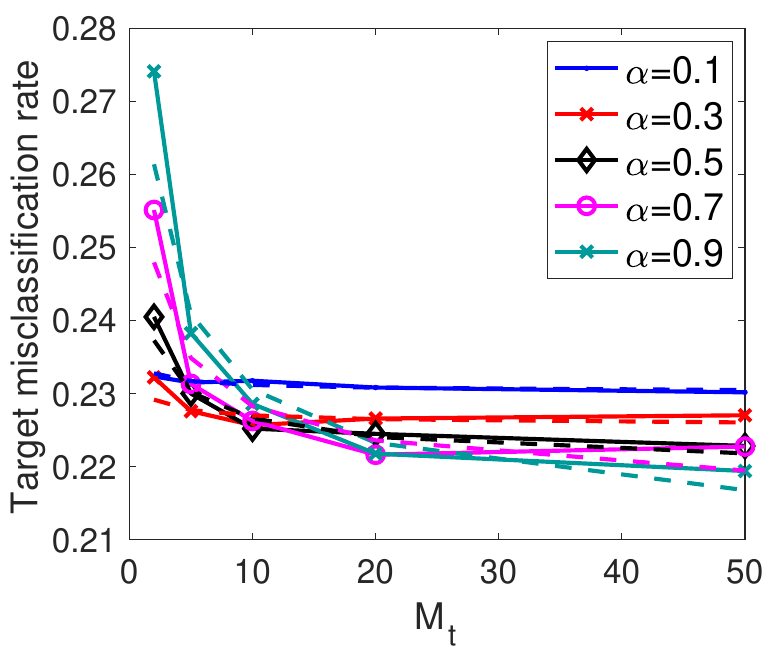}}
     \subfigure[]
       {\label{fig:toy_error_tau}\includegraphics[height=4.5cm]{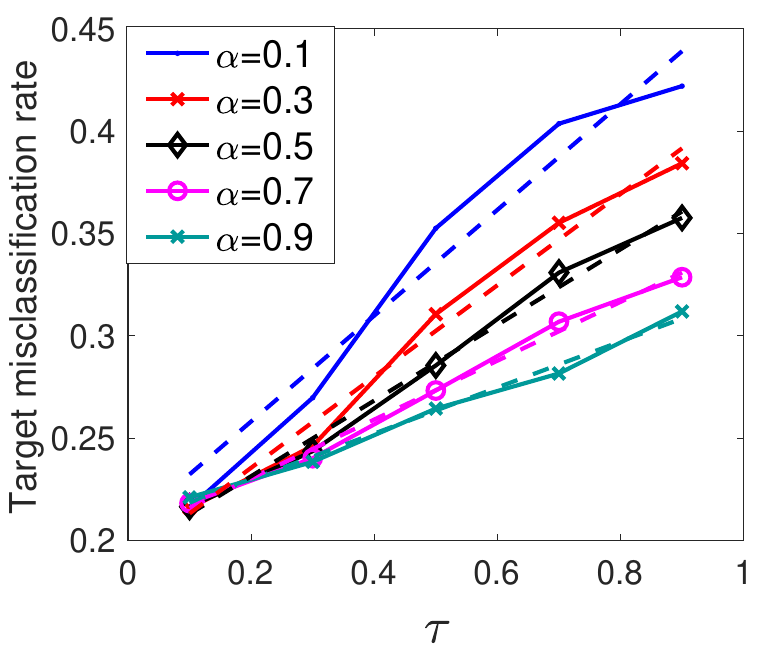}}
 \end{center}
 \vspace{-0.5cm}
 \caption{Variation of the target error on synthetical data with (a) Number of labeled target samples, (b) Distribution distance after transformation. Solid lines indicate experimental data and dashed lines represent theoretical rates of variation.}
 \label{fig:toy_error}
\end{figure}

In Figure \ref{fig:toy_error_Mt}, the variation of the target misclassification rate with the number $\Mt$ of labeled target samples is shown for different values of the weight $\alpha$ for the target loss. In order to interpret these results, it is helpful to recall our theoretical analysis in Section \ref{sec:gen_bounds}: Theorem \ref{thm:gen_defect_target} states that the expected target loss $\Lt(\ft, \h) $ deviates from its reference value based on the empirical weighted loss $\hLw(\fs, \ft, \h)$ and the distance $\D(\fs, \ft)$ by an amount of $\epsilon$. In order to achieve this with high and fixed probability, the term $\Mt \epsilon^2$ in the probability expression \eqref{eq_prob_expr_thm1} must be constant\footnote{We ignore logarithmic factors and assume that the generic covering numbers in Theorem \ref{thm:gen_defect_target} grow at a typical geometric rate of increase as the covering radius decreases.}. This implies that the expected target loss should decrease at rate $\epsilon = O(\sqrt{1/\Mt}) $ as $\Mt$ increases. Considering the target misclassification rate as an accurate approximation of the expected loss $\Lt(\ft, \h) $ in Figure \ref{fig:toy_error_Mt}, we observe that the decay in the target error with  $\Mt$ is consistent with Theorem \ref{thm:gen_defect_target}. In particular, the dashed lines in the plots correspond to fitted theoretical rates of decay  $O(\sqrt{1/\Mt})$, which closely match the experimental data.  We can also observe that large $\Mt$ values favor larger $\alpha$ values, while $\alpha$ must be chosen smaller at small $\Mt$ values. This also aligns with the conclusion drawn from Theorem \ref{thm:gen_defect_target} that the parameter $\alpha$ must be chosen as  $\alpha = O(\sqrt{\Mt})$ in order to control the term $e^{-\frac{\Mt \epsilon^2}{8 \alpha^2 \bls^2}}$ as $\Mt$ decreases.

We then study in Figure \ref{fig:toy_error_tau} the variation of the target misclassification rate with the estimation error $\tau$ of the geometric transformations. The parameter $\tau$ here is taken as the norm of the error matrix that is added to the ground truth transformation matrix. Hence, $\tau$ can be regarded as a parameter proportional to the distribution distance  $\D(\fs, \ft)$. The misclassification rate tends to increase with $\tau$ at an approximately linear rate, as confirmed by the dashed lines representing the theoretical linear rate of increase fitted to the experimental data. These results are coherent with the prediction of Theorem \ref{thm:gen_defect_target} that the expected target loss should increase proportionally to the distribution distance $\D(\fs, \ft)$.

\begin{figure}[t]
  \centering
  \centerline{\includegraphics[width=7.0cm]{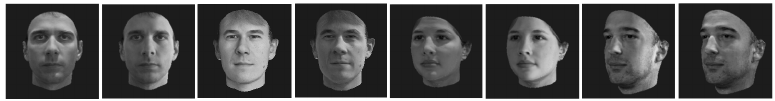}}
  \caption{Sample images from the MIT-CBCL face data set for four different subjects, rendered respectively under poses 1, 2, 5, and 9 for various illumination conditions. }
  \label{fig:face_dataset}
\end{figure}

\begin{figure}[t]
\begin{center}
     \subfigure[]
       {\label{fig:mit_error_Mt}\includegraphics[height=4cm]{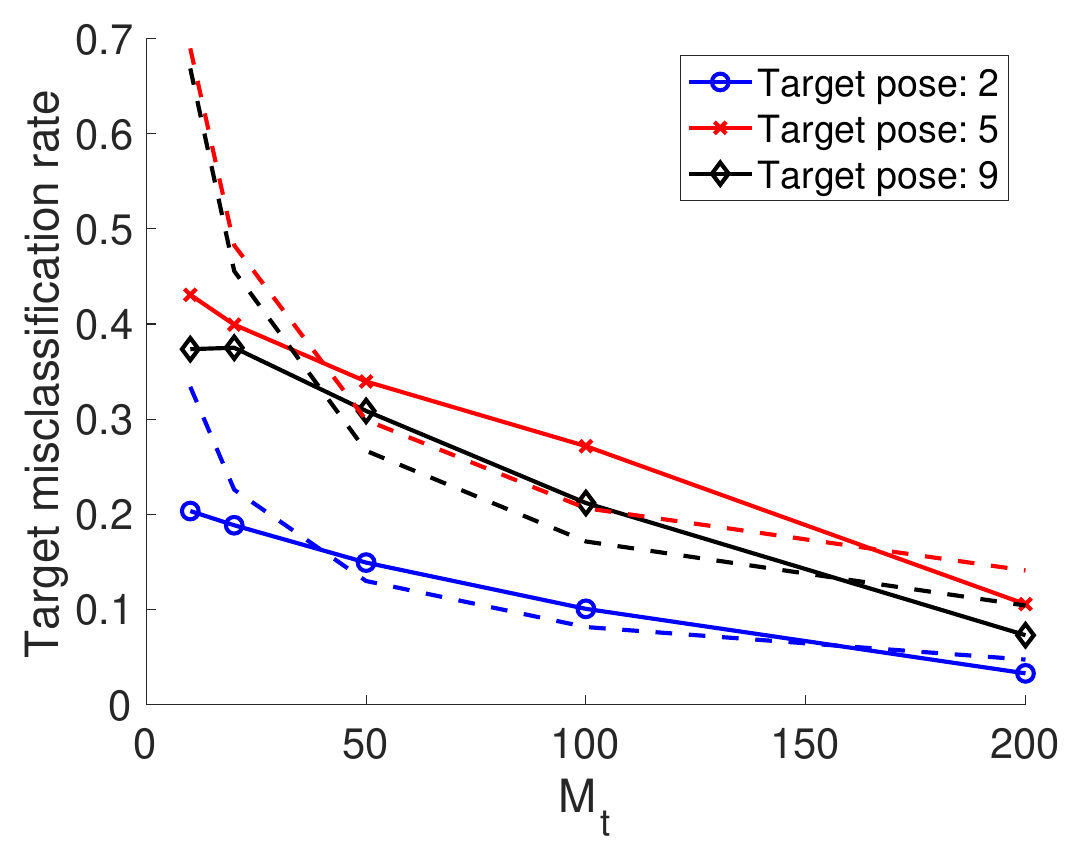}}
     \subfigure[]
       {\label{fig:mit_error_Ms}\includegraphics[height=4cm]{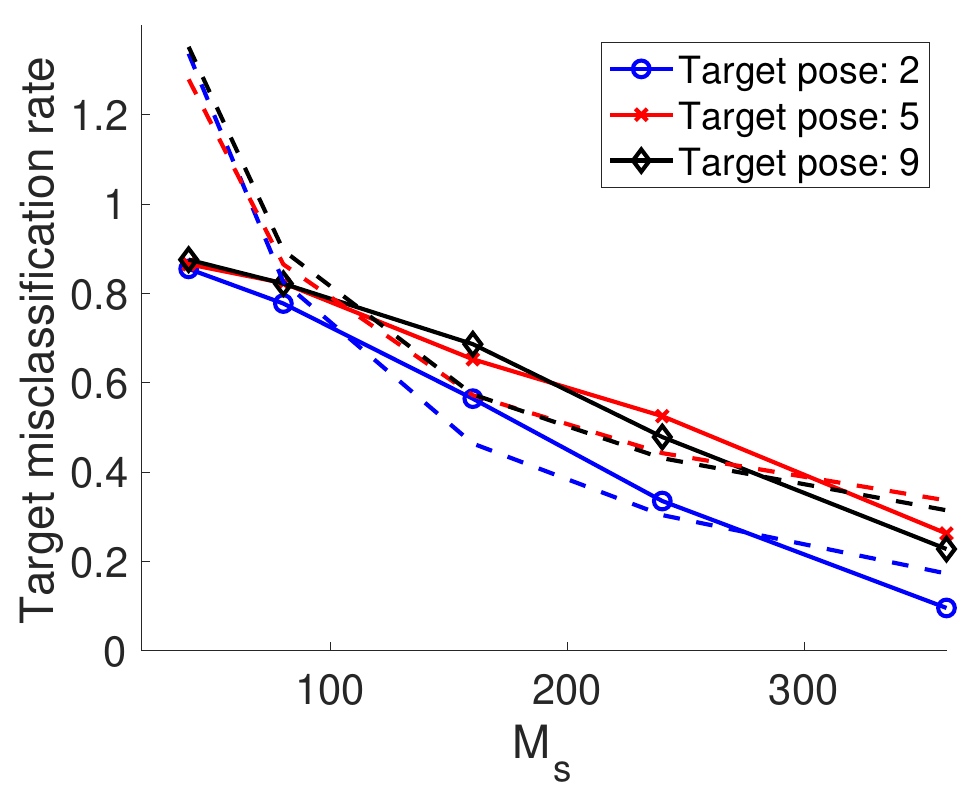}}
 \end{center}
 \vspace{-0.5cm}
 \caption{Variation of the target error on MIT-CBCL face data with (a) Number of labeled target samples, (b) Number of labeled source samples. Solid lines indicate experimental data and dashed lines represent theoretical rates of variation.}
 \label{fig:mit_error}
\end{figure}

Next, we experiment on the MIT-CBCL image data set \cite{MITCBCL}. The data set consists of a total of 3240 synthetic face images belonging to 10 subjects. The images of each subject are rendered under 36 different illumination conditions and 9 poses, with Pose 1 corresponding to the frontal view and Pose 9 corresponding to a nearly profile view. Some example images from Poses 1, 2, 5, 9 are shown in Figure \ref{fig:face_dataset}. We consider the images rendered under Pose 1 as the source domain, and repeat experiments by taking images from Poses 2, 5 and 9 as the target domain in each trial. First, using all labeled and unlabeled images, we compute a mapping between the source and target domains by the method proposed in \cite{FernandoHST13}, which finds a transformation that aligns the PCA bases of the source and target domains. We then train an SVM classifier using all labeled samples from the two domains. The unlabeled target samples are finally classified with the learnt transformation and classifier.

The misclassification rates of unlabeled target samples are plotted in Figures \ref{fig:mit_error_Mt} and \ref{fig:mit_error_Ms},  with respect to the number of labeled target and source samples respectively. We observe that in both figures, the misclassification rates are reduced effectively with the increase in the number of labeled samples. 
As previously discussed, the target loss is expected to asymptotically reduce to an error component resulting from the empirical loss and the distribution distance, at rates $O(\sqrt{1/\Mt})$ and $O(\sqrt{1/\Ms})$ with increasing $\Mt$ and $\Ms$. The experimental results in Figures \ref{fig:mit_error_Mt} and \ref{fig:mit_error_Ms} seem consistent with this expectation.  The theoretical curves fitted to the experimental data with the expected rates of decrease are also indicated with dashed lines in the plots for visual comparison.

\subsection{Domain-adaptive neural networks} 
\label{ssec_exp_dan}

We next aim to experimentally verify our results in Theorems \ref{thm_main_result_da_mmd} and \ref{thm_main_result_dann} regarding the sample complexity of domain-adaptive neural networks. We present our results for MMD-based and adversarial domain adaptation networks, respectively in Section \ref{ssec_exp_mmd_dan} and  Section \ref{exp_adv_da}. For both  architectures, our purpose is to experimentally characterize the sample complexity of the network with respect to the depth $\numL$ and the width $\dcom$ of the network. We additionally investigate the optimal value of the weight $\alpha$ of the target loss in the objective function for both cases.

In our experiments, the MNIST handwritten digit data set \cite{mnist} is used as the source data set, which consists of 60000 images. The target data set is taken as MNIST-M \cite{mnistm}, which contains 59000 handwritten digit images with colored backgrounds. We train the neural networks with labeled and unlabeled training samples from the source and target domains, and then evaluate the target accuracy of the learnt models, defined as the correct classification rate of test samples from the target domain. In all experiments, algorithm hyperparameters and fixed variables are chosen  to keep the neural network in the overfitting regime, enabling the characterization of the sample complexity of the models under consideration.

\subsubsection{MMD-based domain adaptation networks}
\label{ssec_exp_mmd_dan}

\begin{figure}[t]
\begin{center}
     \subfigure[]
       {\label{fig_mmd-layer-Ms-linefit}\includegraphics[height=4.5cm]{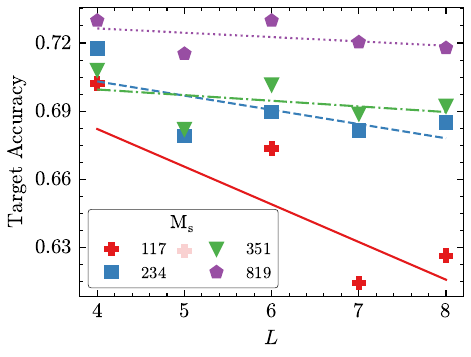}}
     \subfigure[]
       {\label{fig_mmd-layer-Ms-quadprog}\includegraphics[height=4.5cm]{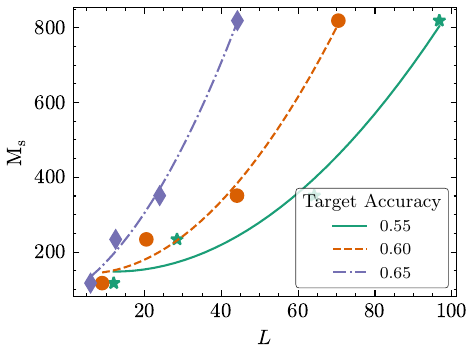}}
            \subfigure[]
       {\label{fig_mmd-layer-Ns-linefit}\includegraphics[height=4.5cm]{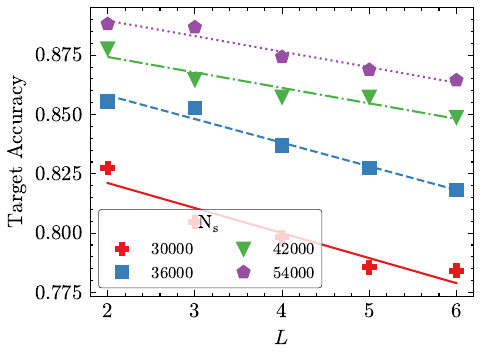}}
     \subfigure[]
       {\label{fig_mmd-layer-Ns-quadprog}\includegraphics[height=4.5cm]{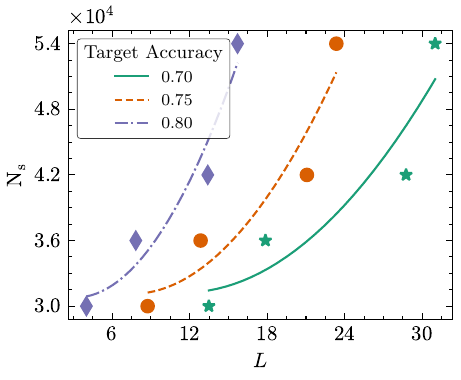}}
\end{center}
 \vspace{-0.5cm}
 \caption{Sample complexity of labeled samples ($\Ms$) and all samples ($\Ns$) with respect to the depth $\numL$ of MMD-based domain adaptation networks. Left panels (a),(c): Variation of target accuracy with $\numL$. Right panels (b),(d): Variation of the number of samples ($\Ms, \Ns$) required for attaining a desired target accuracy level with $\numL$.}
 \label{fig_mmd_results_L}
\end{figure}

\begin{figure}[t!]
\begin{center}
     \subfigure[]
       {\label{fig_mmd-dim-Ms-linefit}\includegraphics[height=4.5cm]{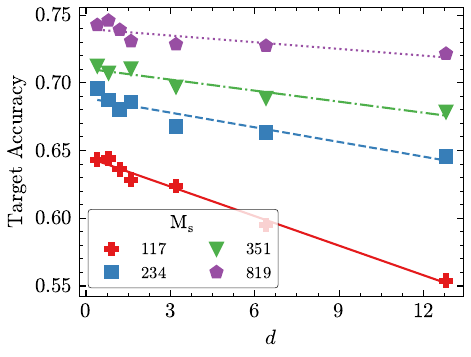}}
     \subfigure[]
       {\label{fig_mmd-dim-Ms-quadprog}\includegraphics[height=4.5cm]{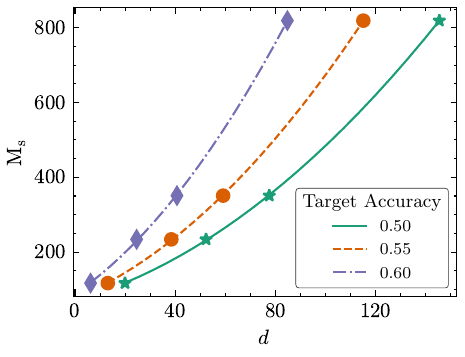}}
 \end{center}
 \vspace{-0.5cm}
 \caption{Sample complexity of labeled samples ($\Ms$) with respect to the width $\dcom$ of MMD-based domain adaptation networks. (a) Variation of target accuracy with $\dcom$. (b) Variation of the number of samples ($\Ms$) required for attaining a desired target accuracy level with $\dcom$.}
 \label{fig_mmd_results_d}
\end{figure}

In our analysis of MMD-based domain adaptation networks, we consider the architecture proposed in the pioneering study \cite{LongCWJ15} as our benchmark. We build on our previous experimental study \cite{KaracaAAAAUV23} and employ a neural network structure similar to the baseline model in  \cite{LongCWJ15}, beginning with convolutional layers and followed by several fully connected MMD layers. The MMD layer parameters are coupled between the source and target domains. The dimensions (widths) of all MMD layers are set as equal. Batch normalization is applied after each layer in order to stabilize the performance. We use the PyTorch implementation of the network available in \cite{mmd_code_pytorch} and adapt it for the minimization of the objective function 
\begin{equation} 
 \frac{1-\alpha}{\Ms} \sum_{i=1}^{\Ms}  \loss( \h \circ f(\xis), \yis )  
+ \frac{\alpha}{\Mt} \sum_{i=1}^{\Mt} \loss( \h \circ f(\xjt),  \yjt  ) 
 +  \beta \sum_{\lay=1}^{\numL-1}  (\hD^\lay)^2 (f^l, f^l)
\end{equation} 
where $\loss(\cdot, \cdot)$ is set as the cross-entropy loss function and the source and target feature transformations are coupled as $\fs=\ft=f$ and $\fsl=\ftl=f^l$.

In Figure \ref{fig_mmd_results_L}, we study the sample complexity of labeled source samples $\Ms$ and all source samples $\Ns$ with respect to the number  $\numL$ of MMD layers in the network. Figures \ref{fig_mmd-layer-Ms-linefit} and \ref{fig_mmd-layer-Ns-linefit} show the decrease in the target accuracy as the number  $\numL$ of MMD layers increases when the network is in the overfitting regime, for different $\Ms$ and $\Ns$ values. We aim to characterize the sample complexity of $\Ms$ and $\Ns$ with respect to $\numL$ in this experiment. Therefore, we determine several desired target accuracy levels for the results in  Figures \ref{fig_mmd-layer-Ms-linefit} and \ref{fig_mmd-layer-Ns-linefit}, and identify the smallest $\Ms$ and $\Ns$ values that ensure this target accuracy as $\numL$ grows\footnote{In cases where obtaining the exact value of $\numL$ exceeded our computational resources, we resorted to linear extrapolation of the curves in Figures \ref{fig_mmd-layer-Ms-linefit} and \ref{fig_mmd-layer-Ns-linefit} to approximately infer the corresponding $\numL$ value.}, which are plotted respectively in Figures \ref{fig_mmd-layer-Ms-quadprog} and \ref{fig_mmd-layer-Ns-quadprog}. We recall from Theorem \ref{thm_main_result_da_mmd} that the sample complexities of $\Ms$ and $\Ns$ are expected to grow at quadratic rates $\Ms=O(\numL^2)$ and $\Ns=O(\numL^2)$ as the network depth $\numL$ increases. The experimental findings in Figures \ref{fig_mmd-layer-Ms-quadprog} and \ref{fig_mmd-layer-Ns-quadprog} confirm this prediction, as the increase in the required sample size for attaining a reference target accuracy level indeed follows a quadratic increase with $\numL$. The curves in  \ref{fig_mmd-layer-Ms-quadprog} and \ref{fig_mmd-layer-Ns-quadprog} are obtained by fitting quadratic polynomials to the experimental data for visual evaluation. 

A similar experiment is conducted in Figure \ref{fig_mmd_results_d}, where the sample complexity is studied with respect to the network width this time. The parameter $\dcom$  in Figures \ref{fig_mmd-dim-Ms-linefit} and \ref{fig_mmd-dim-Ms-quadprog} represent the factor by which the network width in the original implementation  \cite{mmd_code_pytorch} is multiplied in our experiment. Hence, $\dcom$ is directly proportional to the shared width parameter of the MMD layers. The results in \ref{fig_mmd-dim-Ms-quadprog} are also consistent with the theoretical findings in Theorem \ref{thm_main_result_da_mmd}, which states that the sample complexity must increase at a quadratic rate $\Ms=O(\dcom^2)$ as the network width increases.

\begin{figure}[ht]
\begin{center}
     \subfigure[]
       {\label{fig_mmd-optalpha_quadfit_Ms234}\includegraphics[height=4.5cm]{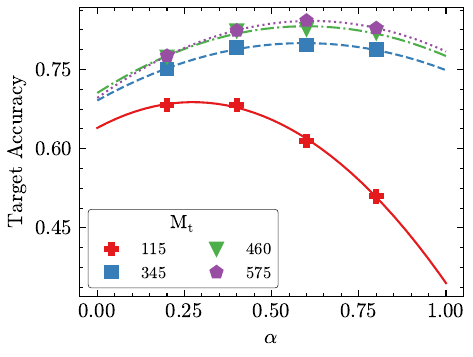}}
     \subfigure[]
       {\label{fig_mmd-optalpha-asqrtx_fit_den_Mss}\includegraphics[height=4.5cm]{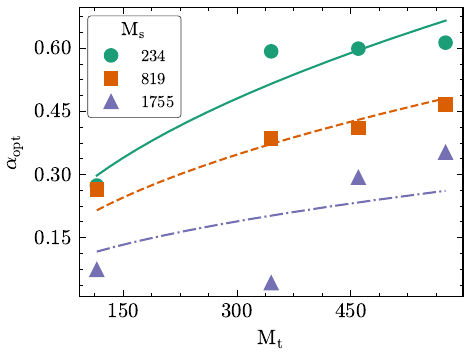}}
 \end{center}
 \caption{(a) Variation of target accuracy with target loss weight parameter $\alpha$ for MMD-based domain adaptation networks (obtained at $\Ms = 234$). (b) Variation of optimal weight $\alpha_{opt}$ with number of labeled target samples $\Mt$.}
 \label{fig_mmd_results_alpha}
\end{figure}

We also recall from Theorem \ref{thm_main_result_da_mmd} that, in order to maximize the target accuracy, the weight parameter $\alpha$ of the target classification loss must scale as $\alpha=O(\sqrt{\Mt})$ as the number $\Mt$ of labeled target samples varies. We experimentally validate this result in Figure \ref{fig_mmd_results_alpha}. In Figure \ref{fig_mmd-optalpha_quadfit_Ms234}, we examine the variation of the target accuracy with the weight parameter $\alpha$. Here, the target accuracy follows a non-monotonic variation with $\alpha$ as expected. We approximately identify the optimal value $\alpha_{opt}$ of the weight parameter for each value of $\Mt$ by applying polynomial fitting to the plots in Figure \ref{fig_mmd-optalpha_quadfit_Ms234}.  The variation of the optimal weight  $\alpha_{opt}$ with $\Mt$ is then plotted in Figure \ref{fig_mmd-optalpha-asqrtx_fit_den_Mss}. In order to visually observe the prediction of Theorem  \ref{thm_main_result_da_mmd}, we also fit a curve of $O(\sqrt{\Mt})$ to each data sequence in Figure \ref{fig_mmd-optalpha-asqrtx_fit_den_Mss}. The experimental data in Figure  \ref{fig_mmd-optalpha-asqrtx_fit_den_Mss} seems consistent with the fitted  curves, which supports the statement of Theorem \ref{thm_main_result_da_mmd} that the optimal weight parameter must scale at rate $\alpha_{opt}=O(\sqrt{\Mt})$.

\subsubsection{Adversarial domain adaptation networks}
\label{exp_adv_da}

In order to experimentally evaluate our findings in Section \ref{ssec_adv_da_net}, we adopt the model proposed in \cite{GaninUAGLLML16}, which is a well-known representative of adversarial domain adaptation architectures. We use the PyTorch implementation of this model available in \cite{fungtion_DANN_py3}, by adapting it to the semi-supervised setting studied in our analysis. We train the adversarial network to minimize the objective function
\begin{equation*} 
\begin{split}
&  \frac{1-\alpha}{\Ms} \sum_{i=1}^{\Ms}  \loss( \h \circ f(\xis), \yis )  
+ \frac{\alpha}{\Mt} \sum_{i=1}^{\Mt} \loss( \h \circ f(\xjt),  \yjt  ) \\
& - 
\frac{\beta}{N_s + N_t}
\left(
 \sum_{i=1}^{\Ns}  \loss_\dom( \ddan \circ f(\xis), \ydoms_i )
 +
  \sum_{j=1}^{\Nt} \loss_\dom( \ddan \circ f(\xjt), \ydomt_j  ) 
  \right)
  \end{split}
\end{equation*} 
where the label loss $\loss(\cdot, \cdot)$ and the domain discrimator loss $\loss_\dom(\cdot, \cdot)$ are selected as the negative log likelihood function, and the source and target feature extractor networks are coupled as $\fs=\ft=f$.

\begin{figure}[ht]
\begin{center}
     \subfigure[]
       {\label{fig_adv-layer-Ms-linefit}\includegraphics[height=4.5cm]{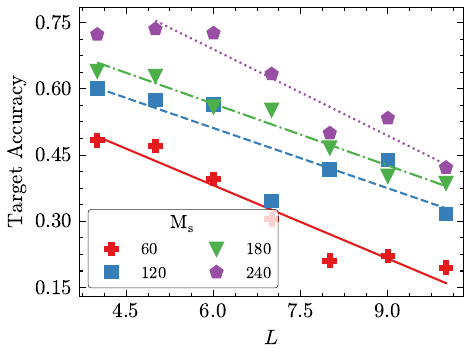}}
     \subfigure[]
       {\label{fig_adv-layer-Ms-quadprog}\includegraphics[height=4.5cm]{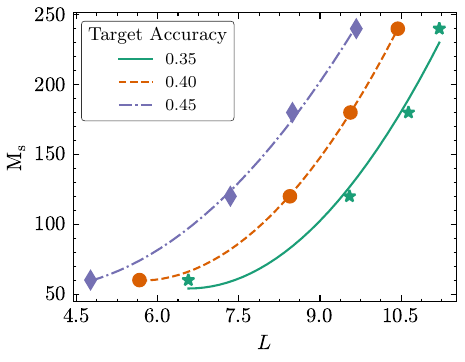}}
            \subfigure[]
       {\label{fig_adv-layer-Ns-linefit}\includegraphics[height=4.5cm]{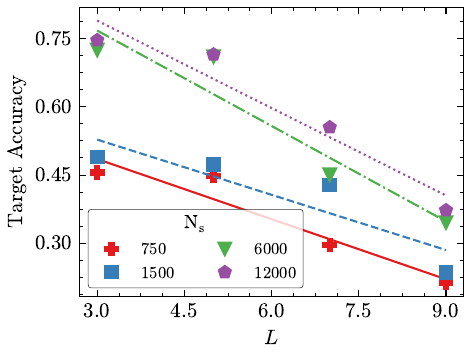}}
     \subfigure[]
       {\label{fig_adv-layer-Ns-quadprog}\includegraphics[height=4.5cm]{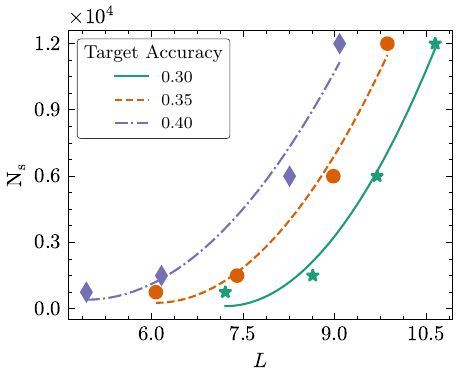}}
\end{center}
 \caption{Sample complexity of labeled samples ($\Ms$) and all samples ($\Ns$) with respect to the depth $\numL$ of adversarial domain adaptation networks. Left panels (a),(c): Variation of target accuracy with $\numL$. Right panels (b),(d): Variation of the number of samples ($\Ms, \Ns$) required for attaining a desired target accuracy level with $\numL$.}
 \label{fig_adv_results_L}
\end{figure}

\begin{figure}[t]
\begin{center}
     \subfigure[]
       {\label{fig_adv-dcm-Ms-linefit}\includegraphics[height=4.5cm]{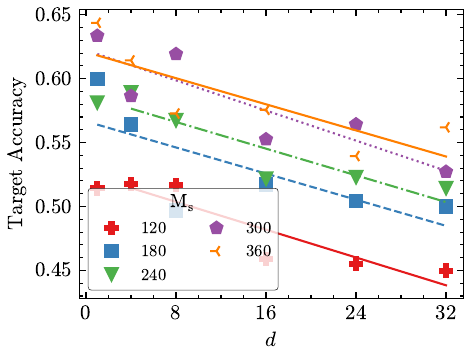}}
     \subfigure[]
       {\label{fig_adv-dcm-Ms-quadprog}\includegraphics[height=4.5cm]{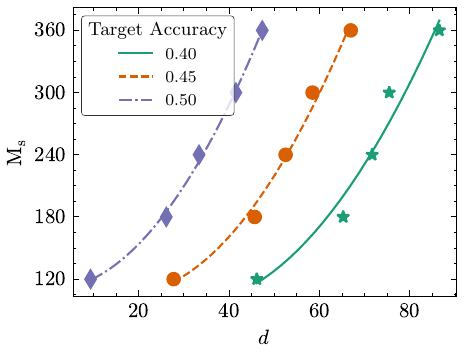}}
            \subfigure[]
       {\label{fig_adv-dcm-Ns-linefit}\includegraphics[height=4.5cm]{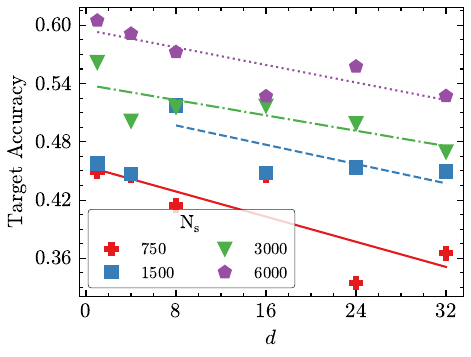}}
     \subfigure[]
       {\label{fig_adv-dcm-Ns-quadprog}\includegraphics[height=4.5cm]{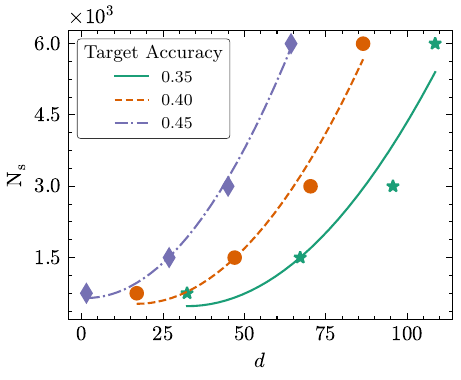}}
 \end{center}
 \caption{Sample complexity of labeled samples ($\Ms$) and all samples ($\Ns$) with respect to the width $\dcom$ of adversarial domain adaptation networks. Left panels (a),(c): Variation of target accuracy with $\dcom$. Right panels (b),(d): Variation of the number of samples ($\Ms, \Ns$) required for attaining a desired target accuracy level with $\dcom$.}
 \label{fig_adv_results_d}
\end{figure}

\begin{figure}[ht]
\begin{center}
     \subfigure[]
       {\label{fig_adv-optalpha_quadfit_Ms240}\includegraphics[height=4.5cm]{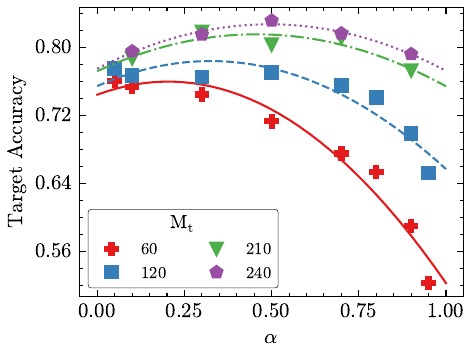}}
     \subfigure[]
       {\label{fig_adv-optalpha-asqrtx_fit_den_Mss.pdf}\includegraphics[height=4.5cm]{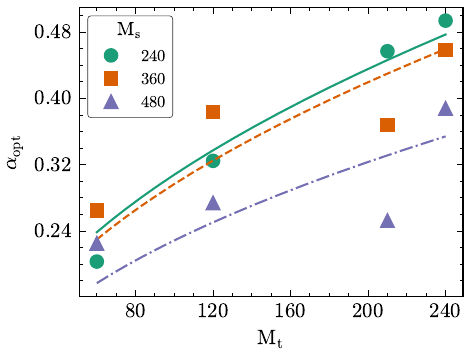}}
 \end{center}
 \caption{(a) Variation of target accuracy with target loss weight parameter $\alpha$ for adversarial domain adaptation networks (obtained at $\Ms = 240$). (b) Variation of optimal weight $\alpha_{opt}$ with number of labeled target samples $\Mt$.}
 \label{fig_adv_results_alpha}
\end{figure}

The feature extractor network contains only convolutional layers, while the label predictor and domain discriminator networks consist of fully connected layers in the implementation in \cite{fungtion_DANN_py3}. In order to adapt our experiments to this structure, when analyzing the sample complexity of labeled data ($\Ms$), we set the number of layers in the feature extractor and label predictor networks as equal, which is represented by the parameter $\numL$. Likewise, when studying the sample complexity of all data ($\Ns$), the number of layers in the feature extractor and domain discriminator networks are equated and denoted as  $\numL$. We use a similar strategy to adjust the network width, where we scale the number of convolutional channels and the fully connected layer width in the original paper  \cite{GaninUAGLLML16}  with the same factor $\dcom$. Hence, the number of convolutional channels is scaled proportionally to the width of the label predictor and the domain discriminator networks, respectively, when studying the sample complexities of $\Ms$ and $\Ns$. Batch normalization and ReLU layers are included after each convolutional or fully connected layer, following standard practice.

The sample complexities of the number of source samples with the network depth $\numL$ and  width $\dcom$ are presented, respectively in Figures \ref{fig_adv_results_L} and \ref{fig_adv_results_d}. Similarly to the experiments in Section \ref{ssec_exp_mmd_dan}, left panels (a) and (c) show the variation of the target accuracy with $\numL$ or $\dcom$ at different $\Ms$ and $\Ns$ values. The plots in the right panels (b) and (d) are then obtained by investigating the smallest $\Ms$ and $\Ns$ values ensuring a reference target accuracy level as $\numL$ or $\dcom$ increases. The results of these experiments align with the theoretical bounds in Theorem \ref{thm_main_result_dann}, confirming the quadratic growth in the sample complexities $\Ms, \Ns=O(\numL^2)$  and $\Ms, \Ns = O(\dcom^2)$ as the network depth $\numL$ and width $\dcom$ increase. 

We lastly study the choice of the parameter $\alpha$ weighting the target classification loss in the objective function for the adversarial setting. The results presented in Figure \ref{fig_adv_results_alpha} confirm the theoretical prediction that the  optimal value  of the weight parameter should scale at rate $\alpha_{opt} =O( \sqrt{\Mt})$ as the number of labeled samples varies. 

Overall, our experimental findings in Section \ref{ssec_exp_dan} are in line with the theoretical bounds presented in Theorems \ref{thm_main_result_da_mmd} and \ref{thm_main_result_dann}, supporting our sample complexity and optimal weight choice analyses for both MMD-based and adversarial domain adaptation networks.

\section{Conclusion}
\label{sec:conclusion}

We have presented a theoretical analysis of semi-supervised domain adaptation methods that jointly learn feature transformations that map the source and target domains to a shared space, along with a classifier defined in that space. We have first derived general performance bounds applicable to arbitrary function classes and domain discrepancy measures. We have then specialized these results under the assumption that the domain alignment is measured using the maximum mean discrepancy (MMD) metric. Our results show that the number of labeled source samples must scale logarithmically with the covering number of the combined hypothesis class comprising the feature transformation and the classifier, while the total sample sizes must scale logarithmically with the covering numbers of the feature transformation classes alone.

Building on these results, we have then extended our analysis to characterize the sample complexity of domain-adaptive neural networks. Our treatment relies on a detailed examination of the covering numbers of the corresponding function classes in deep architectures. We have focused on two types of neural networks, which perform domain alignment via MMD-based transformations or through adversarial objectives. In both cases, our analysis indicates that the sample complexities for both labeled and unlabeled data grow quadratically with the network depth and width. We have also shown that the scarcity of labeled target data can be effectively mitigated by scaling the weight of the target classification loss proportionally to the square root of the number of labeled target samples. 

To the best of our knowledge, our study provides the first comprehensive theoretical characterization of the sample complexity of domain-adaptive neural networks.

\section*{Acknowledgement}

The authors would like to thank \"Ozlem Akg\"ul, \"Omer Faruk Arslan, Atilla Can Aydemir, Firdevs Su Ayd{\i}n and Enes Ata \"Unsal for their help with the experiments in Section \ref{ssec_exp_mmd_dan}.

\appendix

\section{Proof of Lemma \ref{lem:weight_loss_gen}}
\label{pf_lem_weight_loss_gen}

\begin{proof}
We characterize the complexity of function spaces via covering numbers \cite{CuckerS02}. We first derive a bound for the deviation between the expected and empirical target losses. Let the open balls of radius $\frac{\epsilon}{8 \alpha \Lls}$ around the functions $ \{ \gt_k \}_{k=1}^{\Kt}$ be a cover for the function space $\Hs \circ \Ft$ with covering number 

\[
{\Kt} = \N( \Hs \circ \Ft, \frac{\epsilon}{8 \alpha \Lls}, \dt).
\]

Take any $\gt_k = \h_k \circ \ft_k$, for $k=1, \dots, \Kt$. The random variables $\loss( \gt_k (\xjt), \yjt ) $, $j= 1, \dots, \Mt$ are independent identically distributed, bounded as $ | \loss( \gt_k (\xjt), \yjt ) | \leq \bls$, and they have mean $\Lt (\ft_k, \h_k)$. From Hoeffding's inequality, we get that for each $k$, the deviation between the empirical loss and the expected loss is bounded as

\[
P\left( | \hLt( \ft_k, \h_k) - \Lt (\ft_k, \h_k) | \geq \frac{\epsilon}{4 \alpha} \right) \leq 2 e^{-\frac{\Mt \epsilon^2}{8 \alpha^2 \bls^2}}.
\]

Then, from union bound, with probability at least $1 - 2 \Kt e^{-\frac{\Mt \epsilon^2}{8 \alpha^2 \bls^2}}$, the inequality 

\[
| \hLt( \ft_k, \h_k) - \Lt (\ft_k, \h_k) | \leq \frac{\epsilon}{4 \alpha}
\]
holds for all $k= 1, \dots, \Kt$. Now for any $\gt = \h \circ \ft \in \Hs \circ \Ft$, there exists at least one $\gt_k $ such that 

\[
\dt(\gt, \gt_k) <  \frac{\epsilon}{8 \alpha \Lls}.
\]
This gives
\begin{equation*}
\begin{split}
& |  \Lt(  \ft, \h ) - \Lt(  \ft_k, \h_k )   | 
= \left | \int_{\Zt}  \left( \loss( \gt (\xt), \yt ) -   \loss( \gt_k (\xt), \yt )  \right) \, d \mut  \right | \\
& \leq  \int_{\Zt}   \left | \loss(  \gt  (\xt), \yt ) -   \loss(  \gt_k(\xt), \yt )  \right | \, d \mut  
 \leq  \int_{\Zt}  \Lls \|  \gt  (\xt) -  \gt_k(\xt)  \| \, d \mut  \\
&\leq  \Lls  \int_{\Zt} \dt(\gt, \gt_k)  \, d \mut  
 <  \frac{\epsilon}{8 \alpha }.
\end{split}
\end{equation*}

It is easy to show similarly that
\[
|  \hLt(  \ft, \h ) - \hLt(  \ft_k, \h_k ) | <  \frac{\epsilon}{8 \alpha }.
\]
Then with probability at least 

\[1 - 2 \N( \Hs \circ \Ft, \frac{\epsilon}{8 \alpha \Lls}, \dt) e^{-\frac{\Mt \epsilon^2}{8 \alpha^2 \bls^2}}
\]
for any $\gt = \h \circ \ft \in \Hs \circ \Ft$ we have
\begin{equation*}
\begin{split}
& |  \Lt(  \ft, \h ) - \hLt(  \ft, \h )   |  \\
&\leq   |  \Lt(  \ft, \h ) - \Lt(  \ft_k, \h_k ) | 
 + |   \Lt(  \ft_k, \h_k ) -    \hLt( \ft_k, \h_k) 
  + |   \hLt( \ft_k, \h_k) -  \hLt(  \ft, \h )   |  \\
& <   \frac{\epsilon}{8 \alpha } + \frac{\epsilon}{4 \alpha} + \frac{\epsilon}{8 \alpha }
=  \frac{\epsilon}{2 \alpha }.
\end{split}
\end{equation*}
Replacing $\alpha$ with $1-\alpha$ and applying the same steps for the function space $\Hs \circ \Fs$, we similarly obtain that with probability at least 

\[
1 - 2 \N( \Hs \circ \Fs, \frac{\epsilon}{8 (1-\alpha) \Lls}, \ds) e^{-\frac{\Ms \epsilon^2}{8 (1-\alpha)^2 \bls^2}}
\]
the difference between the expected and empirical source losses is bounded for any $\fs$ and $\h$ as
\begin{equation*}
\begin{split}
& |  \Ls(  \fs, \h ) - \hLs(  \fs, \h )   |  <  
  \frac{\epsilon}{2 (1-\alpha) }.
\end{split}
\end{equation*}
Combining these results, we get that with probability at least 
\begin{equation}
\begin{split}
&1 - 2 \N( \Hs \circ \Ft, \frac{\epsilon}{8 \alpha \Lls}, \dt) e^{-\frac{\Mt \epsilon^2}{8 \alpha^2 \bls^2}} 
-2 \N( \Hs \circ \Fs, \frac{\epsilon}{8 (1-\alpha) \Lls}, \ds) e^{-\frac{\Ms \epsilon^2}{8 (1-\alpha)^2 \bls^2}}
\end{split}
\end{equation}
the largest difference between the expected and empirical total weighted losses is bounded as
\begin{equation*}
\begin{split}
\sup_{\fs \in \Fs, \ft \in \Ft, \h \in \Hs}  & |  \Lw( \fs, \ft, \h ) - \hLw( \fs,  \ft, \h )   | \\
& \leq \alpha \sup  |  \Lt(  \ft, \h ) - \hLt(  \ft, \h )   | 
+ (1-\alpha) \sup   |  \Ls(  \fs, \h ) - \hLs(  \fs, \h )   | \\
& \leq \epsilon.
\end{split}
\end{equation*}
\end{proof}

\section{Proof of Lemma \ref{lem:bernstein_src_trg}}
\label{pf_lem_bernstein_src_trg}

\begin{proof}
Our proof is based on the following result by Yurinskii \cite{Yurinski76}.

\begin{theorem} \cite[Theorem 2.1]{Yurinski76}
\label{thm:yurinski}
Let $\zeta_1, \dots, \zeta_N \in \mathcal{B}$ be independent random vectors, where $\mathcal{B}$ is a Banach space. Assume for all $i=1, \dots, N$
\begin{equation}
\label{eq:cond_mom_yur}
E[ \| \zeta_i \|^k ] \leq \frac{k!}{2} \, b_i^2 \, C^{k-2}, \text{ for } k=2,3, \cdots.
\end{equation}
If $x>\beta_N / B_N$ where
\begin{equation}
\label{eq:cond_betan_Bn}
\beta_N \geq E[ \| \zeta_1 + \dots + \zeta_N \| ], 
\quad
B_N^2 = b_1^2 + \dots + b_N^2,
\end{equation}
then 
\begin{equation*}
P\left(   \| \zeta_1 + \dots + \zeta_N \|  \geq x B_N  \right) \leq
  \exp \left(  
  -\frac{1}{8}\left(x-\frac{\beta_N}{B_N}\right)^2 \frac{1}{1+\left(x-\frac{\beta_N}{B_N} \right)\frac{C}{2 B_N}}
   \right).
\end{equation*}
\end{theorem}

Based on Theorem \ref{thm:yurinski}, we first derive the stated result for the source domain, whose generalization to the target domain is straightforward. First notice that, due to the assumptions \eqref{eq:var_bnd_fs_ft}, \eqref{eq:mom_bnd_fs_ft}, the random vectors $ \fs(\xis)- E[\fs(\xs)]$ for $i=1, \dots, \Ns$ satisy the condition \eqref{eq:cond_mom_yur}, for the choices $b_i=\sigma_s$ and $C=\Cs$. 

Next, we derive a constant $\beta_{\Ns}$ for which the zero-mean random vectors $\zeta_i = \fs(\xis)- E[\fs(\xs)]$ for $i=1, \dots, \Ns$ satisfy the condition \eqref{eq:cond_betan_Bn} for $N=\Ns$. From \eqref{eq:var_bnd_fs_ft}, we have
\begin{equation*}
E[ \| \zeta_i  \|^2 ] \leq \vars.
\end{equation*}
We consider now
\begin{equation*}
\begin{split}
E\left[ \left  \|   \sum_{i=1}^{\Ns} \zeta_i   \right \|^2
\right]
&= E\left[ \left \langle  \sum_{i=1}^{\Ns} \zeta_i ,  
\sum_{j=1}^{\Ns} \zeta_j   \right \rangle
\right]
= \sum_{i=1}^{\Ns} \sum_{j=1}^{\Ns}  E[ \langle \zeta_i, \zeta_j  \rangle ]  \\
&= \sum_{i=1}^{\Ns} E[ \langle \zeta_i, \zeta_i  \rangle ] 
+   \sum_{i=1}^{\Ns} \sum_{j \neq i, \, j=1}^{\Ns} E[ \langle \zeta_i, \zeta_j  \rangle ] 
 \leq \vars \Ns
\end{split}
\end{equation*}
where the last inequality follows from $E[ \| \zeta_i  \|^2 ] \leq \vars$, and the fact that we have $E[ \langle \zeta_i, \zeta_j  \rangle ] =0$ for independent and zero-mean $\zeta_i$ and $\zeta_j$ for $i \neq j$.
From the nonnegativity of the variance, we have $(E[Y])^2 \leq E[Y^2]$ for any random variable $Y$. Taking 
\begin{equation*}
Y= \left  \|  \sum_{i=1}^{\Ns} \zeta_i   \right \|
\end{equation*}
then yields
\begin{equation*}
\begin{split}
E\left[ \left  \|   \sum_{i=1}^{\Ns} \zeta_i   \right \|  
\right]
\leq
\left( 
E\left[ \left  \|  \sum_{i=1}^{\Ns} \zeta_i   \right \|^2
\right]
\right)^{1/2}
\leq 
\sigma_s \sqrt{\Ns}.
\end{split}
\end{equation*}
Hence defining $\beta_{\Ns} = \sigma_s \sqrt{\Ns}$, we get 
 \begin{equation}
 \label{eq:exp_sum_zetai}
 E[  \| \zeta_1 + \dots + \zeta_{\Ns} \|] \leq \beta_{\Ns}.
 \end{equation}
From the choice $b_i=\sigma_s$, we have $B_{\Ns}=\sqrt{\Ns} \sigma_s = \beta_{\Ns}$. Now for given $\epsilon>0$, from the assumption $\Ns > \vars / \epsilon^2$, the following choice for $x$ 
\begin{equation*}
x=\frac{\sqrt{\Ns} \epsilon}{\sigma_s} > 1
\end{equation*}
satisfies the condition $x>\beta_{\Ns} / B_{\Ns}$ as $\beta_{\Ns} = B_{\Ns}$. Then from Theorem \ref{thm:yurinski}, we have
\begin{equation*}
\begin{split}
P\left(   \| \zeta_1 + \dots + \zeta_{\Ns} \|  \geq \Ns \epsilon \right) \leq
  \exp \left(  
  -\frac{1}{8}\left( \frac{\sqrt{\Ns} \epsilon}{\sigma_s} -1\right)^2 \frac{1}{1+\left( \frac{\sqrt{\Ns} \epsilon}{\sigma_s} -1 \right)\frac{\Cs}{2 \sqrt{\Ns} \sigma_s }}
   \right).
\end{split}
\end{equation*}
Replacing  $\zeta_i = \fs(\xis)- E[\fs(\xs)]$ gives the stated result
\begin{equation*}
\begin{split}
&P\left( \left \|  \frac{1}{\Ns}  \sum_{i=1}^{\Ns}  \fs(\xis)  -  E[\fs(\xs)] \right \|  \geq  \epsilon \right) \\
&\leq
  \exp \left(  
  -\frac{1}{8}\left( \frac{\sqrt{\Ns} \epsilon}{\sigma_s} -1\right)^2 \frac{1}{1+\left( \frac{\sqrt{\Ns} \epsilon}{\sigma_s} -1 \right)\frac{\Cs}{2 \sqrt{\Ns} \sigma_s }}
   \right).
\end{split}
\end{equation*}
Applying the same analysis for the target domain, it is easy to show similarly that the upper bound for the target domain in \eqref{eq:lem_trg_empmean} also holds. 
\end{proof}


\section{Proof of Lemma \ref{lem:unif_bnd_D_hD}}
\label{pf_lem_unif_bnd_D_hD}

\begin{proof}

We begin with bounding the deviation $| \D(\fs, \ft) - \hD (\fs, \ft)  | $ between the MMD and its empirical estimate for a fixed pair of transformations. Let $\fs$ and $\ft$ be a given, fixed pair of transformations. We have
\begin{equation}
\label{eq:d_min_hD_v1}
\begin{split}
& | \D(\fs,\ft) - \hD(\fs,\ft) | \\
&= \left |  \|  E[\fs(\xs)] - E[ \ft(\xt) ] \|  
- \big \|
\frac{1}{\Ns} \sum_{i=1}^{\Ns} \fs(\xis) -  \frac{1}{\Nt} \sum_{j=1}^{\Nt} \ft(\xjt) 
\big \|
\right | \\
& \leq 
\big \|    
\frac{1}{\Ns} \sum_{i=1}^{\Ns} \fs(\xis) - E[\fs(\xs)] 
 \big \|
 + 
 \big \|    
\frac{1}{\Nt} \sum_{j=1}^{\Nt} \ft(\xjt)  - E[ \ft(\xt) ]
 \big \|.
\end{split}
\end{equation}
Replacing $\epsilon$ by $\epsilon/4$ in Lemma \ref{lem:bernstein_src_trg}, we observe that with probability at least
\begin{equation*}
1 - \exp(-\as(\Ns, \epsilon)) - \exp(-\at(\Nt, \epsilon))
\end{equation*}
we have
\begin{equation*}
\begin{split}
\big \|    
\frac{1}{\Ns} \sum_{i=1}^{\Ns} \fs(\xis) - E[\fs(\xs)] 
 \big \| \leq \frac{\epsilon}{4}, 
 \quad
  \big \|    
\frac{1}{\Nt} \sum_{j=1}^{\Nt} \ft(\xjt)  - E[ \ft(\xt) ]
 \big \|  \leq \frac{\epsilon}{4}
\end{split}
\end{equation*}
which yields from \eqref{eq:d_min_hD_v1}
\begin{equation*}
| \D(\fs,\ft) - \hD(\fs,\ft) |  \leq \frac{\epsilon}{2}.
\end{equation*}

In order to extend the above bound to the whole space of transformations, we consider covers of the function classes $\Fs$ and $\Ft$, consisting of open balls of radius $\epsilon/8$ respectively  around the functions $\{ \fs_k \}_{k=1}^{\Ks}$ and $\{ \ft_l \}_{l=1}^{\Kt}$, where $\Ks$ and $\Kt$ are the covering numbers
\begin{equation*}
\Ks = \N(\Fs, \frac{\epsilon}{8}, \dXs), 
\quad
\Kt = \N(\Ft, \frac{\epsilon}{8}, \dXt).
\end{equation*}
From the union bound, it follows that with probability at least 
\begin{equation*}
1 - \Ks \exp(-\as(\Ns, \epsilon)) - \Kt \exp(-\at(\Nt, \epsilon))
\end{equation*}
for all $k=1, \dots, \Ks$ and $l=1, \dots, \Kt$,
\begin{equation}
\label{eq:bnd_D_hD_fsk_ftl}
\begin{split}
| \D(\fs_k,\ft_l) - \hD(\fs_k,\ft_l) |  \leq \frac{\epsilon}{2}.
\end{split}
\end{equation}

Now, let us consider an arbitrary pair of transformations $\fs \in \Fs$ and $\ft \in \Ft$. As the balls around $\{ \fs_k \}_{k=1}^{\Ks}$ and $\{ \ft_l \}_{l=1}^{\Kt}$ form $\epsilon/8$-covers of the function classes, there exists a source transformation $\fs_k$ and a target transformation $\ft_l$ such that
\begin{equation*}
\dXs (\fs, \fs_k) < \frac{ \epsilon}{8},
\quad
\dXt (\ft, \ft_l) < \frac{ \epsilon}{8}.
\end{equation*}
We can then bound the difference between the MMD and its sample mean for $\fs$ and $\ft$ as follows.
\begin{equation}
\label{eq:form_D_Dhat_fs_ft}
\begin{split}
| \D(\fs, \ft) - \hD(\fs, \ft) |
&\leq  | \D(\fs, \ft) - \D(\fs_k, \ft_l) |
 + | \D(\fs_k, \ft_l) - \hD(\fs_k, \ft_l) | \\
 &+  | \hD(\fs_k, \ft_l) - \hD(\fs, \ft) | 
\end{split}
\end{equation}
Next, we bound each one of the terms on the right hand side of the above inequality. The first term can be upper bounded as
\begin{equation}
\label{eq:bnd_Dfsft_Dfskftl}
\begin{split}
| \D(\fs, \ft) - \D(\fs_k, \ft_l) | 
&=
\left | 
\|  
E[\fs(\xs)] - E[\ft(\xt)]
 \|
 - 
 \|  
E[\fs_k(\xs)] - E[\ft_l(\xt)]
 \|
\right | \\
&\leq
\|
E[\fs(\xs)] - E[\fs_k(\xs)] 
\|
+
\|
E[\ft(\xt)] - E[\ft_l(\xt)] 
\| \\
&=
\|
E[\fs(\xs)  - \fs_k(\xs)] 
\|
+
\|
E[\ft(\xt)  - \ft_l(\xt)] 
\| \\
& \leq
E[ \| \fs(\xs)  - \fs_k(\xs)  \| ] 
+ E[ \| \ft(\xt)  - \ft_l(\xt) \| ] 
\end{split}
\end{equation}
where the last inequality follows from Jensen's inequality, observing the fact that a norm over a Hilbert space is a convex function. From the definition of the metrics $\dXs$ and $\dXt$, we have 
\begin{equation*}
\begin{split}
\| \fs(\xs)  - \fs_k(\xs)  \|  \leq \dXs(\fs, \fs_k) \\
\| \ft(\xt)  - \ft_l(\xt)  \|  \leq \dXt(\ft, \ft_l) \
\end{split}
\end{equation*}
for all $\xs \in \Xs$ and $\xt \in \Xt$. Using this in \eqref{eq:bnd_Dfsft_Dfskftl}, we get
\begin{equation*}
\begin{split}
| \D(\fs, \ft) - \D(\fs_k, \ft_l) | 
\leq 
\dXs(\fs, \fs_k) +  \dXt(\ft, \ft_l)
< 
\frac{ \epsilon}{8} + \frac{ \epsilon}{8} = \frac{ \epsilon}{4}.
\end{split}
\end{equation*}
With a similar analysis by replacing the expectations with the sample means, it is easy to show that the third term in the inequality \eqref{eq:form_D_Dhat_fs_ft} can also be upper bounded as
\begin{equation*}
| \hD(\fs_k, \ft_l) - \hD(\fs, \ft) | < \frac{ \epsilon}{4}.
\end{equation*}
Now, remembering also the probabilistic upper bound \eqref{eq:bnd_D_hD_fsk_ftl} that holds for the second term in \eqref{eq:form_D_Dhat_fs_ft} for all $k$ and $l$, we get that with probability at least 
\begin{equation*}
1 - \Ks \exp(-\as(\Ns, \epsilon)) - \Kt \exp(-\at(\Nt, \epsilon))
\end{equation*}
we have for all $\fs \in \Fs$ and $\ft \in \Ft$,
\begin{equation*}
\begin{split}
| \D(\fs, \ft) - \hD(\fs, \ft) | < \frac{ \epsilon}{4} + \frac{ \epsilon}{2} + \frac{ \epsilon}{4}
= \epsilon.
\end{split}
\end{equation*}
Hence, we get the stated result
\begin{equation*}
\begin{split}
& P \left(
\sup_{\fs \in \Fs, \ft \in \Ft}  | \D(\fs, \ft) - \hD (\fs, \ft)  | < \epsilon
\right) \\
&\geq 1 - \Ks \exp(-\as(\Ns, \epsilon)) - \Kt \exp(-\at(\Nt, \epsilon))  \\
&= 1 - \N(\Fs, \frac{\epsilon}{8}, \dXs) \exp(-\as(\Ns, \epsilon)) 
- \N(\Ft, \frac{\epsilon}{8}, \dXt) \exp(-\at(\Nt, \epsilon)).
\end{split}
\end{equation*}
\end{proof}


\section{Proof of Lemma \ref{lem_fs_ft_measble}}
\label{pf_lem_fs_ft_measble}

\begin{proof}
We prove the statements only for the source domain, as the proofs for the target domain are the same. Let $\hsl (\xs) \in \Rdl$ denote the feature in layer $\lay$ for the source input $\xs\in \R^\dz$, where we regard $\hsl(\cdot): \R^\dz \rightarrow \Rdl$ as a function. In the relation 
\[
\hsl (\xs) =\actl( \Wsl \hslm (\xs)+ \bsl)
\]
the expression $ \Wsl \hslm (\xs)+ \bsl$ is a continuous mapping of $\hslm (\xs)$, and the function $\actl $ is continuous. Hence, based on a simple induction argument it follows that $\hsl (\cdot): \Xs= \R^\dz \rightarrow \Rdl$ is a continuous, thus measurable function (a Borel map).

We next show that the mappings $\fsl: \Xs \rightarrow \Xl$ are measurable.   Let $\B(\cdot)$ denote the Borel $\sigma$-algebra of a metric space. We recall from \eqref{eq_defn_fsl_ftl} that $\fsl(\xs)=\phil(\hsl(\xs)) \in \Xl $. Consider a sequence $\{\hsl_n\} \subset \Rdl$ with $\lim_{n \rightarrow \infty} \hsl_n = \hsl_*$ for some $\hsl_* \in \Rdl$. As the kernel $\krl(\cdot , \cdot ) $ is assumed to be a continuous function, we have
\begin{equation*}
\begin{split}
\lim_{n\rightarrow \infty} \| \phil(\hsl_n) - \phil(\hsl_*) \|^2_{\Xl} 
=\lim_{n\rightarrow \infty} 
\left(
 \krl(\hsl_n, \hsl_n) -2 \krl(\hsl_n, \hsl_* )  +\krl(\hsl_*, \hsl_*) \right)
 =0
\end{split}
\end{equation*}
where $\| \cdot \|_{\Xl}$ denotes the norm in the RKHS $\Xl$. It thus follows that
\[ \lim_{n \rightarrow \infty} \phil(\hsl_n) = \phil ( \hsl_*)\] 
and hence $\phil: \Rdl \rightarrow \Xl$ is a continuous function. $\phil $ is thus measurable with respect to the Borel $\sigma$-algebra $ \B(\Xl)$   of the RKHS $\Xl$. Since $\hsl (\cdot): \Xs \rightarrow \Rdl$  is a measurable mapping as well, we conclude that the mapping $\fsl=\phil(\hsl(\cdot)) : \Xs \rightarrow \Xl$ is measurable with respect to $\B(\Xl)$, for $\lay=1, \dots, \numL-1$.

We next show that the mappings $\fs \in \Fs$ are measurable. Since the kernel $\krl(\cdot , \cdot ) $ is  assumed to be continuous, the RKHS $\Xl$ is separable for all $\lay$ \cite{SubediC19}. The separability of the RKHSs ensures that
\[
\B(\X) = \bigotimes_{l=1}^{\numL-1} \B(\Xl)
\]
where the right hand side denotes the $\sigma$-algebra generated by all finite products of Borel sets in $\B(\Xl)$'s \cite{Bogachev07}. Hence, denoting the set product of some collection of Borel sets $B^1 \in \B(\X^1), \ \cdots, B^{\numL-1} \in \B(\X^{\numL-1})$ as 
\[
B^1 \times B^2 \times \cdots \times B^{\numL-1}
 = \{ (f^1, f^2, \dots \, , f^{\numL-1}) : f^\lay \in B^\lay, \, \lay=1, \dots, \numL-1 \},
\]
the $\sigma$-algebra generated by
\[
B=\{ B^1 \times \cdots \times B^{\numL-1} : B^1 \in \B(\X^1), \cdots, B^{\numL-1} \in \B(\X^{\numL-1}) \}
\]
is equal to the Borel $\sigma$-algebra $\B(\X)$. Then, in order to show that $\fs: \Xs \rightarrow \X$ is measurable, it is sufficient to show that the inverse image $(\fs)^{-1}(B)$ of the set $B$ is contained in $\B(\Xs)$. For any element $ B^1 \times \cdots \times B^{\numL-1}$ in $B$, we have
\begin{equation*}
\begin{split}
(\fs)^{-1} (B^1 \times \cdots \times B^{\numL-1}) 
&= \{ \xs \in \Xs : \fs(\xs) \in B^1 \times \cdots \times B^{\numL-1}\} \\
&=\{ \xs \in \Xs : \fsone(\xs)  \in B^1, \ \cdots \ , \fsLm(\xs) \in  B^{\numL-1}\} \\
&= \bigcap_{\lay=1}^{\numL-1} (\fsl)^{-1} (B^\lay).
\end{split}
\end{equation*}
Since each $\fsl$ is measurable, $(\fsl)^{-1} (B^\lay) \in \B(\Xs)$. Hence, $(\fs)^{-1}(B^1 \times \cdots \times B^{\numL-1}) \in \B(\Xs)$ and we conclude that $\fs: \Xs \rightarrow \X$ is a measurable mapping. 

In order to prove the second part of the lemma, let us fix $\boldsymbol{\xi} \in \Rdl$, and for fixed $\boldsymbol{\xi}$ consider the function $\fsl(\cdot)(\boldsymbol{\xi}): \Xs = \R^\dz \rightarrow \R$ given by
\[
\fsl(\cdot)(\boldsymbol{\xi})=\krl(\hsl(\cdot),\boldsymbol{\xi}).
\]
From the continuity of the kernel $\krl$ and the measurability of the function $\hsl(\cdot)$, it is easy to conclude that the function $\fsl(\cdot)(\boldsymbol{\xi})$ is measurable for any fixed $\boldsymbol{\xi}$. Hence, based on the Borel probability measure $\mus$ in the source domain, the expectation $E_{\xs}[\fsl(\xs)(\boldsymbol{\xi})]$ for fixed $\boldsymbol{\xi}$ is well defined, as well as the function $E_{\xs}[\fsl(\xs)]: \Rdl \rightarrow \R$ given by
\[
E_{\xs}[\fsl(\xs)](\boldsymbol{\xi}) \triangleq E_{\xs}[\fsl(\xs)(\boldsymbol{\xi})].
\]

Next, we would like to show that $E_{\xs}[\fsl(\xs)] \in \Xl$. Consider the linear functional $T_{\mus}: \Xl \rightarrow \R$ on the RKHS $\Xl$ defined by
\[
T_{\mus}(\psi) \triangleq E_{\xs}[ \psi(\hsl) ]  
\]
for $\psi \in \Xl$. Following the steps as in the proof of \cite[Lemma 3]{GrettonBRSS12}, the linear functional $T_{\mus}$ is observed to be bounded since 
\begin{equation*}
\begin{split}
| T_{\mus}(\psi) | &= \left| E_{\xs}[ \psi(\hsl) ] \right |
\leq  E_{\xs} \left[ | \psi(\hsl) | \right]  
=  E_{\xs} \left[ \left|  \langle \krl( \hsl, \cdot) , \psi(\cdot)    \rangle_{\Xl} \right| \right]  \\
& \leq E_{\xs}  \left[    \|   \krl( \hsl, \cdot)  \|_{\Xl}  \|  \psi \|_{\Xl}   \right]
= E_{\xs} \left[ \sqrt{\krl( \hsl, \hsl)}   \right]   \|  \psi \|_{\Xl}.
\end{split}
\end{equation*}
Hence, by the Riesz Representation Theorem \cite[Theorem 12.5]{BachmanN66},\cite[Lemma 3]{GrettonBRSS12}, there exists an element $\psi^{sl} \in \Xl$ in the RKHS $\Xl$ (called the mean embedding), such that 
\[
T_{\mus}(\psi) = \langle  \psi, \psi^{sl} \rangle_{\Xl}
\]
for all $\psi \in \Xl$. In particular, setting $\psi = \phil(\boldsymbol{\xi})$ for an arbitrary $\boldsymbol{\xi} \in \Rdl$, we have
\begin{equation}
\label{eq_Tmus_phil1}
T_{\mus}(\phil(\boldsymbol{\xi})) = \langle  \phil(\boldsymbol{\xi}), \psi^{sl} \rangle_{\Xl}
=\psi^{sl} (\boldsymbol{\xi}). 
\end{equation}
But it also holds that
\begin{equation}
\label{eq_Tmus_phil2}
\begin{split}
T_{\mus}(\phil(\boldsymbol{\xi}))  &=  E_{\xs}[ \phil(\boldsymbol{\xi})(\hsl) ] = E_{\xs}[ \krl(\boldsymbol{\xi}, \hsl) ]
= E_{\xs}[ \krl(\hsl, \boldsymbol{\xi}) ] \\
&= E_{\xs}[ \phil(\hsl) (\boldsymbol{\xi}) ]
=E_{\xs}[ \fsl(\xs) (\boldsymbol{\xi}) ] = E_{\xs}[ \fsl(\xs)  ] (\boldsymbol{\xi}).
\end{split}
\end{equation}
From the equality of the expressions in \eqref{eq_Tmus_phil1} and \eqref{eq_Tmus_phil2}, we observe that
\[
 E_{\xs}[ \fsl(\xs)  ] = \psi^{sl} \in \Xl.
\] 
It then simply follows from the construction of $\X$ that 
\[
 E_{\xs}[\fs(\xs)] \triangleq ( E_{\xs}[\fsone (\xs)], \dots \, , E_{\xs}[\fsLm(\xs)]    )  \\ 
\]
is in the Hilbert space $\X$.

\end{proof}


\section{Derivation of Lipschitz constants for common nonlinear activation functions}
\label{sec_app_lip_nonlinact}

Here we derive Lipschitz constants for some widely used nonlinear activation functions. Let $\act: \Rdl \rightarrow \Rdl$ represent an activation function in layer $\lay$ giving the output $\zeta= \act (\boldsymbol{\xi})$ for the input $\boldsymbol{\xi} \in \Rdl$.   

\subsection{ReLU activation}
We begin with the rectified linear unit (ReLU) function $\act_{R}: \Rdl \rightarrow \Rdl$ given by
\begin{equation}
\begin{split}
\boldsymbol{\zeta}(k) = \max \{ 0, \boldsymbol{\xi}(k) \}
\end{split}
\end{equation}
where $\boldsymbol{\zeta}= \act_R (\boldsymbol{\xi})$, and the notation $(\cdot)(k)$ denotes the $k$-th entry of a vector. For two vectors $\boldsymbol{\xi}_1, \boldsymbol{\xi}_2 \in \Rdl$, we have
\begin{equation}
\begin{split}
\|  \act_R(\boldsymbol{\xi}_1) - \act_R(\boldsymbol{\xi}_2)  \|^2 
&=\sum_{k=1}^{\dl} ( \max \{ 0, \boldsymbol{\xi}_1(k) \} -    \max \{ 0, \boldsymbol{\xi}_2(k) \} )^2 \\
&\leq \sum_{k=1}^{\dl} ( \boldsymbol{\xi}_1(k) - \boldsymbol{\xi}_2(k))^2 
= \| \boldsymbol{\xi}_1 - \boldsymbol{\xi}_2 \|^2  \\
\end{split}
\end{equation}
where $\max\{\cdot, \cdot\}$ denotes the maximum of two scalar values. We thus get
\[
\|  \act_R(\boldsymbol{\xi}_1) - \act_R(\boldsymbol{\xi}_2)  \| \leq \| \boldsymbol{\xi}_1 - \boldsymbol{\xi}_2 \|
\]
which gives the Lipschitz constant of the ReLU function as $\Lact_R =1$.

\subsection{Softplus activation}
Next, we consider the softplus function $\act_{SP}: \Rdl \rightarrow \Rdl$ given by
\begin{equation}
\begin{split}
\boldsymbol{\zeta}(k) = \log \left(1+e^{\boldsymbol{\xi}(k)}\right)
\end{split}
\end{equation}
where $\boldsymbol{\zeta}= \act_{SP} (\boldsymbol{\xi})$. The derivative of the components of the softplus function can be upper bounded as
\begin{equation}
\begin{split}
\left | \frac{d}{dt} \log(1+e^t)  \right |= \left | \frac{e^t}{1+e^t} \right | 
<1
\end{split}
\end{equation}
for all $t \in \R$. Then for  $\boldsymbol{\zeta}_1= \act_{SP} (\boldsymbol{\xi}_1)$ and $\boldsymbol{\zeta}_2= \act_{SP} (\boldsymbol{\xi}_2)$ with $\boldsymbol{\xi}_1, \boldsymbol{\xi}_2 \in \Rdl$, from the mean value theorem we get
\begin{equation}
\begin{split}
| \boldsymbol{\zeta}_1(k) -  \boldsymbol{\zeta}_2(k) | \leq | \boldsymbol{\xi}_1(k) - \boldsymbol{\xi}_2(k) |
\end{split}
\end{equation}
which implies 
\begin{equation}
\begin{split}
\| \act_{SP}(\boldsymbol{\xi}_1) -  \act_{SP}(\boldsymbol{\xi}_2) \|  \leq \|  \boldsymbol{\xi}_1 - \boldsymbol{\xi}_2 \|.
\end{split}
\end{equation}
Hence, we obtain the Lipschitz constant of the softplus function as $\Lact_{SP} =1$.

\subsection{Softmax activation}

Lastly, we consider the softmax function  $\act_{SM}: \Rdl \rightarrow \Rdl$ given by 
\[
\act_{SM}(\boldsymbol{\xi}) = [\act_{SM}^1(\boldsymbol{\xi}) \ \act_{SM}^2(\boldsymbol{\xi}) \ \cdots  \ \act_{SM}^{\dl}(\boldsymbol{\xi}) ]^T
\]
where $\boldsymbol{\xi} \in \Rdl$ and each $k$-th component $\act_{SM}^k(\boldsymbol{\xi}): \Rdl \rightarrow \R$ of the softmax activation is defined as
\begin{equation}
\begin{split}
\act_{SM}^k(\boldsymbol{\xi}) = \frac{e^{\boldsymbol{\xi}(k)}}{ \sum_{n=1}^{\dl} e^{\boldsymbol{\xi}(n)}}.
\end{split}
\end{equation}
Since the functions $\act_{SM}^k(\boldsymbol{\xi})$ are differentiable for all $k$, for any two $\boldsymbol{\xi}_1, \boldsymbol{\xi}_2 \in \Rdl$, it follows from the multivariable mean value theorem that there exists some $\boldsymbol{\xi} \in \Rdl$ lying in the line segment between $\boldsymbol{\xi}_1$ and $\boldsymbol{\xi}_2$ such that
\[
\act_{SM}^k(\boldsymbol{\xi}_1) - \act_{SM}^k(\boldsymbol{\xi}_2) = (\nabla \act_{SM}^k(\boldsymbol{\xi}))^T (\boldsymbol{\xi}_1 - \boldsymbol{\xi}_2)
\]
where $\nabla \act_{SM}^k (\boldsymbol{\xi}) \in \Rdl$ denotes the gradient of $\act_{SM}^k$ at $\boldsymbol{\xi}$. The following inequality is then obtained
\begin{equation}
\label{eq_bnd_grad_sm}
| \act_{SM}^k(\boldsymbol{\xi}_1) - \act_{SM}^k(\boldsymbol{\xi}_2) | \leq 
\sup_{\boldsymbol{\xi} \in \Rdl} \| \nabla \act_{SM}^k (\boldsymbol{\xi})   \|  \, \| \boldsymbol{\xi}_1 - \boldsymbol{\xi}_2 \|.
\end{equation}
In the sequel, in order to find a Lipschitz constant for the softmax function, we derive a bound on the norm $\| \nabla \act_{SM}^k (\boldsymbol{\xi})   \| $ of its gradient. 

For the case $k\neq n$, the derivative of $\act_{SM}^k (\boldsymbol{\xi})  $ with respect to the $n$-th entry $\boldsymbol{\xi} (n)$ of $\boldsymbol{\xi} \in \Rdl$ is obtained as
\begin{equation*}
\begin{split}
\frac{\partial \act_{SM}^k (\boldsymbol{\xi})  }{\partial \boldsymbol{\xi}(n)} 
=  \frac{\partial }{\partial \boldsymbol{\xi}(n)} 
\left( \frac{e^{\boldsymbol{\xi}(k)}}{ \sum_{r=1}^{\dl} e^{\boldsymbol{\xi}(r)}} \right)
= -  \frac{e^{\boldsymbol{\xi}(k)} e^{\boldsymbol{\xi}(n)} }{\left( \sum_{r=1}^{\dl} e^{\boldsymbol{\xi}(r)} \right)^2}.
\end{split}
\end{equation*}
Since all $e^{\boldsymbol{\xi}(1)}, \dots , e^{\boldsymbol{\xi}(\dl)}$ are positive, it is easy to show that $(e^{\boldsymbol{\xi}(1)} + \dots , +e^{\boldsymbol{\xi}(\dl)})^2 \geq 4 e^{\boldsymbol{\xi}(k)} e^{\boldsymbol{\xi}(n)}$. Using this in the above expression, we get the bound
\begin{equation}
\label{eq_lip_sm_kneqn}
\begin{split}
\left | \frac{\partial \act_{SM}^k (\boldsymbol{\xi})  }{\partial \boldsymbol{\xi}(n)} \right |
\leq \frac{1}{4}.
\end{split}
\end{equation}

Next, for the case $k= n$, we have
\begin{equation*}
\begin{split}
\frac{\partial \act_{SM}^k (\boldsymbol{\xi})  }{\partial \boldsymbol{\xi}(k)} 
=  \frac{\partial }{\partial \boldsymbol{\xi}(k)} 
\left( \frac{e^{\boldsymbol{\xi}(k)}}{ \sum_{r=1}^{\dl} e^{\boldsymbol{\xi}(r)}} \right)
= \left( \frac{e^{\boldsymbol{\xi}(k)}}{ \sum_{r=1}^{\dl} e^{\boldsymbol{\xi}(r)}} \right) \left( 1- \frac{e^{\boldsymbol{\xi}(k)}}{ \sum_{r=1}^{\dl} e^{\boldsymbol{\xi}(r)}} \right).
\end{split}
\end{equation*}
Letting $\alpha =  e^{\boldsymbol{\xi}(k)} / \sum_{r=1}^{\dl} e^{\boldsymbol{\xi}(r)}  $ in the above expression and observing that the maximum value of the function $\alpha (1-\alpha)$ in the interval $\alpha \in [0,1]$ is $1/4$, we get
\begin{equation}
\label{eq_lip_sm_keqn}
\begin{split}
\left | \frac{\partial \act_{SM}^k (\boldsymbol{\xi})  }{\partial \boldsymbol{\xi}(k)} \right |
\leq \frac{1}{4}.
\end{split}
\end{equation}
Combining the results \eqref{eq_lip_sm_kneqn} and \eqref{eq_lip_sm_keqn}, the gradient of $ \act_{SM}^k (\boldsymbol{\xi})   $  can be bounded as
\begin{equation*}
\begin{split}
\| \nabla \act_{SM}^k (\boldsymbol{\xi})   \|   \leq \frac{\sqrt{\dl}}{4}
\end{split}
\end{equation*}
for any $\boldsymbol{\xi} \in \Rdl$. Using this in \eqref{eq_bnd_grad_sm} gives
\[
| \act_{SM}^k(\boldsymbol{\xi}_1) - \act_{SM}^k(\boldsymbol{\xi}_2) |  \leq \frac{\sqrt{\dl}}{4}  \| \boldsymbol{\xi}_1 - \boldsymbol{\xi}_2 \|
\]
for any $\boldsymbol{\xi}_1, \boldsymbol{\xi}_2 \in \Rdl$, which implies
\begin{equation*}
\begin{split}
\|  \act_{SM}(\boldsymbol{\xi}_1) - \act_{SM}(\boldsymbol{\xi}_2)  \| \leq \frac{\dl}{4} \| \boldsymbol{\xi}_1 - \boldsymbol{\xi}_2 \|.
\end{split}
\end{equation*}
Defining
\begin{equation*}
\begin{split}
\dlmax=\max_{\lay=1, \dots, \numL} \dl
\end{split}
\end{equation*}
we thus get the Lipschitz constant of the softmax function as $\Lact_{SM} =\dlmax/4$.


\section{Proof of Lemma \ref{lem_Fs_Ft_HFs_HFt_comp}}
\label{pf_lem_Fs_Ft_HFs_HFt_comp}

\begin{proof}
We prove the statements only for $\Fs$ and $\Gs$ as the proofs for the target domain are similar. We first show that $\Fs$ is compact with respect to the metric $\dXs $. Let
\begin{equation*}
\begin{split}
\Phis&=\{ \Thetas=(\boldsymbol{\Theta}^{s1}, \dots, \boldsymbol{\Theta}^{s \numL}):  \ |\Thetasl_{ij}|  \leq \Bnet, \forall i, j, \lay \} \\
\end{split}
\end{equation*}
denote the parameter space over which the source network parameters are defined. Regarding $\Phis $ as the Cartesian product of the corresponding matrix spaces at layers $\lay=1, \dots, \numL$, it follows from the bound $ |\Thetasl_{ij}|  \leq \Bnet$ on the network parameters that the finite dimensional set $\Phis $ is closed and bounded, hence compact.
 
We next define a mapping $\mapFs: \Phis \rightarrow \Fs $ such that
\begin{equation}
\label{eq_defn_mapFs}
\begin{split}
\mapFs(\Thetas) = \fs_{\Thetas}=\big( \fsone_{\Thetas} , \, \dots \, , \fsLm_{\Thetas}    \big) 
\end{split}
\end{equation}
where the notation $ \fs_{\Thetas}(\xs)$ stands for the function $ \fs(\xs)$ defined in \eqref{eq_Fs_Ft_defn_dl} by explicitly referring to its dependence on the network parameters $\Thetas $. In the following, we show that the mapping $\mapFs $ is continuous. Let us consider a sequence $\{ \Thetas_n \}\subset \Phis $ converging to an element $ \Thetas_* \in \Phis$.  Since the relation \eqref{eq_defn_hsl_htl} between the features of adjacent layers is given by a linear mapping followed by a continuous activation function $\actl$, the mapping $\hsl_{\Thetas} (\xs) $ is a continuous function of $\boldsymbol{\Theta}^s$, i.e. 
\begin{equation}
\label{eq_hsl_cont}
\lim_{n \rightarrow \infty} \hsl_{\Thetas_n} (\xs) =\hsl_{\Thetas_*} (\xs).
\end{equation}
In  fact, due to the assumptions on the boundedness \eqref{eq_bnd_xs_xt} of the source samples, the boundedness \eqref{eq_bnd_Thetaij} of the network parameters, and the Lipschitz continuity \eqref{eq_Lipcont_act} of the activation functions $\actl$, it is easy to show that the convergence in \eqref{eq_hsl_cont} is uniform on $\Xs$. Hence, for any given $\epsilon >0$, one can find some $n_0$ such that for $n \geq n_0$, we have
\[
\| \hsl_{\Thetas_n} (\xs) -  \hsl_{\Thetas_*} (\xs)  \| < \epsilon
\]
for all $\xs \in \Xs$, for $\lay=1, \dots, \numL-1$. Then we have
%
%
\begin{equation*}
\begin{split}
& \|  \fsl_{\Thetas_n} (\xs)  - \fsl_{\Thetas_*} (\xs) \|_{\Xl}^2
=  \|  \phil(\hsl_{\Thetas_n} (\xs))   -   \phil(\hsl_{\Thetas_*} (\xs))   \|_{\Xl}^2 \\
&=   \krl \big(\hsl_{\Thetas_n} (\xs), \hsl_{\Thetas_n} (\xs) \big)
- 2 \krl \big(\hsl_{\Thetas_n} (\xs), \hsl_{\Thetas_*} (\xs) \big)
+ \krl \big(\hsl_{\Thetas_*} (\xs), \hsl_{\Thetas_*} (\xs) \big)  \\
& \leq 2 \Lk \| \hsl_{\Thetas_n} (\xs) -  \hsl_{\Thetas_*} (\xs) \|  
< 2 \Lk \epsilon
\end{split}
\end{equation*}
for all $\xs \in \Xs$ due to the Lipschitz continuity of the kernels $\krl$. This gives
%
%
\begin{equation*}
\begin{split}
 \|  \fs_{\Thetas_n} (\xs)  - \fs_{\Thetas_*} (\xs) \|_{\X}^2
= \sum_{\lay=1}^{\numL-1} \|  \fsl_{\Thetas_n} (\xs)  - \fsl_{\Thetas_*} (\xs) \|_{\Xl}^2 
< 2 (\numL-1) \Lk  \epsilon.
\end{split}
\end{equation*}
We have thus obtained
\begin{equation*}
\begin{split}
 \|  \fs_{\Thetas_n} (\xs)  - \fs_{\Thetas_*} (\xs) \|_{\X}
< \sqrt{2 (\numL-1) \Lk } \sqrt{ \epsilon}
\end{split}
\end{equation*}
for all $n \geq n_0$ and for all $\xs \in \Xs$, which shows that $\fs_{\Thetas_n} (\xs)$ converges to $\fs_{\Thetas_*} (\xs)$ uniformly on $\Xs$. Then we have
\begin{equation*}
\begin{split}
\lim_{n \rightarrow \infty}  \dXs(\fs_{\Thetas_n} , \fs_{\Thetas_*}   ) 
&= \lim_{n \rightarrow \infty} \sup_{\xs \in \Xs} 
 \|  \fs_{\Thetas_n} (\xs)  - \fs_{\Thetas_*} (\xs) \|_{\X} \\
 &= \sup_{\xs \in \Xs} \lim_{n \rightarrow \infty}  \|  \fs_{\Thetas_n} (\xs)  - \fs_{\Thetas_*} (\xs) \|_{\X}
=0
\end{split}
\end{equation*}
where the second equality follows from the uniform convergence of $\fs_{\Thetas_n} (\xs)$. We have thus shown that the mapping $\mapFs: \Phis \rightarrow \Fs$ defined in \eqref{eq_defn_mapFs} is continuous. Since the set $\Phis$ is compact, we conclude that the function space $\Fs $ is a compact metric space.

Next, in order to show the compactness of $\Gs$, we proceed in a similar fashion. Let us define a mapping $\mapGs: \Phis  \rightarrow \Gs$ with $\mapGs(\Thetas) = \gs_{\Thetas}$, where the notation $\gs_{\Thetas}(\xs) = \hsL_{\Thetas} (\xs)$ refers to the network output function defined in \eqref{eq_gs_gt_defn} by clarifying its dependence on the network parameters. Similarly to \eqref{eq_hsl_cont}, it is easy to observe that $\hsL_{\Thetas} (\xs) $ is a continuous function of $\Thetas$ and for any sequence  $\{ \Thetas_n \} $ converging to an element $\Thetas_*  \in \Phis$
\begin{equation*}
\begin{split}
\lim_{n \rightarrow \infty} \gs_{\Thetas_n}(\xs) 
 = \lim_{n \rightarrow \infty} \hsL_{\Thetas_n} (\xs)  = \hsL_{\Thetas_*} (\xs) 
 =\gs_{\Thetas_*}(\xs) 
\end{split}
\end{equation*}
uniformly. Hence,
\begin{equation*}
\begin{split}
\lim_{n \rightarrow \infty}  \ds(\gs_{\Thetas_n} , \gs_{\Thetas_*}   ) 
&= \lim_{n \rightarrow \infty} \sup_{\xs \in \Xs} 
 \|  \gs_{\Thetas_n}(\xs)  - \gs_{\Thetas_*}(\xs)  \| \\
 &=  \sup_{\xs \in \Xs} \lim_{n \rightarrow \infty}  \|  \gs_{\Thetas_n}(\xs)  - \gs_{\Thetas_*}(\xs)  \|
 =0.
\end{split}
\end{equation*}
Hence, the mapping $\mapGs: \Phis  \rightarrow \Gs$ is continuous. Then, from the compactness of $\Phis $, it follows that the function space $\Gs$ is compact as well.
\end{proof}


\section{Proof of Lemma \ref{lem_cov_num_Fs_Ft}}
\label{pf_lem_cov_num_Fs_Ft}

\begin{proof}
We obtain the bound only for the source domain, as the derivation for the target domain is identical. Our proof is based on constructing an $\epsilon $-cover for the compact metric space $\Fs$. For two mappings $\fs_1, \fs_2 \in \Fs$ defined respectively by the parameter vectors $\Thetas_1, \Thetas_2$ we have
\begin{equation}
\label{eq_dX_ito_dhsl}
\begin{split}
(\dXs(\fs_1, \fs_2) )^2
&= \sup_{\xs \in \Xs}  \| \fs_1(\xs) - \fs_2(\xs)  \|_{\X}^2 \\
&= \sup_{\xs \in \Xs} \sum_{\lay=1}^{\numL-1}
	\| \phil(\hsl_{\Thetas_1} (\xs))   - \phil(\hsl_{\Thetas_2} (\xs))  \|_{\Xl}^2 \\
&= \sup_{\xs \in \Xs}  \sum_{\lay=1}^{\numL-1}
\krl \left(  \hsl_{\Thetas_1} (\xs) , \hsl_{\Thetas_1} (\xs) \right)
 - 2 \krl \left(  \hsl_{\Thetas_1} (\xs) , \hsl_{\Thetas_2} (\xs) \right) \\
& \quad \quad +\krl \left(  \hsl_{\Thetas_2} (\xs) , \hsl_{\Thetas_2} (\xs) \right) \\
&\leq 
\sup_{\xs \in \Xs}  \sum_{\lay=1}^{\numL-1}
\left |   \krl \left(   \hsl_{\Thetas_1} (\xs) ,  \hsl_{\Thetas_1} (\xs) \right)
- 
\krl \left(  \hsl_{\Thetas_1} (\xs) ,  \hsl_{\Thetas_2} (\xs) 
  \right )
  \right | \\
&  \quad \quad 
+
  \left |   \krl \left(   \hsl_{\Thetas_2} (\xs) ,  \hsl_{\Thetas_2} (\xs) \right)
- 
\krl \left(  \hsl_{\Thetas_1} (\xs) , \hsl_{\Thetas_2} (\xs)
  \right )
  \right | \\
 &  \leq   \sup_{\xs \in \Xs} \sum_{\lay=1}^{\numL-1}
2 \Lk 
 \|   \hsl_{\Thetas_1} (\xs)  -  \hsl_{\Thetas_2} (\xs)   \|
\end{split}
\end{equation}
where the last inequality is due to the Lipschitz continuity of the kernels $\krl$.  We next construct a cover for the set of parameter vectors $\Thetas$, which will define a cover for $\Fs$ using the relation in \eqref{eq_dX_ito_dhsl}. From \eqref{eq_bnd_Thetaij} the network parameter vectors of layer $\lay$ are in the compact set
\begin{equation}
\label{eq_defn_thetasetl}
\begin{split}
\ThetaSetl=\{ \Thetal = [\Wl \ \bl] \in \R^{\dl \times (\dlm+1)} :  | \Wl_{ij} | \leq \Bnet, | \bl_{i} | \leq \Bnet,  \,\forall i, j, \lay \}.
\end{split}
\end{equation}
Then there exists a cover of $\ThetaSetl$ consisting of open balls around a set $\grid^\lay = \{ \Thetal_m \}_{m=1}^{\Kl}$ of regularly sampled grid points, with a distance of $\radTheta $ between adjacent grid centers in each dimension. The maximal overall distance between two adjacent grid centers is then $\radTheta  \sqrt{\dl(\dlm+1)}$. Hence, the distance between any parameter vector $\Thetal \in \ThetaSetl $ and the nearest grid center $\Thetal_m $ is at most
\begin{equation*}
\begin{split}
\frac{\radTheta  \sqrt{\dl(\dlm+1)}}{2}
\end{split}
\end{equation*}
with the number of balls in the cover being
\begin{equation*}
\begin{split}
\Kl = \left( \frac{2 \Bnet}{\radTheta}+1  \right)^{ \dl(\dlm+1) }.
\end{split}
\end{equation*}
From the Cartesian product of the grid centers at layers $\lay=1, \dots, \numL-1$, we then obtain a product grid
\begin{equation}
\label{eq_grid_Fs}
\begin{split}
\grid= \grid^1 \times \dots \times \grid^{\numL-1}  =  \{ \boldsymbol{\Theta}_k  \}_{k=1}^{\Kgen^1 \dots \ \Kgen^{\numL-1}}
\end{split}
\end{equation}
which defines a cover for the overall parameter space
\begin{equation*}
\begin{split}
\PhiCom&=\{ \ThetaCom=(\ThetaCom ^{1}, \dots, \ThetaCom ^{\numL-1}) :  \ |\ThetaCom^{\lay}_{ij}|  \leq \Bnet, \,\forall i, j, \lay \} \\
\end{split}
\end{equation*}
consisting of 
\begin{equation*}
\begin{split}
\Kgen_\grid = \prod_{\lay=1}^{\numL-1}  \Kl = \prod_{\lay=1}^{\numL-1}  \left( \frac{2 \Bnet}{\radTheta}+1  \right)^{ \dl(\dlm+1) }
\end{split}
\end{equation*}
balls. Then for any $\fs \in \Fs$ with parameters $\Thetas$, there exists some $\fs_k \in \Fs$ with parameters $\boldsymbol{\Theta}_k=(\boldsymbol{\Theta}_{k}^1, \boldsymbol{\Theta}_{k}^2, \dots, \boldsymbol{\Theta}_{k}^{L-1}) \in \grid $ in the product grid such that
\begin{equation}
\label{eq_dist_thetasl_thetakl}
\begin{split}
\| \Thetasl -  \boldsymbol{\Theta}_{k}^\lay \| < \radTheta  \sqrt{\dl(\dlm+1)}.
\end{split}
\end{equation}
For any $\xs \in \Xs$, the distance between the $\lay$-th layer features of these parameters can be bounded as
\begin{equation}
\label{eq_bnd_hsl_thetas_thetak}
\begin{split}
& \| \hsl_{\Thetas}(\xs) - \hidl_{\boldsymbol{\Theta}_k}(\xs)  \|   
= \left \|  \actl \left( \Wsl \hslm_{\Thetas}(\xs) + \bsl \right) 
- 
\actl \left( \Wl_k \, \hidlm_{\boldsymbol{\Theta}_k}(\xs) + \bl_k \right)
 \right \| \\
& \leq
 \Leta \left \| \Wsl \hslm_{\Thetas}(\xs) + \bsl 
 -
 \Wl_k \, \hidlm_{\boldsymbol{\Theta}_k}(\xs) - \bl_k
   \right \| \\
&  = 
 \Leta \left \| \Wsl \hslm_{\Thetas}(\xs) 
 - \Wsl  \hidlm_{\boldsymbol{\Theta}_k}(\xs)
 + \Wsl  \hidlm_{\boldsymbol{\Theta}_k}(\xs)
 -  \Wl_k \, \hidlm_{\boldsymbol{\Theta}_k}(\xs)  
 + \bsl - \bl_k
   \right \| \\
  & \leq
 \Leta 
 \|  \Wsl \|  \,  \|   \hslm_{\Thetas}(\xs)  - \hidlm_{\boldsymbol{\Theta}_k}(\xs)  \|
 +\Leta
 \|  \Wsl  - \Wl_k \| \, \| \hidlm_{\boldsymbol{\Theta}_k}(\xs) \|
 +\Leta
 \| \bsl - \bl_k \|
\end{split}
\end{equation}
where $\Wl_k$, $\bl_k$, and $\hidlm_{ \boldsymbol{\Theta}_k }$ denote the $\lay$-th layer network parameters and features generated by the parameter vector  $\boldsymbol{\Theta}_{k}$; and $\|  \cdot \|$ and $\| \cdot \|_F$ respectively denote the operator norm and the Frobenius norm of a matrix. From \eqref{eq_defn_thetasetl} and \eqref{eq_dist_thetasl_thetakl}, we have 
\begin{equation*}
\begin{split}
\| \Wsl \| & \leq \| \Wsl \|_F \leq \Bnet \sqrt{\dl \dlm} \\
 \|  \Wsl  - \Wl_k \| 
 & \leq  \|  \Wsl  - \Wl_k \|_F    <  \radTheta \sqrt{\dl \dlm} \\
 \|  \bsl - \bl_k  \| 
 & <   \radTheta \sqrt{\dl}.
\end{split}
\end{equation*}
These bounds together with the inequality in \eqref{eq_bnd_hsl_thetas_thetak} yield
\begin{equation}
\label{eq_hsl_thetas_thetak_xs_bnd}
\begin{split}
\| \hsl_{\Thetas}(\xs) - \hidl_{\boldsymbol{\Theta}_k}(\xs)  \|   
&< 
 \Leta 
 \Bnet \sqrt{\dl \dlm} \,  \|   \hslm_{\Thetas}(\xs)  - \hidlm_{\boldsymbol{\Theta}_k}(\xs)  \| \\
 &+\Leta
 \radTheta \sqrt{\dl \dlm} \, \| \hidlm_{\boldsymbol{\Theta}_k}(\xs) \|
 +\Leta
 \radTheta \sqrt{\dl}.
\end{split}
\end{equation}
In order to study \eqref{eq_hsl_thetas_thetak_xs_bnd}, we first obtain an upper bound on the term $\| \hidl_{\boldsymbol{\Theta}_k}(\xs) \|$. Notice that for the condition \eqref{eq_bnd_act_value}, we simply have 
\begin{equation}
\label{eq_bnd_hidl_ito_beta}
\begin{split}
\| \hidl_{\boldsymbol{\Theta}_k}(\xs)  \| 
&=
\| \actl \left(  \Wl    \hidlm_{\boldsymbol{\Theta}_k}(\xs)  + \bl  \right) \| 
= \left( 
\sum_{i=1}^\dl
 \left( \actl_i  ( \Wl    \hidlm_{\boldsymbol{\Theta}_k}(\xs)  + \bl  ) \right)^2
  \right)^{1/2} \\
&  \leq \Beta \sqrt{\dl}.
 \end{split}
\end{equation}
Next, for the condition \eqref{eq_bnd_act_op} we have
\begin{equation*}
\begin{split}
\|  \hidz_{\boldsymbol{\Theta}_k}(\xs) \| &=  \| \xs \| \leq \Binp \\
\|  \hidone_{\boldsymbol{\Theta}_k}(\xs) \|  & = \| \actone \left( \Wone \hidz_{\boldsymbol{\Theta}_k}(\xs) + \bone \right) \|
\leq 
\Bopeta \, \| \Wone \hidz_{\boldsymbol{\Theta}_k}(\xs) + \bone \| \\
& \leq
\Bopeta \, (  \| \Wone \|  \| \hidz_{\boldsymbol{\Theta}_k}(\xs) \| +  \| \bone \| )
\leq \Bopeta  \Bnet \sqrt{\done \dz}  \Binp +  \Bopeta \Bnet \sqrt{\done}
\end{split}
\end{equation*}
for layers $\lay=0$ and $\lay=1$. For $\lay \geq 2$, one can similarly establish a recursive relation between the parameter vectors of layers $\lay$ and $\lay-1$, which yields 
\begin{equation*}
\begin{split}
\|  \hidl_{\boldsymbol{\Theta}_k}(\xs) \|  
&\leq 
\Bopeta \, \left(  \| \Wl \|  \| \hidlm_{\boldsymbol{\Theta}_k}(\xs) \| +  \| \bl \| \right) \\
&\leq 
\Bopeta  \Bnet \sqrt{\dl \dlm}  \| \hidlm_{\boldsymbol{\Theta}_k}(\xs) \|  +  \Bopeta \Bnet \sqrt{\dl} \\
&\leq 
(\Bopeta \Bnet) ^l  (\Binp \sqrt{\dz}+1) \sqrt{\done}
\prod_{k=1}^{\lay-1} \sqrt{\dkplone \dk} \\
&+
\sum_{i=2}^{\lay-1} (\Bopeta \Bnet)^{\lay+1-i} \sqrt{\di}
\prod_{k=1}^{\lay-1} \sqrt{\dkplone \dk}
+
\Bopeta \Bnet \sqrt{\dl}.
\end{split}
\end{equation*}
Hence, combining this with \eqref{eq_bnd_hidl_ito_beta}, we get 
\begin{equation}
\label{eq_bnd_norm_Rl}
\begin{split}
\|  \hidl_{\boldsymbol{\Theta}_k}(\xs) \|  \leq \Bdiml
\end{split}
\end{equation}
for $\lay=2, \dots, \numL-1 $, where $\Bdiml$ is the constant defined in Lemma \ref{lem_cov_num_Fs_Ft}. Using this in \eqref{eq_hsl_thetas_thetak_xs_bnd}, we obtain
\begin{equation}
\label{eq_hsl_thetas_thetak_v2}
\begin{split}
\| \hsl_{\Thetas}(\xs) - \hidl_{\boldsymbol{\Theta}_k}(\xs)  \|   
&< 
 \Leta 
 \Bnet \sqrt{\dl \dlm} \,  \|   \hslm_{\Thetas}(\xs)  - \hidlm_{\boldsymbol{\Theta}_k}(\xs)  \| \\
 &+\Leta
 \radTheta \sqrt{\dl \dlm} \, \Bdimlm 
 +\Leta
 \radTheta \sqrt{\dl}.
\end{split}
\end{equation}
For layer $\lay=1$, we have
\begin{equation*}
\begin{split}
\| \hsone_{\Thetas}(\xs) - \hidone_{\boldsymbol{\Theta}_k}(\xs)  \|   
&< 
\Leta 
 \Bnet \sqrt{\done \dz} \,  \|   \hsz_{\Thetas}(\xs)  - \hidz_{\boldsymbol{\Theta}_k}(\xs)  \| \\
 &+\Leta
 \radTheta \sqrt{\done \dz} \, \Bdimz 
 +\Leta
 \radTheta \sqrt{\done}\\
 &=\Leta
 \radTheta \sqrt{\done \dz} \, \Bdimz 
 +\Leta
 \radTheta \sqrt{\done} 
\end{split}
\end{equation*}
since $ \hsz_{\Thetas}(\xs)  = \hidz_{\boldsymbol{\Theta}_k}(\xs) = \xs$. This relation together with the recursive inequality in \eqref{eq_hsl_thetas_thetak_v2} yields
\begin{equation}
\label{eq_bnd_norm_Ql_delta}
\begin{split}
\| \hsl_{\Thetas}(\xs) - \hidl_{\boldsymbol{\Theta}_k}(\xs)  \|   
&< \radTheta 
\bigg(
(\Leta \Bdimlm \sqrt{\dl \dlm}+ \Leta \sqrt{\dl}) \\
&+
\sum_{i=1}^{\lay-1}
(\Leta \Bdimim \sqrt{\di \dimm}
+ \Leta \sqrt{\di})
\prod_{k=i+1}^{\lay}
\Leta \Bnet \sqrt{\dk \dkm}
\bigg)\\
&= \BQdiml  \radTheta
\end{split}
\end{equation}
for $\lay=1, \dots, \numL-1 $. Hence, we have shown that for any $\fs \in \Fs$ with parameters $\Thetas$, there exists some $\fs_k \in \Fs$ with parameters $\boldsymbol{\Theta}_k \in \grid $ in the product grid such that 
\begin{equation*}
\begin{split}
\| \hsl_{\Thetas}(\xs) - \hidl_{\boldsymbol{\Theta}_k}(\xs)  \|    < \BQdiml  \radTheta
\end{split}
\end{equation*}
for any $\xs \in \Xs$. We can now use this in \eqref{eq_dX_ito_dhsl} to bound the distance $\dXs(\fs, \fs_k)$ as
\begin{equation}
\begin{split}
(\dXs(\fs, \fs_k) )^2
& \leq   \sup_{\xs \in \Xs} \sum_{\lay=1}^{\numL-1}
2 \Lk 
 \|   \hsl_{\Thetas} (\xs)  -  \hidl_{\boldsymbol{\Theta}_k} (\xs)   \|
< 
2 \Lk  \radTheta \sum_{\lay=1}^{\numL-1} \BQdiml
= 2 \Lk  \radTheta \BQ.
\end{split}
\end{equation}
Therefore, the set $\{ \fs_k \}_{k=1}^{\Kgen_\grid} \subset \Fs$ provides a cover for $\Fs$  with covering radius $ \sqrt{2 \Lk  \radTheta \BQ}$. In order to obtain a covering radius of $\epsilon=  \sqrt{2 \Lk  \radTheta \BQ}$, we set 
\begin{equation*}
\begin{split}
\radTheta = \frac{\epsilon^2}{2 \Lk \BQ}
\end{split}
\end{equation*}
which provides a grid consisting of
\begin{equation*}
\begin{split}
\prod_{\lay=1}^{\numL-1}  \Kl 
= \prod_{\lay=1}^{\numL-1}  \left( \frac{4 \Bnet  \Lk \BQ}{\epsilon^2}+1  \right)^{ \dl(\dlm+1) }
\end{split}
\end{equation*}
balls that covers $\Fs$. Hence, we obtain the upper bound 
\begin{equation*}
\begin{split}
\N (  \Fs, \epsilon, \dXs)  \leq \prod_{\lay=1}^{\numL-1}  \left( \frac{4 \Bnet  \Lk \BQ}{\epsilon^2}+1  \right)^{ \dl(\dlm+1) }
\end{split}
\end{equation*}
for the covering number stated in the lemma. 
\end{proof}


\section{Proof of Lemma \ref{lem_cov_num_HFs_HFt}}
\label{pf_lem_cov_num_HFs_HFt}

\begin{proof}
We prove the statement of the lemma only for the source function space $\Hs \circ \Fs$, as the derivations for the target domain are identical. In order to bound the covering number for  $\Hs \circ \Fs$, we proceed as in the proof of Lemma \ref{lem_cov_num_Fs_Ft} and extend the grid construction in \eqref{eq_grid_Fs} to include layer $\numL$ as well. This defines a grid 
\begin{equation}
\label{eq_grid_HFs}
\begin{split}
\grid_{\Hs \circ \F} = \grid^1 \times \dots \times \grid^{\numL}  =  \{ \boldsymbol{\Theta}_k  \}_{k=1}^{\Kgen^1 \dots \ \Kgen^{\numL}}
\end{split}
\end{equation}
providing a cover for the parameter space
\begin{equation*}
\begin{split}
\PhiCom_{\Hs \circ \F} &=\{ \ThetaCom=(\ThetaCom ^{1}, \dots, \ThetaCom ^{\numL}) :  \ |\ThetaCom^{\lay}_{ij}|  \leq \Bnet, \,\forall i, j, \lay \} \\
\end{split}
\end{equation*}
consisting of 
\begin{equation}
\label{eq_num_balls_HsF}
\begin{split}
 \prod_{\lay=1}^{\numL}  \Kl = \prod_{\lay=1}^{\numL}  \left( \frac{2 \Bnet}{\radTheta}+1  \right)^{ \dl(\dlm+1) }
\end{split}
\end{equation}
balls. Then for any   $\gs \in \Hs \circ \Fs$ with network parameters $\Thetas$, there exists some $\gs_k \in \Hs \circ \Fs$ with network parameters $\boldsymbol{\Theta}_k=(\boldsymbol{\Theta}_{k}^1, \boldsymbol{\Theta}_{k}^2, \dots, \boldsymbol{\Theta}_{k}^{L}) \in \grid_{\Hs \circ \F} $ in the grid such that
\begin{equation*}
\begin{split}
\| \Thetasl -  \boldsymbol{\Theta}_{k}^\lay \| < \radTheta  \sqrt{\dl(\dlm+1)}
\end{split}
\end{equation*}
for $\lay=1, \dots, \numL$. Proceeding in a similar fashion to the derivations in \eqref{eq_bnd_hsl_thetas_thetak} and \eqref{eq_hsl_thetas_thetak_xs_bnd}, we obtain
\begin{equation}
\begin{split}
 \| \hsL_{\Thetas}(\xs) - \hidL_{\boldsymbol{\Theta}_k}(\xs)  \|   
&   \leq
 \Leta 
 \|  \WsL \|  \,  \|   \hsLm_{\Thetas}(\xs)  - \hidLm_{\boldsymbol{\Theta}_k}(\xs)  \| \\
 &+\Leta
 \|  \WsL  - \WL_k \| \, \| \hidLm_{\boldsymbol{\Theta}_k}(\xs) \| 
 +\Leta
 \| \bsL - \bL_k \| \\
&< 
 \Leta 
 \Bnet \sqrt{\dL \dLm} \,  \|   \hsLm_{\Thetas}(\xs)  - \hidLm_{\boldsymbol{\Theta}_k}(\xs)  \| \\
 &+\Leta
 \radTheta \sqrt{\dL \dLm} \, \| \hidLm_{\boldsymbol{\Theta}_k}(\xs) \|
 +\Leta
 \radTheta \sqrt{\dL}
\end{split}
\end{equation}
for any $\xs \in \Xs$. Combining this inequality with the bounds in \eqref{eq_bnd_norm_Rl} and \eqref{eq_bnd_norm_Ql_delta} gives
\begin{equation*}
\begin{split}
 \| \hsL_{\Thetas}(\xs) - \hidL_{\boldsymbol{\Theta}_k}(\xs)  \|   
&   <
 \Leta 
 \Bnet \sqrt{\dL \dLm} \,  \BQdimLm  \radTheta \\
 &+\Leta
 \radTheta \sqrt{\dL \dLm} \, \BdimLm
 +\Leta
 \radTheta \sqrt{\dL} \\
 &= \BQdimL \radTheta.
\end{split}
\end{equation*}
Recalling the definition of the distance $\ds $ in \eqref{eq_defn_ds_dt}, we then have
\begin{equation*}
\begin{split}
\ds(\gs, \gs_k) &= \sup_{\xs \in \Xs} \| \gs(\xs) - \gs_k(\xs)  \| 
=  \sup_{\xs \in \Xs} \|  \hsL_{\Thetas} (\xs)  -  \hidL_{\boldsymbol{\Theta}_k} (\xs)  \| 
< \BQdimL \radTheta.
\end{split}
\end{equation*}
%
 %
Hence, the grid $\grid_{\Hs \circ \F}$ in \eqref{eq_grid_HFs} provides a cover for $ \Hs \circ \Fs$ with covering radius $\BQdimL \radTheta$. For a covering radius of $\epsilon $, we set $\epsilon=\BQdimL \radTheta$, which results in a cover with 
\begin{equation}
\begin{split}
\prod_{\lay=1}^{\numL}  \left( \frac{2 \Bnet \BQdimL}{\epsilon}+1  \right)^{ \dl(\dlm+1) }
\end{split}
\end{equation}
balls due to \eqref{eq_num_balls_HsF}. We thus get the covering number upper bound
\begin{equation*}
\begin{split}
\N( \Hs \circ \Fs, \epsilon, \ds) \leq  \prod_{\lay=1}^{\numL}  \left( \frac{2 \Bnet \BQdimL}{\epsilon}+1  \right)^{ \dl(\dlm+1) } \\
\end{split}
\end{equation*}
stated in the lemma.

\end{proof}


\section{Proof of Corollary \ref{cor_covnum_rate}}
\label{pf_cor_covnum_rate}

\begin{proof}
In order to analyze the dependence of $\N (  \Fs, \epsilon, \dXs) $ on $\dcom$ and $\numL$, we first study how the term $\Bdiml$ in Lemma \ref{lem_cov_num_Fs_Ft} grows with the dimension $\dcom$ and the number of layers $\numL$.
For condition \eqref{eq_bnd_act_value}, we have 
\[
\Bdiml =\Beta \sqrt{\dl} = O(\dcom^{1/2}).
\]
For condition \eqref{eq_bnd_act_op}, representing the relevant constant terms as $c$ for simplicity, we have
\begin{equation*}
\begin{split}
\Bdiml = O\big( (c\dcom)^\lay \big).
\end{split}
\end{equation*}
We next study the term $\BQdiml$ in \eqref{eq_defn_Ql}. For condition  \eqref{eq_bnd_act_value}, we obtain
\begin{equation*}
\begin{split}
\BQdiml = O( c^{\lay-1} \, \dcom^{\lay+ \frac{1}{2}})
\end{split}
\end{equation*}
which results in
\begin{equation}
\label{eq_BQ_bnd_35}
\begin{split}
\BQ = O(c^{\numL-2} \dcom^{\numL-\frac{1}{2}}).
\end{split}
\end{equation}
Meanwhile, condition \eqref{eq_bnd_act_op} yields
\begin{equation*}
\begin{split}
\BQdiml = O\big( (\lay-1) \, c^{\lay-1} \,  \dcom^\lay \big)
\end{split}
\end{equation*}
resulting in
\begin{equation}
\label{eq_BQ_bnd_36}
\begin{split}
\BQ = O\big( (\numL-2) \, c^{\numL-2} \,  \dcom^{\numL-1} \big).
\end{split}
\end{equation}
For simplicity, we may combine the results in \eqref{eq_BQ_bnd_35} and \eqref{eq_BQ_bnd_36} through a slightly more pessimistic but brief common upper bound as
\begin{equation*}
\begin{split}
\BQ = O \big( \numL \, c^{\numL-2} \dcom^\numL   \big)
\end{split}
\end{equation*}
which is valid for both of the conditions in \eqref{eq_bnd_act_value} and \eqref{eq_bnd_act_op}.  Then, from the expressions of the covering numbers $\N (  \Fs, \epsilon, \dXs) $ and $\N (  \Ft, \epsilon, \dXt) $ in Lemma \ref{lem_cov_num_Fs_Ft}, we conclude
\begin{equation*}
\begin{split}
\N (  \Fs, \epsilon, \dXs)  
= O\left( \left( \frac{  c\BQ}{\epsilon^2}  \right)^{\dcom^2 \numL} \right)
= O \left( \left( \frac{\numL}{\epsilon} \right) ^{\dcom^2 \numL} \, (c \dcom)^{\dcom^2 \numL^2}  \right)
\end{split}
\end{equation*}
where we have taken the liberty to replace the $\epsilon^2$ term in the denominator with  $\epsilon$ for simplicity, as they will lead to equivalent bounds. Similarly,
\begin{equation*}
\begin{split}
\N (  \Ft, \epsilon, \dXt)  
= O \left( \left( \frac{\numL}{\epsilon} \right) ^{\dcom^2 \numL}  (c \dcom)^{\dcom^2 \numL^2}  \right).
\end{split}
\end{equation*}

We next analyze the covering number  $\N ( \Hs \circ \Fs, \epsilon, \ds) $ for the hypothesis space $ \Hs \circ \Fs$. For condition  \eqref{eq_bnd_act_value}, we have
\begin{equation*}
\begin{split}
\BQdimL = O( c^{\numL-1} \, \dcom^{\numL+ \frac{1}{2}})
\end{split}
\end{equation*}
which gives from Lemma \ref{lem_cov_num_HFs_HFt}
\begin{equation}
\label{eq_covnum_Hfs_35}
\begin{split}
\N ( \Hs \circ \Fs, \epsilon, \ds)  
= O\left( \left(\frac{c \BQdimL}{\epsilon} \right)^{\dcom^2 \numL} \right)
= O\left(  \frac{  (c \dcom)^{\dcom^2 \numL^2}  }{\epsilon^{\dcom^2 \numL}}  \right)
\end{split}
\end{equation}
if the $\dcom^2 \numL/2$ term added to the $\dcom^2 \numL^2$ term in the exponent is ignored for simplicity. Next, for condition \eqref{eq_bnd_act_op} we obtain
\begin{equation*}
\begin{split}
\BQdimL = O\big( (\numL-1) \, c^{\numL-1} \,  \dcom^\numL \big)
\end{split}
\end{equation*}
resulting in
\begin{equation}
\label{eq_covnum_Hfs_36}
\begin{split}
\N ( \Hs \circ \Fs, \epsilon, \ds) 
= O\left( \left(\frac{c \BQdimL}{\epsilon} \right)^{\dcom^2 \numL} \right)
 = O\left( \left( \frac{  \numL }{\epsilon}  \right)^{\dcom^2 \numL}   (c \dcom)^{\dcom^2 \numL^2} \right).
\end{split}
\end{equation}
Combining the bounds in \eqref{eq_covnum_Hfs_35} and \eqref{eq_covnum_Hfs_36}, we arrive at the common upper bound
\begin{equation*}
\begin{split}
\N ( \Hs \circ \Fs, \epsilon, \ds) 
 = O\left( \left( \frac{  \numL }{\epsilon}  \right)^{\dcom^2 \numL}   (c \dcom)^{\dcom^2 \numL^2} \right)
\end{split}
\end{equation*}
which covers both conditions. Identical derivations for the target domain yield
\begin{equation*}
\begin{split}
\N ( \Hs \circ \Ft, \epsilon, \dt) 
 = O\left( \left( \frac{  \numL }{\epsilon}  \right)^{\dcom^2 \numL}   (c \dcom)^{\dcom^2 \numL^2} \right).
\end{split}
\end{equation*}

\end{proof}


\section{Proof of Theorem \ref{thm_main_result_da_mmd}}
\label{pf_thm_main_result_da_mmd}

\begin{proof}
We first notice that, owing to Lemma \ref{lem_fs_ft_measble}, we can analyze MMD-based domain adaptation networks within the setting of Theorem \ref{thm:main_result_mmd}. The compactness of the function spaces $\Fs$, $\Ft$, $\Hs \circ \Fs$, and $\Hs \circ \Ft$ follow from Assumptions \ref{assum_Ax_Atheta}-\ref{assum_Lk_Leta} due to Lemma \ref{lem_Fs_Ft_HFs_HFt_comp}. Assumptions \ref{assum_HF_comp_Ll_Al} and \ref{assum_Fs_Ft_compact} are thereby satisfied; hence, the statement of Theorem \ref{thm:main_result_mmd} applies to the current setting in consideration.

We recall from Theorem \ref{thm:main_result_mmd} that the expected target loss in \eqref{eq_accuracy_thm4} is attained with probability at least
\begin{equation}
\label{eq_prob_expr_thm4}
\begin{split}
1 &- 2 \N( \Hs \circ \Ft, \frac{\epsilon}{8 \alpha \Lls}, \dt) e^{-\frac{\Mt \epsilon^2}{8 \alpha^2 \bls^2}} 
-2 \N( \Hs \circ \Fs, \frac{\epsilon}{8 (1-\alpha) \Lls}, \ds) e^{-\frac{\Ms \epsilon^2}{8 (1-\alpha)^2 \bls^2}} \\
& - \N(\Fs, \frac{\epsilon}{8}, \dXs) \exp(-\as(\Ns, \epsilon)) 
- \N(\Ft, \frac{\epsilon}{8}, \dXt) \exp(-\at(\Nt, \epsilon)).
\end{split}
\end{equation}
Our proof is then based on identifying the rate at which the number of samples should grow with $\numL$ and $\dcom$ so that each one of the terms subtracted from 1 in the expression \eqref{eq_prob_expr_thm4} remains fixed. This will in return guarantee that the generalization gap of $O(\epsilon)$ in \eqref{eq_accuracy_thm4} be attained with high probability.
 
We begin with the term 
$\N(\Fs, \frac{\epsilon}{8}, \dXs) \exp(-\as(\Ns, \epsilon)) $. Recalling the definition of  $\as(\Ns, \epsilon)$ from Lemma \ref{lem:unif_bnd_D_hD}, we have
\begin{equation*}
\begin{split}
\as(\Ns, \epsilon) = \bm{\theta}(\Ns \epsilon^2)
\end{split}
\end{equation*}
where we use the notation $\bm{\theta}(\cdot)$ to refer to asymptotic tight bounds. Combining this with Corollary \ref{cor_covnum_rate}, we obtain 
\begin{equation*}
\begin{split}
\N(\Fs, \frac{\epsilon}{8}, \dXs) \exp(-\as(\Ns, \epsilon)) 
&=
O\left( 
\left( \frac{  \numL }{\epsilon}  \right)^{\dcom^2 \numL}   (c \dcom)^{\dcom^2 \numL^2} 
\exp(- \Ns \epsilon^2)
\right) \\
&=O\left(
\exp \left(
\dcom^2 \numL \, \log\left( \frac{  \numL }{\epsilon}  \right)
+ 
\dcom^2 \numL^2 \log(c \dcom) 
- 
\Ns \epsilon^2
\right)
\right).
\end{split}
\end{equation*}
We conclude that the total number $\Ns$ of source samples required to ensure a lower bound on the probability expression \eqref{eq_prob_expr_thm4} scales as
\begin{equation*}
\begin{split}
\Ns = O \left(
\frac{\dcom^2 \numL \, \log\left( \frac{  \numL }{\epsilon}  \right)
+ 
\dcom^2 \numL^2 \log(\dcom) }
{\epsilon^2}
\right),
\end{split}
\end{equation*}
yielding the sample complexity stated in the theorem. An identical derivation based on bounding the term $ \N(\Ft, \frac{\epsilon}{8}, \dXt) \exp(-\at(\Nt, \epsilon))$ shows that $\Nt $ has the same sample complexity.

Next, we examine the terms involving the number of labeled samples. Proceeding similarly, we get
\begin{equation*}
\begin{split}
 \N( \Hs \circ \Ft, \frac{\epsilon}{8 \alpha \Lls}, \dt) &e^{-\frac{\Mt \epsilon^2}{8 \alpha^2 \bls^2}}
 =
O\left( 
\left( \frac{  \numL \alpha }{\epsilon}  \right)^{\dcom^2 \numL}   (c \dcom)^{\dcom^2 \numL^2} 
\exp \left(- \frac{\Mt \epsilon^2}{\alpha^2} \right)
\right) \\
&=O\left(
\exp \left(
\dcom^2 \numL \, \log\left( \frac{  \numL \alpha }{\epsilon}  \right)
+ 
\dcom^2 \numL^2 \log(c \dcom) 
- 
\frac{\Mt \epsilon^2}{\alpha^2}
\right)
\right).
\end{split}
\end{equation*}
Recalling that $0\leq \alpha \leq 1$, we conclude that upper bounding the choice of the weight parameter $\alpha$ by the rate
\begin{equation*}
\begin{split}
\alpha
= O \left(
\left(
\frac
{\Mt \epsilon^2}
{\dcom^2 \numL \log \left( \frac{  \numL }{\epsilon} \right)  \, 
+ 
\dcom^2 \numL^2 \log(\dcom) }
\right)^{1/2}
\right)
\end{split}
\end{equation*}
ensures that the probability term $ \N( \Hs \circ \Ft, \frac{\epsilon}{8 \alpha \Lls}, \dt) e^{-\frac{\Mt \epsilon^2}{8 \alpha^2 \bls^2}}$ remain bounded.

Finally, for the number of labeled samples in the source domain, we have
\begin{equation*}
\begin{split}
 \N( \Hs \circ \Fs, & \, \frac{\epsilon}{8 (1-\alpha) \Lls},   \ds) e^{-\frac{\Ms \epsilon^2}{8 (1-\alpha)^2 \bls^2}} \\
 &=
O\left( 
\left( \frac{  \numL (1-\alpha) }{\epsilon}  \right)^{\dcom^2 \numL}   (c \dcom)^{\dcom^2 \numL^2} 
\exp \left(- \frac{\Ms \epsilon^2}{(1-\alpha)^2} \right)
\right) \\
&=O\left(
\exp \left(
\dcom^2 \numL \, \log\left( \frac{  \numL (1-\alpha) }{\epsilon}  \right)
+ 
\dcom^2 \numL^2 \log(c \dcom) 
- 
\frac{\Ms \epsilon^2}{(1-\alpha)^2}
\right)
\right).
\end{split}
\end{equation*}
Recalling again the bound $0\leq 1-\alpha \leq 1$, we observe that the sample complexity 
\begin{equation*}
\begin{split}
\Ms = 
 O \left(
\frac{\dcom^2 \numL \, \log\left( \frac{  \numL }{\epsilon}  \right)
+ 
\dcom^2 \numL^2 \log(\dcom) }
{\epsilon^2}
\right)
\end{split}
\end{equation*}
ensures a lower bound on the probability expression \eqref{eq_prob_expr_thm4}, which concludes the proof of the theorem.
\end{proof}


\section{Derivation of the bound and the Lipschitz constant for the cross-entropy loss}
\label{sec_app_crossent}

We first discuss the magnitude bound $\bls$ for the widely used cross-entropy loss function. Let $\y_1, \y_2 \in \Y \subset \R^m$ be two nonnegative label vectors in the label set $\Y=[0,1] \times \cdots \times [0,1] \subset \R^m$. In its na\"ive form, the cross-entropy loss between $\y_1$ and  $\y_2$  is given by
\begin{equation}
\label{eq_crossent_naive}
\begin{split}
\loss(\y_1, \y_2) = - \sum_{k=1}^m \log(\y_1(k)) \, \y_2(k)
\end{split}
\end{equation}
where $\y(k)$ denotes the $k$-th entry of the vector $\y$. While the original form \eqref{eq_crossent_naive} of the cross-entropy loss is not bounded, often the following modification is made in order to avoid numerical issues in practical implementations
\begin{equation*}
\begin{split}
\loss(\y_1, \y_2) = - \sum_{k=1}^m \log(\y_1(k)+\delta) \, \y_2(k)
\end{split}
\end{equation*}
where $0<\delta < 1$ is a positive constant. We then have
\begin{equation*}
\begin{split}
|\loss(\y_1, \y_2) | \leq \sum_{k=1}^m | - \log(\y_1(k)+ \delta) \y_2(k)  |
\leq m \max\{| \log(\delta) | , \ \log(1+\delta)  \}.
\end{split}
\end{equation*}
Assuming that $\delta$ is very small, we get the following bound on the loss magnitude
\begin{equation*}
\begin{split}
|\loss(\y_1, \y_2) | \leq \bls \triangleq  m \, | \log(\delta) |.
\end{split}
\end{equation*}

We next derive the Lipschitz constant $\Lls$ of the cross-entropy loss function. For any $\y, \y_1, \y_2 \in \Y$ we have
\begin{equation}
\label{eq_lip_loss_step1}
\begin{split}
| \loss(\y_1, \y) -  \loss(\y_2, \y) |
&= \left |   -\sum_{k=1}^m  \log( \y_1(k) + \delta) \y(k)    
 + \sum_{k=1}^m  \log( \y_2(k) + \delta) \y(k)   \right|  \\
 &\leq \sum_{k=1}^m \left |  \, \log( \y_2(k) + \delta)  -   \log( \y_1(k) + \delta)  \, \right |.
\end{split}
\end{equation}
For any $t\geq \delta$, we have
\[
\left |  \frac{d}{dt} \log(t)   \right | = \left |  \frac{1}{t}   \right | 
\leq \frac{1}{\delta} 
\]
which gives
\begin{equation*}
\begin{split}
 \left|  \frac{  \log( \y_2(k) + \delta)  -   \log( \y_1(k) + \delta) }{\y_2(k) - \y_1(k)} \right |
 \leq  \frac{1}{\delta} 
\end{split}
\end{equation*}
due to the mean value theorem. Using this in \eqref{eq_lip_loss_step1}, we get
\begin{equation*}
\begin{split}
| \loss(\y_1, \y) -  \loss(\y_2, \y) |
 \leq \sum_{k=1}^m  \delta^{-1}  |  \y_2(k) - \y_1(k) |  
 \leq \delta^{-1} \sqrt{m} \, \|  \y_2 - \y_1 \|
\end{split}
\end{equation*}
which shows that the cross-entropy loss is Lipschitz continuous with respect to the first argument with constant
\[
\Lls \triangleq \delta^{-1} \sqrt{m} .
\]


\section{Proof of Lemma \ref{lem_ddan_hddan_dev}}
\label{pf_lem_ddan_hddan_dev}

\begin{proof}

Due to the assumption of compactness of the function classes $\Vs$ and $\Vt$, there exists an $\epsilon$-cover of each function space. Let us denote the cover numbers of $\Vs$ and $\Vt$ as
\begin{equation*}
\Ks = \N(\Vs, \epsilon, \dVs), 
\quad \quad
\Kt = \N(\Vt, \epsilon, \dVt)
\end{equation*}
respectively, and the corresponding sets of ball centers as  $\{ \vs_k \}_{k=1}^{\Ks}$ and $\{ \vt_l \}_{l=1}^{\Kt}$. Then, for any $\vs \in \Vs$ and any $\vt \in \Vt$ there exist some $\vs_k \in \Vs$ and $\vt_l \in \Vt$ such that
\begin{equation}
\label{eq_dvs_dvt_ball_rad}
\begin{split}
\dVs (\vs , \vs_k) &= \sup_{\xs \in \Xs}  | \vs(\xs) - \vs_k(\xs) | < \epsilon \\
\dVt (\vt , \vt_l) &= \sup_{\xt \in \Xt}  | \vt(\xt) - \vt_l(\xt) | < \epsilon. \\
\end{split}
\end{equation}
Let us denote
\begin{equation*}
\begin{split}
\D(\vs_k, \vt_l)  &\triangleq \left |  E[\vs_k(\xs)] - E[\vt_l(\xt)] \right |  \\
\hD(\vs_k, \vt_l) &\triangleq
\left | 
 \frac{1}{\Ns} 
  \sum_{i=1}^{\Ns} \vs_k (\xis)
  - 
  \frac{1}{\Nt} 
  \sum_{j=1}^{\Nt} \vt_l (\xjt) 
  \right |.
\end{split}
\end{equation*}
Take any $\fs \in \Fs$, $\ft \in \Ft$ and $\ddan \in \Dspace$. We have
\begin{equation}
\label{eq_ddan_hddan_dev}
\begin{split}
& |  \Ddan(\fs, \ft)  - \hDdan(\fs, \ft)  |  \\
&= 
|  \Ddan(\fs, \ft)  - \D(\vs_k, \vt_l) + \D(\vs_k, \vt_l)  
- \hD(\vs_k, \vt_l)  + \hD(\vs_k, \vt_l) 
- \hDdan(\fs, \ft)  | \\
&\leq 
|  \Ddan(\fs, \ft)  - \D(\vs_k, \vt_l)  |
+
| \D(\vs_k, \vt_l)  
- \hD(\vs_k, \vt_l)  |
+
| \hD(\vs_k, \vt_l) 
- \hDdan(\fs, \ft) | . 
\end{split}
\end{equation}
We proceed by bounding each one of the three terms at the right hand side of the inequality in \eqref{eq_ddan_hddan_dev}. The first term can be upper bounded as 
\begin{equation}
\label{eq_ddan_hddan_term1}
\begin{split}
 | \Ddan(\fs, \ft)  - \D(\vs_k, \vt_l) |
&=
\left | 
|  E[\vs(\xs) ]  - E [\vt(\xt) ]   |
- 
|  E[\vs_k(\xs)] - E[\vt_l(\xt)] | 
\right |  \\
&\leq
\left | 
 E[\vs(\xs) ]  - E [\vt(\xt) ]   
- 
  E[\vs_k(\xs)] + E[\vt_l(\xt)] 
\right |  \\
&\leq
| E[\vs(\xs) ] -   E[\vs_k(\xs)] |
+
|  E [\vt(\xt)] - E[\vt_l(\xt)] |
<
2\epsilon
\end{split}
\end{equation}
where the last inequality follows from \eqref{eq_dvs_dvt_ball_rad}. For the third term in \eqref{eq_ddan_hddan_dev}, one can similarly show that 
\begin{equation}
\label{eq_ddan_hddan_term3}
\begin{split}
| \hD(\vs_k, \vt_l) 
- \hDdan(\fs, \ft) |  < 2 \epsilon.
\end{split}
\end{equation}
We lastly study the second term in \eqref{eq_ddan_hddan_dev}. We have
\begin{equation}
\label{eq_ddan_hddan_bnd1}
\begin{split}
&
| \D(\vs_k, \vt_l)  
- \hD(\vs_k, \vt_l)  | \\
&=
\left |
\left |  E[\vs_k(\xs)] - E[\vt_l(\xt)] \right | 
- 
\left | 
 \frac{1}{\Ns} 
  \sum_{i=1}^{\Ns} \vs_k (\xis)
  - 
  \frac{1}{\Nt} 
  \sum_{j=1}^{\Nt} \vt_l (\xjt) 
  \right | 
  \right | \\
& \leq
\left |
 E[\vs_k(\xs)] - E[\vt_l(\xt)] 
- 
 \frac{1}{\Ns} 
  \sum_{i=1}^{\Ns} \vs_k (\xis)
  +
  \frac{1}{\Nt} 
  \sum_{j=1}^{\Nt} \vt_l (\xjt) 
\right | \\
&\leq
\left | 
 \frac{1}{\Ns}   \sum_{i=1}^{\Ns} \vs_k (\xis) -  E[\vs_k(\xs)]
 \right |
+
\left | 
  \frac{1}{\Nt}  \sum_{j=1}^{\Nt} \vt_l (\xjt)  - E[\vt_l(\xt)] 
\right |.
\end{split}
\end{equation}
As the domain discriminator is bounded due to Assumption \ref{assum_ddan_bdd}, from Hoeffding's inequality we have
\begin{equation*}
\begin{split}
P \left( 
\left |
\frac{1}{\Ns} 
  \sum_{i=1}^{\Ns} \vs_k (\xis)
 - E[\vs_k(\xs)]
\right |
\geq \epsilon
\right)
\leq
2 \exp \left(
-\frac{\Ns \epsilon^2}{2 \Bddansq}
\right)
\end{split}
\end{equation*}
for a fixed $\vs_k \in \Vs$, and a similar inequality can be obtained for a fixed  $\vt_l \in \Vt$. Applying the union bound over all ball centers  $\{ \vs_k \}_{k=1}^{\Ks}$ 
and $\{ \vt_l \}_{l=1}^{\Kt}$, we get that with probability at least 
\begin{equation*}
\begin{split}
1 - 2 \Ks \exp \left( -\frac{\Ns \epsilon^2}{2 \Bddansq} \right) 
- 2 \Kt \exp \left( -\frac{\Nt \epsilon^2}{2 \Bddansq} \right) 
\end{split}
\end{equation*}
we have
\begin{equation*}
\begin{split}
\left |
\frac{1}{\Ns} 
  \sum_{i=1}^{\Ns} \vs_k (\xis)
 - E[\vs_k(\xs)]
\right |  < \epsilon 
\quad
\text{ and }
\quad
\left |
\frac{1}{\Nt} 
  \sum_{j=1}^{\Nt} \vt_l (\xjt)
 - E[\vt_l(\xt)]
\right |  < \epsilon 
\end{split}
\end{equation*}
for all ball centers, which implies from \eqref{eq_ddan_hddan_bnd1}
\begin{equation*}
\begin{split}
| \D(\vs_k, \vt_l)  
- \hD(\vs_k, \vt_l)  |  
< 2 \epsilon.
\end{split}
\end{equation*}
Combining this result with the bounds in  \eqref{eq_ddan_hddan_dev}-\eqref{eq_ddan_hddan_term3}, we get
\begin{equation*}
\begin{split}
P & \left(
\sup_{\fs \in \Fs, \ft \in \Ft, \ddan \in \Dspace} 
 |  \Ddan(\fs, \ft)  - \hDdan(\fs, \ft)  | \leq 6 \epsilon
 \right ) \\
& \geq
 1 - 2 \Ks \exp \left( -\frac{\Ns \epsilon^2}{2 \Bddansq} \right) 
- 2 \Kt \exp \left( -\frac{\Nt \epsilon^2}{2 \Bddansq} \right) .
\end{split}
\end{equation*}
Replacing $ \epsilon$ with $\epsilon/6$, we get the statement of the lemma.
\end{proof}


\section{Proof of Theorem \ref{thm_main_result_dann}}
\label{pf_thm_main_result_dann}

\begin{proof}

We begin by bounding the expected target loss as
\[
 \Lt(\ft, \h) \leq \Ls (\fs, \h) +  \LLsdan \, \Ddan(\fs, \ft)
\]
using Assumption \ref{assum_existence_LLsdan}. It follows that
\begin{equation}
\label{eq_Lt_Lw_RADA_ddan}
\begin{split}
\Lt(\ft, \h) &= \alpha \Lt(\ft, \h) +  (1-\alpha) \Lt(\ft, \h) \\
	&\leq \alpha \Lt(\ft, \h) +  (1-\alpha) \left( \Ls (\fs, \h) + \LLsdan \, \Ddan(\fs, \ft) \right)\\
&= \Lw(\fs, \ft, \h)+ (1-\alpha) \LLsdan \, \Ddan(\fs, \ft) .
\end{split}
\end{equation}

We next aim to upper bound the expected loss $\Lw(\fs, \ft, \h)$ and the expected distribution distance $ \Ddan(\fs, \ft) $ in terms of their empirical counterparts. It follows from Assumptions \ref{assum_Ax_Atheta} and \ref{assum_actddan_cont_Lip} that the source hypothesis space $\Gs=\Hs \circ \Fs$, the target hypothesis space $\Gt= \Hs \circ \Ft$, the source domain discriminator space $\Vs = \Dspace \circ \Fs$ and the target domain discriminator space $\Vt = \Dspace \circ \Ft$ are compact with respect to the metrics $\ds, \dt, \dVs, \dVt$ respectively, which can be shown by following similar steps as in the proof of Lemma \ref{lem_Fs_Ft_HFs_HFt_comp} in Appendix \ref{pf_lem_Fs_Ft_HFs_HFt_comp}.

Due to the compactness of $\Gs, \Gt$ and the assumptions on the classification loss function $\loss$, we have 
\begin{equation}
\label{eq_cons_lem_weight_loss_gen}
\begin{split}
&P\left (\sup_{\fs \in \Fs, \ft \in \Ft, \h \in \Hs} |  \Lw( \fs, \ft, \h ) - \hLw( \fs,  \ft, \h )   |  \leq \epsilon \right) \\
& \geq 1 - 2 \N( \Hs \circ \Ft, \frac{\epsilon}{8 \alpha \Lls}, \dt) e^{-\frac{\Mt \epsilon^2}{8 \alpha^2 \bls^2}} 
-2 \N( \Hs \circ \Fs, \frac{\epsilon}{8 (1-\alpha) \Lls}, \ds) e^{-\frac{\Ms \epsilon^2}{8 (1-\alpha)^2 \bls^2}}
\end{split}
\end{equation}
from Lemma \ref{lem:weight_loss_gen}. Similarly, the compactness of $\Vs, \Vt$ together with Assumption \ref{assum_ddan_bdd} implies that 
\begin{equation}
\label{eq_cons_lem_ddan_hhdan_dev}
\begin{split}
P & \left(
\sup_{\fs \in \Fs, \ft \in \Ft, \ddan \in \Dspace} 
 |  \Ddan(\fs, \ft)  - \hDdan(\fs, \ft)  | \leq  \epsilon
 \right ) \\
& \geq
 1 - 2 \N(\Vs, \frac{\epsilon}{6}, \dVs)  \exp \left( -\frac{\Ns \epsilon^2}{72 \Bddansq} \right) 
- 2 \N(\Vt, \frac{\epsilon}{6}, \dVt) \exp \left( -\frac{\Nt \epsilon^2}{72 \Bddansq} \right) 
\end{split}
\end{equation}
due to Lemma \ref{lem_ddan_hddan_dev}. 

Combining the results in \eqref{eq_Lt_Lw_RADA_ddan}, \eqref{eq_cons_lem_weight_loss_gen}, and \eqref{eq_cons_lem_ddan_hhdan_dev}, we get that with probability at least
\begin{equation}
\label{eq_prob_exp_thm4}
\begin{split}
1 &-2 \N( \Hs \circ \Fs, \frac{\epsilon}{8 (1-\alpha) \Lls}, \ds) e^{-\frac{\Ms \epsilon^2}{8 (1-\alpha)^2 \bls^2}} 
- 2 \N( \Hs \circ \Ft, \frac{\epsilon}{8 \alpha \Lls}, \dt) e^{-\frac{\Mt \epsilon^2}{8 \alpha^2 \bls^2}} 
\\
&- 2 \N(\Vs, \frac{\epsilon}{6}, \dVs)  \exp \left( -\frac{\Ns \epsilon^2}{72 \Bddansq} \right) 
- 2 \N(\Vt, \frac{\epsilon}{6}, \dVt) \exp \left( -\frac{\Nt \epsilon^2}{72 \Bddansq} \right) 
\end{split}
\end{equation}
the expected target loss is bounded as
\begin{equation*}
\begin{split}
\Lt(\ft, \h) & \leq \hLw(\fs, \ft, \h)+ (1-\alpha) \LLsdan \, \hDdan(\fs, \ft) 
 + (1- \alpha) \LLsdan \epsilon +  \epsilon.
\end{split}
\end{equation*}

In the sequel, we examine each one of the terms in the probability expression in \eqref{eq_prob_exp_thm4}. As for the covering numbers of $\Hs \circ \Fs$ and $\Hs \circ \Ft$, Assumptions \ref{assum_Ax_Atheta}, \ref{assum_bnd_act_val_op}, and \ref{assum_actddan_cont_Lip} ensure that the result in Lemma \ref{assum_bnd_act_val_op} applies to this setting as well, which implies that the rate of growth of $\N( \Hs \circ \Fs, \epsilon, \ds) $ and $\N( \Hs \circ \Ft, \epsilon, \dt) $ with $\numL$ and $\dcom$ is upper bounded by
\begin{equation*}
\begin{split}
O\left( \left( \frac{  \numL }{\epsilon}  \right)^{\dcom^2 \numL}   (c \dcom)^{\dcom^2 \numL^2} \right)
\end{split}
\end{equation*}
due to Corollary \ref{cor_covnum_rate}. Then, following the very same steps as in the proof of Theorem \ref{thm_main_result_da_mmd}, we get that  upper bounding the  weight parameter $\alpha$ by 
\begin{equation*}
\begin{split}
\alpha
= O \left(
\left(
\frac
{\Mt \epsilon^2}
{\dcom^2 \numL \log \left( \frac{  \numL }{\epsilon} \right)  \, 
+ 
\dcom^2 \numL^2 \log(\dcom) }
\right)^{1/2}
\right),
\end{split}
\end{equation*}
together with scaling $\Ms $ at rate
\begin{equation*}
\begin{split}
\Ms = 
 O \left(
\frac{\dcom^2 \numL \, \log\left( \frac{  \numL }{\epsilon}  \right)
+ 
\dcom^2 \numL^2 \log(\dcom) }
{\epsilon^2}
\right)
\end{split}
\end{equation*}
ensures an upper bound on the terms 
\[
\N( \Hs \circ \Fs, \frac{\epsilon}{8 (1-\alpha) \Lls}, \ds) e^{-\frac{\Ms \epsilon^2}{8 (1-\alpha)^2 \bls^2}} 
\]
and 
\[
\N( \Hs \circ \Ft, \frac{\epsilon}{8 \alpha \Lls}, \dt) e^{-\frac{\Mt \epsilon^2}{8 \alpha^2 \bls^2}} 
\]
in the probability expression in \eqref{eq_prob_exp_thm4}. 

Then, in order to analyze the covering numbers of $\Vs$ and $\Vt$, we proceed with the following reasoning: Noting the paralel between the structures of the domain discriminator and the feature extractor network parameters considered in Assumptions \ref{assum_actddan_cont_Lip}, \ref{assum_bnd_act_val_op} and \ref{assum_bnd_act_val_op_ddan}, we observe that the function space $\Vs=\Dspace \circ \Fs$ has an identical construction to the function space $\Gs=\Hs \circ \Fs$, if the metric 
\[
\ds(\gs_1, \gs_2) = \sup_{\xs \in \Xs}  \| \gs_1(\xs) - \gs_2(\xs) \|
\]
based on the Euclidean distance in $\R^{m}$ is replaced by its counterpart
\[
\dVs(\vs_1, \vs_2) = \sup_{\xs \in \Xs}  | \vs_1(\xs) - \vs_2(\xs) | 
\]
which uses the Euclidean distance in $\R$ instead. Hence, the latter is a special case of the former that can be obtained by setting $m=1$. Consequently, the analysis of the covering number $\N ( \Hs \circ \Fs, \epsilon, \ds) $  in Corollary \ref{cor_covnum_rate} immediately applies to $\N ( \Dspace \circ \Fs, \epsilon, \dVs) $ as well, only by replacing the number of layers $\numL$ with the total number of layers $\numL+\numLdan -1$ in the cascade network formed by the combination of the feature extractor and the domain discriminator networks. We thus get
\begin{equation*}
\begin{split}
\N (  \Vs, \epsilon, \dVs) 
 = O\left( \left( \frac{  \numL + \numLdan}{\epsilon}  \right)^{\dcom^2 (\numL+ \numLdan)}   (c \dcom)^{\dcom^2 (\numL + \numLdan)^2} \right)
\end{split}
\end{equation*}
which yields
\begin{equation}
\label{eq_NVs_exp}
\begin{split}
&\N(\Vs, \frac{\epsilon}{6}, \dVs)  \exp \left( -\frac{\Ns \epsilon^2}{72 \Bddansq} \right) \\
&=
O\left( \left( \frac{  \numL + \numLdan}{\epsilon}  \right)^{\dcom^2 (\numL+ \numLdan)}   (c \dcom)^{\dcom^2 (\numL + \numLdan)^2}
\exp \left( -\frac{\Ns \epsilon^2}{72 \Bddansq} \right) 
 \right) \\
 &=
 O\left( 
 \exp \left(
  \dcom^2 (\numL+ \numLdan) \log \left( \frac{  \numL + \numLdan}{\epsilon}  \right)       
 +
  \dcom^2 (\numL + \numLdan)^2 \log(c \dcom) 
 -
  \frac{\Ns \epsilon^2}{72 \Bddansq} 
 \right)
 \right).
\end{split}
\end{equation}
We thus conclude that the sample complexity
\begin{equation*}
\begin{split}
\Ns = O \left(
\frac{  \dcom^2 (\numL+ \numLdan) \log \left( \frac{  \numL + \numLdan}{\epsilon} \right) 
+
\dcom^2 (\numL + \numLdan)^2 \log( \dcom) 
}
{\epsilon^2}
\right)
\end{split}
\end{equation*}
ensures  an upper bound on the term \eqref{eq_NVs_exp}. The same arguments also hold for the target domain, resulting in the sample complexity
\begin{equation*}
\begin{split}
\Nt = O \left(
\frac{  \dcom^2 (\numL+ \numLdan) \log \left( \frac{  \numL + \numLdan}{\epsilon} \right) 
+
\dcom^2 (\numL + \numLdan)^2 \log( \dcom) 
}
{\epsilon^2}
\right)
\end{split}
\end{equation*}
for the number of target samples, which concludes the proof of the theorem.
\end{proof}


\bibliographystyle{IEEEbib}
\bibliography{refs}

\end{document}